\documentclass{article}

\usepackage{microtype}
\usepackage{graphicx}
\usepackage{booktabs} 
\usepackage{amsfonts}       
\usepackage{nicefrac}
\usepackage{microtype}
\usepackage{subcaption}
\usepackage{booktabs} 
\usepackage{multirow}
\usepackage{float}

\usepackage{hyperref}
\usepackage{bm}

\usepackage{enumitem} 
\usepackage{amsfonts} 

\usepackage{algorithm}
\usepackage{algorithmic}

\usepackage[accepted]{icml2025}

\usepackage{amsmath}
\usepackage{amssymb}
\usepackage{mathtools}
\usepackage{amsthm}
\usepackage{hyperref}
\usepackage{url}

\usepackage[utf8]{inputenc} 
\usepackage[T1]{fontenc}    
\usepackage{hyperref}       
\usepackage{url}            
\usepackage{booktabs}       
\usepackage{amsfonts}       
\usepackage{nicefrac}       
\usepackage{microtype}      
\usepackage{xcolor}         
\usepackage{amsmath,amssymb}
\usepackage{graphicx}
\usepackage{microtype}
\usepackage{booktabs} 
\usepackage{multirow}
\usepackage{float}
\usepackage{hyperref}
\usepackage{bm}

\usepackage{enumitem} 
\usepackage{amsfonts} 

\usepackage{algorithm}
\usepackage{algorithmic}

\usepackage[capitalize,noabbrev]{cleveref}

\theoremstyle{plain}

\theoremstyle{definition}

\theoremstyle{remark}

\usepackage[textsize=tiny]{todonotes}

\icmltitlerunning{Contextual bandits with entropy-based human feedback}

\begin{document}

\twocolumn[
\icmltitle{Contextual bandits with entropy-based human feedback}



\icmlsetsymbol{equal}{*}

\begin{icmlauthorlist}
\icmlauthor{Raihan Seraj}{yyy}
\icmlauthor{Lili Meng }{ind}
\icmlauthor{Tristan Sylvain}{comp}
\end{icmlauthorlist}

\icmlaffiliation{yyy}{Department of Electrical and Computer Engineering, McGill University, Canada}
\icmlaffiliation{comp}{Borealis AI, Montreal, Canada}
\icmlaffiliation{ind}{Independent Researcher}

\icmlcorrespondingauthor{Raihan Seraj}{raihan.seraj@mail.mcgill.ca}

\icmlkeywords{Human Feedback, Contextual Banidts, Reinforcement Learning}

\vskip 0.3in
]



\printAffiliationsAndNotice{\icmlEqualContribution} 

\begin{abstract}
In recent years, preference-based human feedback mechanisms have become essential for enhancing model performance across diverse applications, including conversational AI systems such as ChatGPT. However, existing approaches often neglect critical aspects, such as model uncertainty and the variability in feedback quality. To address these challenges, we introduce an entropy-based human feedback framework for contextual bandits, which dynamically balances exploration and exploitation by soliciting expert feedback only when model entropy exceeds a predefined threshold. Our method is model-agnostic and can be seamlessly integrated with any contextual bandit agent employing stochastic policies. Through comprehensive experiments, we show that our approach achieves significant performance improvements while requiring minimal human feedback, even under conditions of suboptimal feedback quality. This work not only presents a novel strategy for feedback solicitation but also highlights the robustness and efficacy of incorporating human guidance into machine learning systems. Our code is publicly available: \url{https://github.com/BorealisAI/CBHF}
\end{abstract}

\section{Introduction}
\label{sec:introduction }

Contextual bandits (CB) have become a fundamental framework for personalized decision-making across diverse domains including recommendation systems~\citep{li2010contextual,bouneffouf2020survey}, healthcare~\citep{yu2024careforme}, and finance~\citep{zhu2021online}. While traditional CB approaches leverage contextual information to optimize actions, their heavy reliance on implicit feedback signals like clicks introduces inherent limitations due to the biased and incomplete nature of such data~\citep{qi2018bandit}.

This work investigates how explicit human feedback can enhance CB performance. Building on successful integrations of human guidance in reinforcement learning~\citep{christiano2017deep,macglashan2017interactive} and conversational AI~\citep{achiam2023gpt}, we distinguish two primary feedback paradigms: (1) \emph{action-based feedback}, where experts directly prescribe optimal actions for specific contexts~\citep{osa2018algorithmic,li2023reinforcement}, and (2) \emph{preference-based feedback}, where humans compare pairs of learner-generated actions to express relative preferences~\citep{christiano2017deep,saha2023dueling}. While action-based methods require precise expert knowledge, we focus on preference feedback for its practical advantages in scalable data collection, notably its reduced cognitive load on human evaluators. However, this operational simplicity introduces two critical challenges: (1) variable human feedback quality, and (2) model uncertainty propagation. Our central research question emerges: \textbf{How can we effectively incorporate preference-based human feedback into contextual bandits while addressing these fundamental challenges?}

To address this, we propose an entropy-based feedback mechanism that selectively queries an oracle when the agent’s policy exhibits high uncertainty. By dynamically adjusting feedback requests based on entropy, our approach effectively balances exploration and exploitation, reducing unnecessary queries while ensuring informative guidance when needed. This selective feedback mechanism allows the agent to refine its policy more efficiently, leading to tighter regret bounds and improved performance. Furthermore, we introduce two complementary feedback integration strategies: \textit{Action Recommendation} (AR), where experts suggest context-specific actions, and \textit{Reward Manipulation} (RM), where penalties adjust the reward signal for non-recommended actions. By optimizing the timing of human input through adaptive entropy-based solicitation, our framework enhances learning efficiency while mitigating the impact of imperfect human feedback.

Our key contributions include:

\begin{itemize}
\item A unified framework for human-CB collaboration with theoretical analysis comparing AR and RM strategies.

\item An entropy-based solicitation criterion that improves learning efficiency while providing insights into human-algorithm interaction dynamics.

\item Empirical characterization of feedback quality impacts, revealing performance robustness across expert competence levels.

\end{itemize}

Our findings advance both practical CB implementations and theoretical understanding of human feedback integration. By establishing guidelines for effective human-AI collaboration in bandit settings, this work bridges critical gaps between algorithmic decision-making and human expertise.

\section{Related works}
\textbf{Contextual bandits}
Contextual bandits have diverse applications in recommendation systems~\citep{li2010contextual, xu2020contextual}, healthcare~\citep{yu2024careforme}, finance~\citep{zhu2021online}, and other fields~\citep{bouneffouf2020survey}. CBs are a variant of the multi-armed bandit problem where each round is influenced by a specific context, and rewards vary accordingly. This adaptability makes CBs valuable for enhancing various machine learning methods, including supervised learning~\citep{sui2020bayesian}, unsupervised learning~\citep{sublime2018collaborative}, active learning~\citep{bouneffouf2014contextual}, and reinforcement learning~\citep{intayoad2020reinforcement}.

To tackle CB challenges, several algorithms have been developed, such as LINUCB~\citep{li2010contextual}, Neural Bandit~\citep{allesiardo2014neural}, and Thompson sampling~\citep{agrawal2013thompson}. These typically assume a linear dependency between the expected reward and its context. Despite these advancements, CBs often rely on implicit feedback, like user clicks, leading to biased and incomplete evaluations of user preferences~\citep{qi2018bandit}. This reliance complicates accurately gauging user responses and tailoring the learning process.

\textbf{Human feedback in the loop}
Recent advancements in human-in-the-loop methodologies have shown significant successes in real-life applications, such as ChatGPT via reinforcement learning with human feedback (RLHF)~\citep{macglashan2017interactive}, as well as in robotics~\citep{argall2009survey} and health informatics~\citep{holzinger2016interactive}. 

Preference-based feedback can be categorized into three groups: i) action-based preferences~\citep{furnkranz2012preference}, where experts rank actions, ii) state preferences~\citep{wirth2014learning}, and iii) trajectory preferences~\cite{busa2014preference,novoseller2020dueling}. Action-based feedback from humans is explored in~\citep{mandel2017add}, where experts add actions to a reinforcement learning agent to boost performance. Other forms of explicit human feedback include reward shaping~\citep{xiao2020fresh,biyik2022learning,ibarz2018reward,arakawa2018dqn}. These approaches however do not account for acquiring feedback based on the learner's uncertainty or the impact of varying levels of feedback on performance.

\textbf{Contextual bandits with human feedback} Human-in-the-Loop Reinforcement Learning addresses the bias problem of implicit feedback in contextual bandits. The exploration of learning in multi-armed bandits with human feedback is discussed in~\citep{tang2019bandit}, where a human expert provides biased reports based on observed rewards. The learner's goal is to select arms sequentially using this biased feedback to maximize rewards, without direct access to the actual rewards.

Preference-based feedback in contextual and dueling bandit frameworks has been explored in previous studies~\citep{sekhari2023contextual,dudik2015contextual,saha2021optimal,wu2023borda}. The learner presents candidate actions and receives noisy preferences from a human expert, focusing on minimizing regret and active queries. In contrast, we consider a setup where the learner receives direct feedback from human experts and show how the fraction of active queries varies with different sets of experts.

\textbf{Active learning in contexual bandits} Active learning~\citep{judah2014active} enhances performance by selectively querying the most informative data points for labeling, rather than passively receiving labels for randomly or sequentially presented data. In the context of bandit algorithms, active learning has been employed to optimize the exploration-exploitation trade-off by guiding the algorithm to request feedback or labels when it is most uncertain about an action’s outcome~\citep{taylor2009transfer}. For example, \cite{bouneffouf2014contextual} integrated active learning with Thompson sampling and UCB algorithms in contextual bandits, resulting in improved sample efficiency.

While active learning traditionally assumes access to a small set of labeled examples alongside abundant unlabeled data, the contextual bandit setting differs fundamentally in that all feedback is inherently partial - we only observe outcomes for the chosen action. This structural divergence precludes direct application of standard active learning techniques. Nevertheless, we draw methodological inspiration from active learning's core philosophy of strategic information acquisition.

In our work, we build on this idea by combining active learning principles with human feedback, utilizing an entropy-based mechanism to query feedback when necessary. By incorporating these selective querying strategies into our contextual bandit framework, we aim to more effectively balance exploration and exploitation, particularly in scenarios where human feedback is noisy or costly. This approach not only improves sample efficiency but also helps mitigate the challenges posed by varying feedback quality.

\section{Method}
\begin{figure*}[htbp]
    \centering
  
     \includegraphics[width=.65\linewidth]{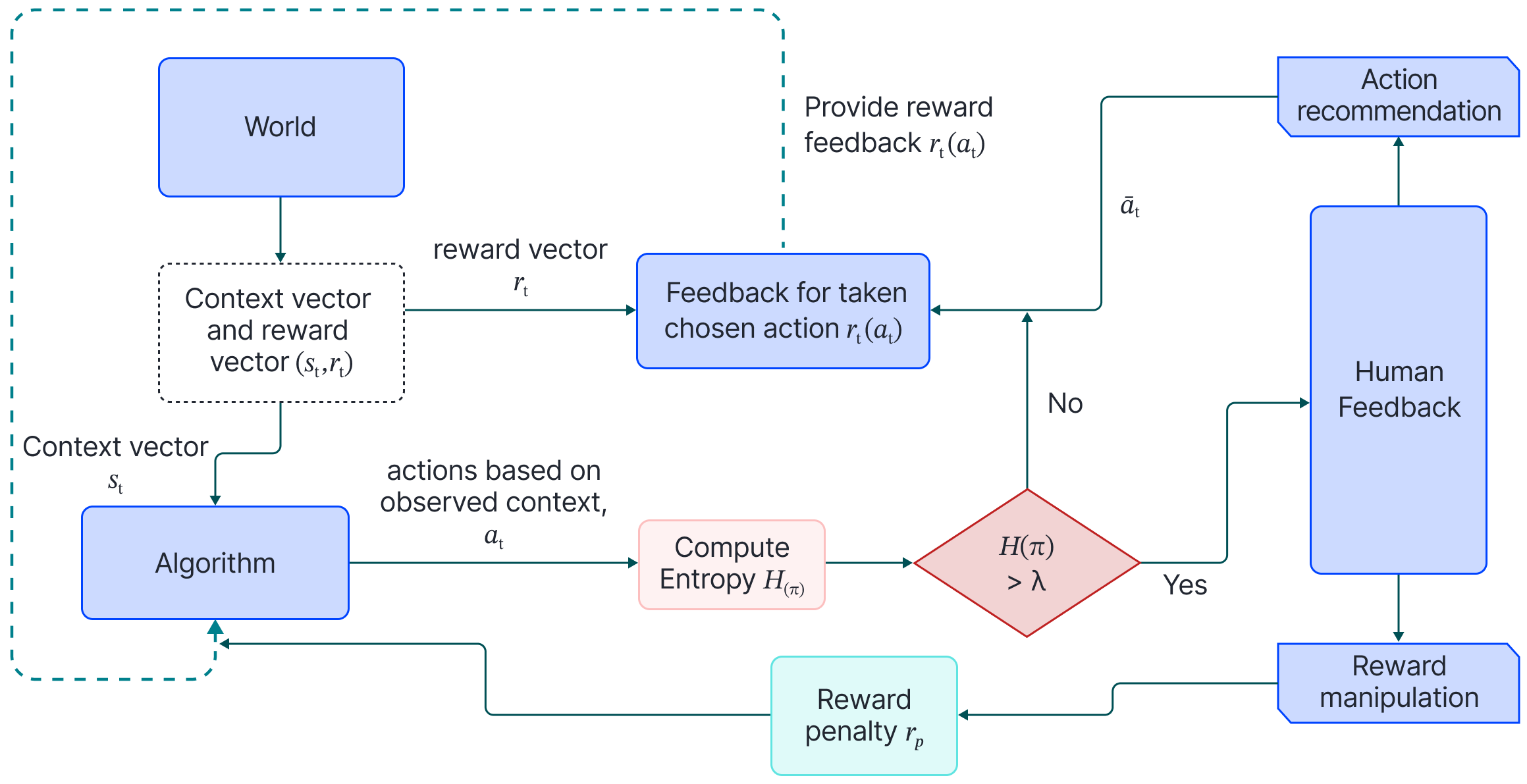}
    \caption{Overview of the architecture. Our framework builds upon a standard contextual bandit setup. The right side of the figure illustrates the human feedback incorporation mechanism, which can be integrated through either \textit{reward manipulation} (directly modifying the bandit's reward signal) or \textit{action recommendation}, which constrains the set of available actions. Rather than triggering human feedback at fixed intervals, we propose an adaptive approach: querying feedback only when the action policy (Eq.~\ref{eq:policy_entropy}) exceeds a predefined uncertainty threshold. This strategy ensures feedback is solicited when it is most valuable, improving efficiency and decision-making.}
    \label{fig:overview}
   
\end{figure*}

The following section provides a description of our method and its subcomponents. A comprehensive representation of the approach is shown in Figure~\ref{fig:overview}. Algorithm~\ref{alg:example} describes our method.

\subsection{Contextual bandit formulation}
We consider an online stochastic contextual bandit framework where at each round $t$, the world generates a context-reward pair $(s_t,r_t)$ sampled independently from a stationary unknown distribution $\mathcal D$. Here $s_t\in \mathcal S= \mathbb{R}^m$ is an $m$ dimensional real valued vector and $r_t=(r_t(1),\dots,r_t(k))\in \{0,1\}^k$ is a $k$-dimensional vector where each element can take values $0$ or $1$. The agent then chooses an action $a_t\in\{1,\dots,k\}$ 
according to a policy $\pi:\mathcal{S}\mapsto \{1,\dots,k\}$ and the environment reveals the reward $r_t(a_t)\in\{0,1\}$.

The objective of the agent is to find a policy $\pi\in \Pi$ that will minimize the cumulative expected regret given by 
\begin{equation}
    \mathbb{E}\Bigl[\sum_{t=1}^T\bigl(r_t(\pi^*(s_t)-r_t(a_t)\bigr)\Bigr],
\end{equation}
where $\pi^*$ denotes the optimal strategy of selecting actions given context $s_t$.
The problem setup described above bears a strong resemblance to a multi-label or multiclass classification problem, where $r_t(a_t)=1$ indicates the correct label choice and $0$ otherwise. However, a key distinction lies in the learner's lack of access to the correct label or label set for each observation. Instead, the learner only discerns whether the chosen label for an observation is correct or incorrect. 



\subsection{Incorporating entropy based human feedback}

In contextual bandits, feedbacks are provided in the form of a predetermined reward signal provided by the designer. These reward signals are not well defined for complex decision-making problems~\citep{blanchard2023adversarial,dragone2019deriving}, and are often learned from data.  An alternative to learning a reward function from data is to obtain preference-based feedback from humans and learn the underlying reward function that the human expert optimizes~\citep{sekhari2024contextual}. In this work, we consider the setup in which the human expert has sufficient expertise and valuable insights stemming from their experience and domain knowledge to provide direct feedback to the learner. These feedbacks can directly impact the actions that a contextual bandit learner takes or the rewards it receives. However, the quality of such explicit feedback may vary depending on the expertise levels of different individuals.
 We provide two ways in which human experts can provide feedback to the contextual bandit learner: i) Action Recommendation through direct supervision (AR) ii) Reward Manipulation (RM). In certain applications, a human expert can directly control the actions that the agent takes; in these cases, feedback in the form of action recommendations (AR) is useful. Conversely, in other applications where the human expert cannot directly influence the agent's actions, feedback through reward manipulation is more beneficial. We describe each of these different feedbacks below. 
 \begin{algorithm}
\caption{Enropy Based - CBHF}
\label{alg:example}
\begin{algorithmic}[1]
\REQUIRE Input parameters: entropy threshold $(\lambda)$, feedback-type $(fb)$, round-number $(n)$, contextual bandit agent $(\mathcal{A})$, human expert quality $(q_t)$
\ENSURE Output: \emph{mean cumulative regret}
\STATE Initialize \emph{mean cumulative regret}$\gets 0$
\FOR{$t = 1$ to $n$}
    \STATE Get context, reward vector $(s_t,r_t)\gets\omega$
    \STATE Get actions and action distribution from the learner ($a_t,\pi(s_t)) \gets \mathcal{A}(s_t)$
    \STATE Compute $H(\pi(s_t))$
    \IF{$H(\pi(s_t))>\lambda$}
    \IF{$fb$ == \emph{AR}}
    \STATE $\hat a\gets\mathcal{E}(s_t,q_t)$
    \STATE $a_t\gets \hat a$
    \STATE $r\gets r_t(a_t)$
    \ELSIF{$fb$== \emph{RM}}
    \STATE $r_p \gets\mathcal E(s_t,q_t)$
    \STATE $r\gets r_t(a_t)+r_p$
    \ENDIF
    \ELSE
    \STATE $r\gets r_t(a_t)$
    \ENDIF
    \STATE Update Agent $\mathcal A$ policy $\pi$ with feedback $r$
    \STATE \emph{mean cumulative regret}$\gets$ evaluate agent $\mathcal A$
\ENDFOR
\STATE Return \emph{mean cumulative regret}
\end{algorithmic}
\end{algorithm}

\subsubsection{Action recommendation via direct supervision}
\label{sec:action_recommendation}

In this form of feedback, the human expert explicitly instructs the actions to take for a given context. We assume that the algorithm always accepts the recommended action. Let $\hat{a}_t$ be a set of actions recommended by the human expert $\mathcal{E}^{\mathrm{AR}}$ for a given context $s_t$ and expert quality $q_t$, where $q_t\in [0,1]$, we elaborate more on the expert quality in Section~\ref{subsec:exp_quality}.  When the expert recommends a set of actions, the learning algorithm randomly chooses an action from the recommended set. 
The final reward $r_t^f$ received by the learner is given by: 
\begin{align}
    &\hat{a}_t = \mathcal{E}^{\mathrm{AR}}(s_t,q_t) \\
    &a_t\sim\mathrm{Uniform}(\hat a_t)\\
    &r_t^f = r_t(a_t)
\end{align}

\subsubsection{Reward manipulation}

\label{sec:reward_penalty}
In this form of feedback, the human expert $\mathcal E^{\mathrm{RM}}$ gives an additional reward penalty when the learner chooses an action not recommended by the expert. Let $r_p$ be the fixed reward penalty for unrecommended actions. Let $a_t$ be the action chosen by the learner in round $t$, and $\hat{a}_t$ be the set of recommended actions of the expert. The final reward $r_t^f$ received by the learner is given by:

\begin{align}
    &r_p=\mathcal{E}^{\mathrm{RM}}(s_t,q_t)\\
    &r_t^f = \begin{cases}
        r_t(a_t) + r_p & \text{if } a_t \notin \hat{a}_t \\
        r_t(a_t) & \text{otherwise}
    \end{cases}
\end{align}

\subsection{When to seek human feedback?}
\label{subsec:when_to_seek_hf}
An important question that naturally arises when integrating human feedback into the contextual bandit algorithm is when the algorithm will actively seek out such feedback. In the contextual duelling bandit setup in~\citep{di2024nearly}, the algorithm presents two options to the human and asks them to choose a preferred one based on a given context. In the case of model misspecification, where the underlying reward function assumed by the algorithm matches the true rewards generated by human preferences, the algorithm can actively query the human expert to obtain feedback on the predicted reward or ranking~\citep{yang2023contextual}. In our work, we take a different approach in which the learner seeks expert feedback based on the uncertainty of the model. The model computes the entropy of the policy at each round $t$ which quantifies the degree of unpredictability in the policy's decision making process using the following expression 

\begin{equation}
H(\pi)=-\sum_{a_t}\pi(a_t\mid s_t)\log(\pi(a_t\mid s_t)),
\label{eq:policy_entropy}
\end{equation}
where $H(\pi)$ denotes the entropy of policy $\pi$. The model then queries for human feedback when the model entropy exceeds a predefined threshold $\lambda$. Appropriate choice of $\lambda$ will depend on the problem domain and are obtained using hyper parameter search. Our proposed entropy based approach for querying the expert depends on the learner's ability to compute an entropy for its policy. Thus for certain models when model uncertainty is not available, we can still obtain two forms of human feedback periodically, we also demonstrate the effect on model performance when these two types of human feedback are incorporated for different periods. 
\subsection{Quality of experts}
\label{subsec:exp_quality}
We consider the effect of learner's performance based on different quality of expert feedback received. We define the quality of feedback in this case as the accuracy of the expert in providing correct recommendation. We first show how the performance of the contextual bandit learner measured by the expected cumulative regret varies for different expert levels of accuracy. Let $q_t\in [0,1]$ be the probability of providing correct recommendation associated with a particular level of expert. During training, the algorithm seeks expert feedback described in Section~\ref{sec:action_recommendation} and~\ref{sec:reward_penalty} when $H(\pi)\geq \lambda$. For action recommendation via direct supervision, the expert provides the correct action with probability $q_t$ and provides a randomized action with probability $1-q_t$. For reward manipulation feedback, the expert wrongly penalizes the learner with a probability of $1-q_t$. 

\subsection{Regret Analysis for Contextual Bandits with Entropy-Based Human Feedback}

We analyze the regret bound for our proposed algorithm, which integrates entropy-based human feedback in a contextual bandit setting. The goal is to quantify the impact of selective oracle feedback on cumulative regret and derive a regret bound that captures the trade-offs.

At each round $t$, the agent observes context $s_t$ and selects an action $a_t \in \mathcal{A}$ using policy $\pi_t$. Oracle feedback is requested if the entropy $H(\pi_t)$ exceeds a threshold $\lambda$, and the observed reward $r_t(a_t)$ combines environment and feedback contributions.

The regret at time $t$ is:

\begin{equation}
\text{Regret}_t = \mathbb{E}[r_t(a_t^{*}) - r_t(a_t)],
\end{equation}

where $a^t = \pi^*(s_t)$ is the optimal action. The total regret over $T$ rounds is:
\begin{equation}
\text{Regret}(T) = \sum_{t=1}^T \text{Regret}_t.
\end{equation}

Let $p = P(H(\pi_t) > \lambda)$ denote the probability of requesting feedback, and let $q_t$ be the accuracy of oracle feedback. The regret bound is:

\begin{equation}
\begin{aligned}
\mathbb{E}[\text{Regret}(T)] \leq  
&\ O\left(\sqrt{(1 - p)T |\mathcal{A}| \log T} \right) \\
&+ O\left( \frac{p T (1 - q)}{1 - q + \log T} \right).
\end{aligned}
\end{equation}

The total regret decomposes into two parts:
\begin{equation}
\text{Regret}(T) =  \sum_{t \notin \mathcal{F}} \text{Regret}_t + \sum_{t \in \mathcal{F}} \text{Regret}_t 
\end{equation}
where $\mathcal{F}$ represents rounds where feedback is requested ($H(\pi_t) > \lambda$).

For rounds without feedback, the regret follows standard contextual bandit analysis:
\begin{equation}
\mathbb{E}[\text{Regret}_{\text{no-feedback}}(T)] \leq O\left(\sqrt{(1 - p)T |\mathcal{A}| \log T}\right).
\end{equation}

For rounds with feedback, regret reduction depends on feedback accuracy and its impact on decision quality:
\begin{equation}
\mathbb{E}[\text{Regret}_{\text{feedback}}(T)] \leq O\left( \frac{p T (1 - q)}{1 - q + \log T} \right).
\end{equation}

This regret bound highlights the trade-off between exploration and feedback solicitation. Increasing $p$ reduces the first term, leading to faster convergence, while higher feedback accuracy $q$ ensures minimal regret in feedback rounds. The entropy threshold $\lambda$ serves as a control parameter to balance feedback frequency and regret minimization. Compared to standard bandit approaches, entropy-driven feedback solicitation provides a principled mechanism to reduce regret in uncertain environments, making it highly effective for practical deployment.

\bigskip

\section{Experiments}
\label{sec:experiments}

\subsection{Experimental Setup}
In this sub-section, we present the environment settings, baselines, and experimental results. We also discuss the effect of entropy thresholds and expert accuracy on model performance.

\textbf{Algorithms and Environments Considered.}  
We conduct experiments across a range of \emph{environments} and \emph{contextual bandit agents}. The agents fall into two categories: (i) classic contextual bandit algorithms and (ii) policy-based reinforcement learning (RL) algorithms with a discount factor of $0$, focusing on immediate rewards.

\textbf{Classic Contextual Bandit Algorithms.}  
For the classic contextual bandit setup, we employ three key algorithms:  
1. \textbf{LinearUCB}~\citep{li2010contextual}: An extension of the traditional Upper Confidence Bound (UCB) algorithm~\citep{auer2002using}, where the expected reward for each action depends linearly on the context or features associated with that action.  
2. \textbf{Bootstrapped Thompson Sampling}~\citep{kaptein2014thompson}: This method replaces the posterior distribution in standard Thompson Sampling with a bootstrapped distribution, enhancing robustness by resampling historical data instead of relying on a parametric model.  
3. \textbf{EE-NET}~\citep{ban2021ee}: This approach utilizes two neural networks—one for exploration and one for exploitation—to learn a reward function and adaptively balance exploration with exploitation.

\textbf{Policy-Based Reinforcement Learning Algorithms.}  
For policy-based RL, we evaluate four algorithms, with the discount factor set to $0$ to prioritize immediate rewards:  
\textbf{Proximal Policy Optimization (PPO)}~\citep{schulman2017proximal},
\textbf{PPO with Long Short-Term Memory (PPO-LSTM)}, \textbf{REINFORCE}~\citep{williams1992simple}, \textbf{Actor-Critic}~\citep{pmlr-v80-haarnoja18b}.

\textbf{Baseline Comparison.}  
We include the \textbf{TAMER framework}~\citep{knox2009interactively} as a baseline, which allows human trainers to provide real-time feedback to the agent, supplementing the predefined environmental reward signal. In our experiments, we simulate human feedback by revealing the true labels during training.

\textbf{Expert Feedback Comparison.}  
For all contextual bandit agents, we compare two types of expert feedback as described in sections~\ref{sec:action_recommendation} and~\ref{sec:reward_penalty}. Expert feedback is solicited only during the training phase, and each learner is evaluated after five independent runs, with the mean cumulative reward reported.

\textbf{Datasets.}  
We use multi-label datasets from the Extreme Classification Repository, including Bibtex, Media Mill, and Delicious~\citep{Bhatia16}. In the contextual bandit framework, the reward function for these supervised learning datasets is defined as:
\begin{align}
r_t(a_t) = \begin{cases}
    1\quad \text{if } a_t \in y_t \\
    0 \quad \text{otherwise}
\end{cases}
\end{align}
where $y_t$ represents the set of correct labels associated with context $s_t$. These datasets are selected for their size, complexity, and diversity, making them suitable for evaluating contextual bandits with human feedback.

\textbf{Implementation Details.}  
We consider a range of entropy thresholds as hyperparameters, controlling how frequently the algorithm seeks to incorporate human feedback. The specific ranges for different datasets are detailed in Appendix~\ref{subsec:entropy_range}. We select the optimal entropy threshold and report the mean cumulative reward for each mode of human expert feedback. The code base for policy-based RL algorithms is implemented in PyTorch, adapted from~\citep{minimalRL}, while the LinearUCB and Bootstrapped Thompson Sampling implementations are adapted from~\citep{cortes2019adapting}. The hyperparameters for the RL algorithms are provided in Appendix~\ref{sec:app_hyperparams_rl}. Additionally, expert quality is varied based on values of $q_t \in [0, 1]$, where with probability $q_t$, the correct label or set of labels associated with context $s_t$ is provided to the learner, as mentioned in Section~\ref{subsec:when_to_seek_hf}.

\subsection{Variation of model performance based on different expert quality}
\begin{figure*}[h]
    \centering
    \begin{subfigure}[b]{0.24\linewidth}
        \includegraphics[width=\linewidth]{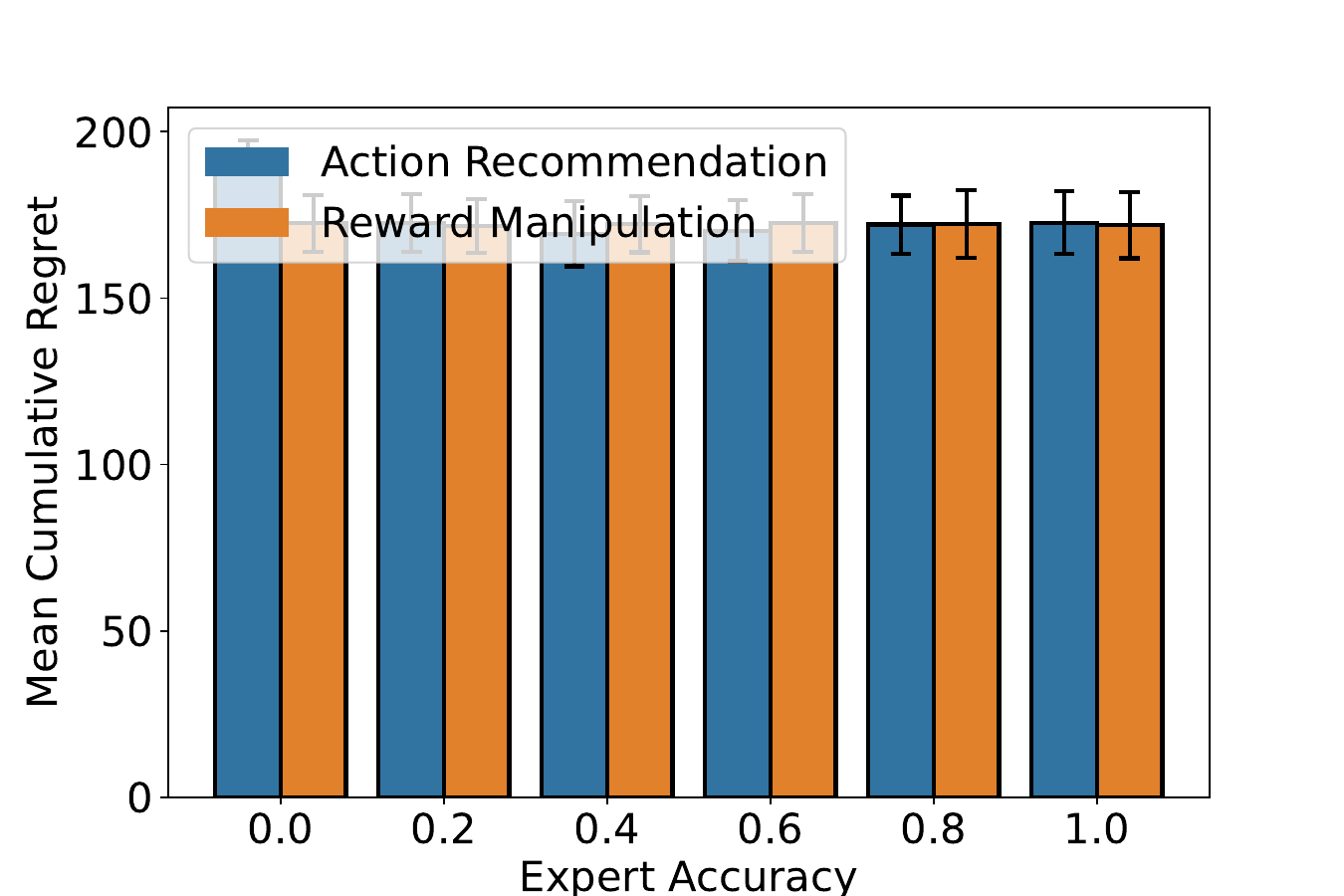}
        \caption{PPO-LSTM (Bibtex)}
        
    \end{subfigure}
    \hfill
    \begin{subfigure}[b]{0.24\linewidth}
        \includegraphics[width=\linewidth]{figures/expert_accuracy_range/bibtex/reinforce/expert_accuracy_range.pdf}
       \caption{Reinforce (Bibtex)}
    \end{subfigure}
    \hfill
    \begin{subfigure}[b]{0.24\linewidth}
        \includegraphics[width=\linewidth]{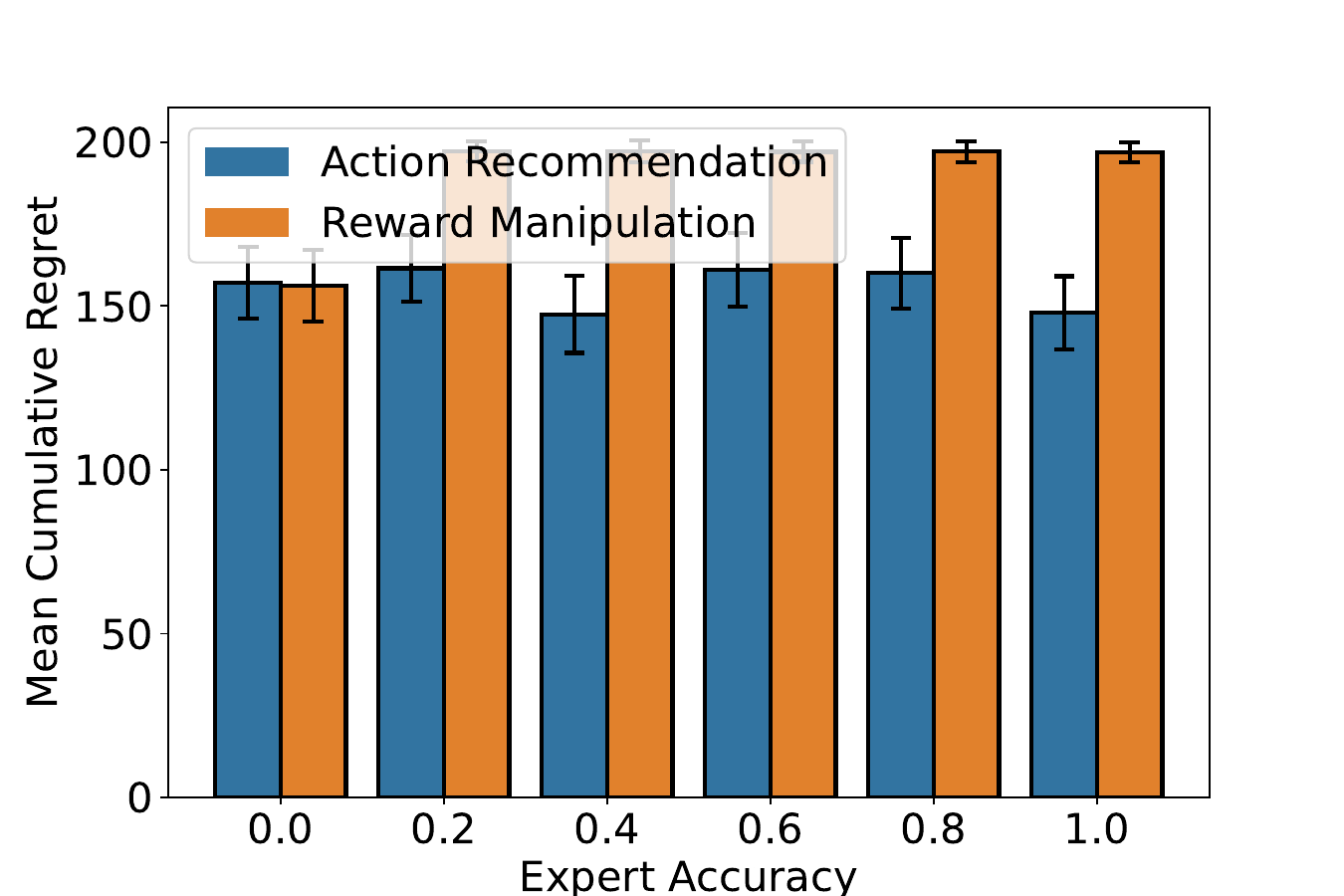}
        \caption{Actor Critic (Bibtex)}
        \end{subfigure}
        \hfill
    \begin{subfigure}[b]{0.24\linewidth}
        \includegraphics[width=\linewidth]{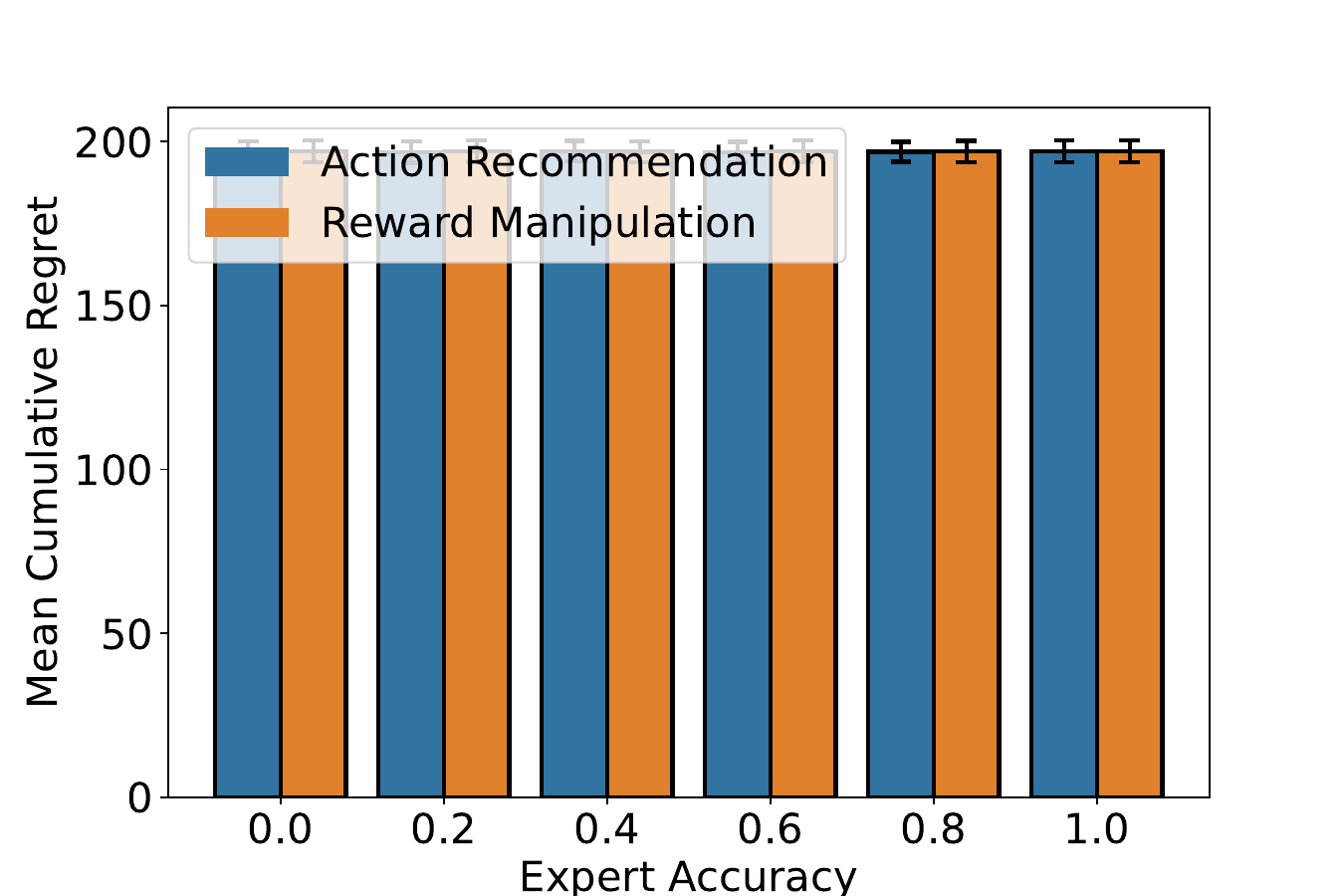}
        \caption{Bootstrapped-TS (Bibtex)}
    \end{subfigure}
        
    
    \begin{subfigure}[b]{0.24\linewidth}
        \includegraphics[width=\linewidth]{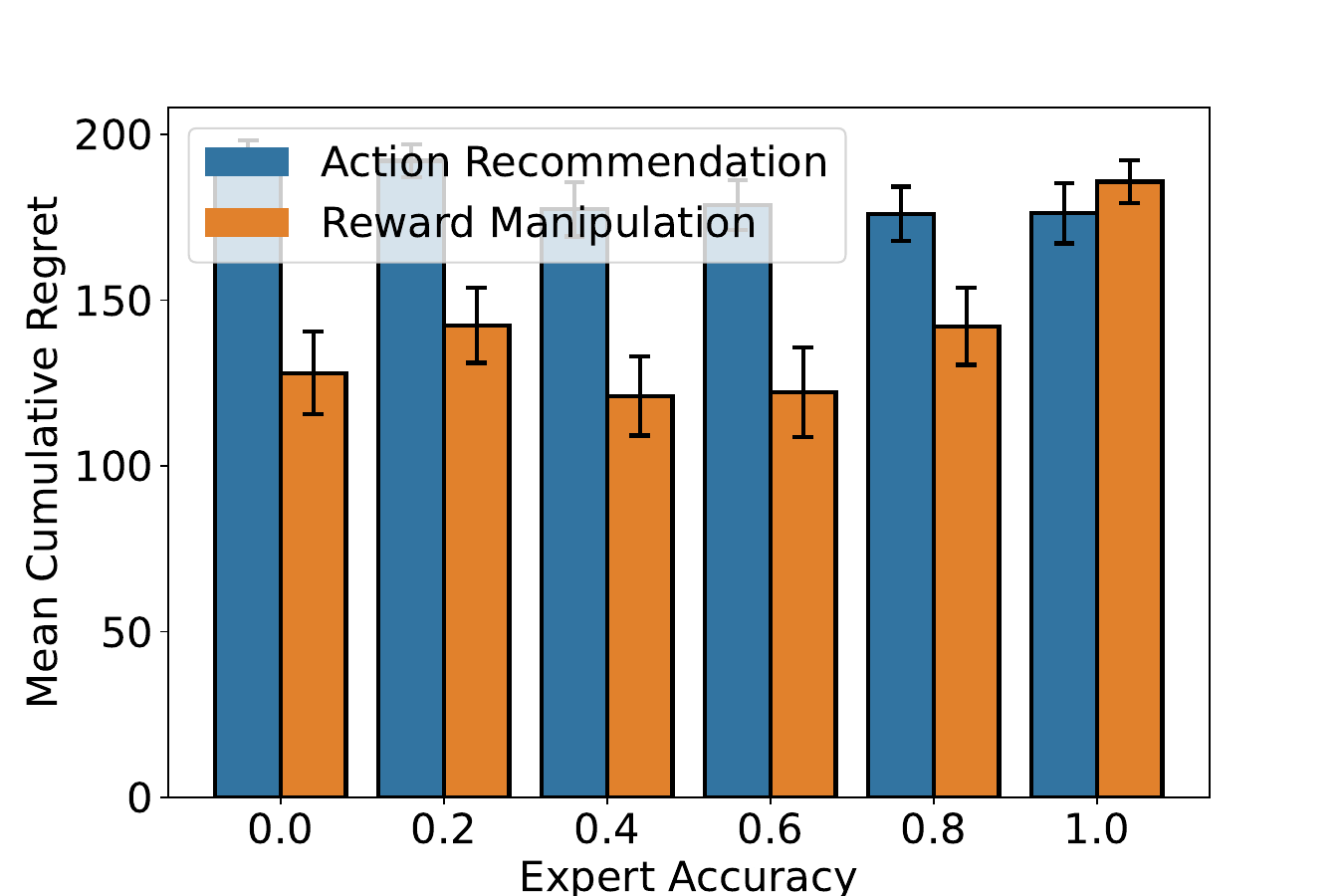}
        \caption{PPO-LSTM (Delicious)}
    \end{subfigure}
    \hfill
    \begin{subfigure}[b]{0.24\linewidth}
        \includegraphics[width=\linewidth]{figures/expert_accuracy_range/delicious/reinforce/expert_accuracy_range.pdf}
        \caption{Reinforce (Delicious)}
    \end{subfigure}
    \hfill
    \begin{subfigure}[b]{0.24\linewidth}
        \includegraphics[width=\linewidth]{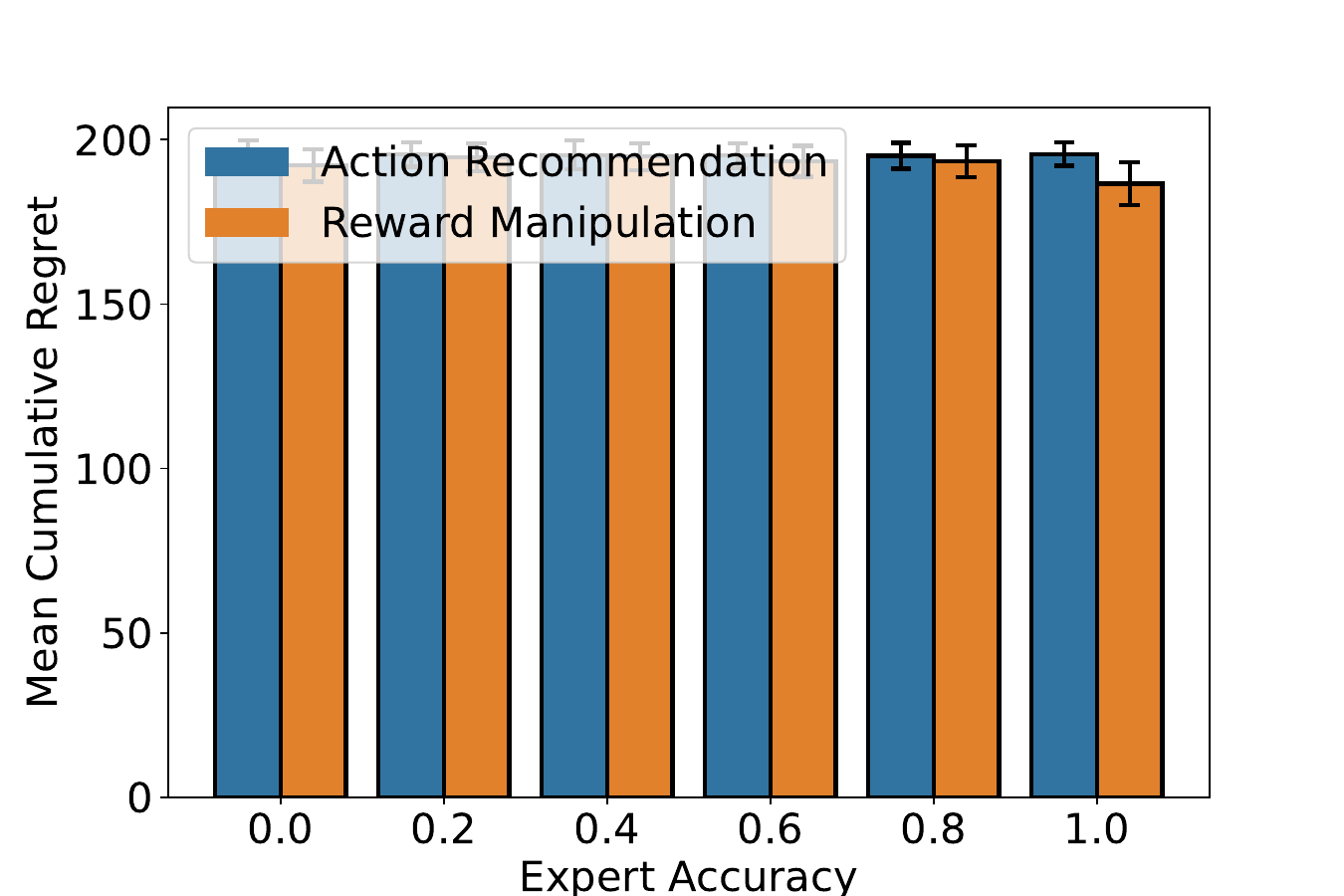}
        \caption{Actor Critic (Delicious)}
    \end{subfigure}
    \hfill
    \begin{subfigure}[b]{0.24\linewidth}
        \includegraphics[width=\linewidth]{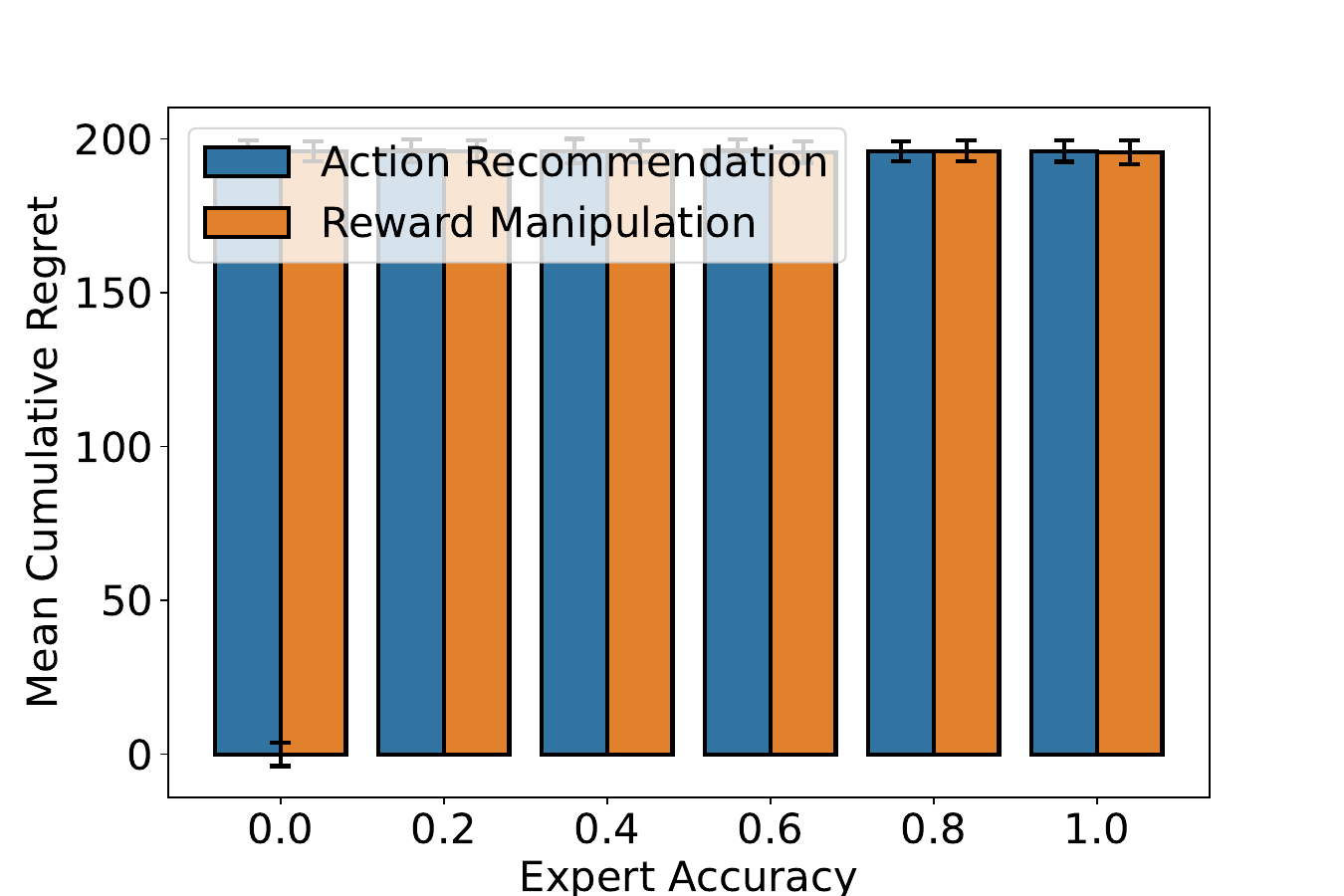}
        \caption{Bootstrapped-TS (Delicious)}
    \end{subfigure}

    \begin{subfigure}[b]{0.24\linewidth}
        \includegraphics[width=\linewidth]{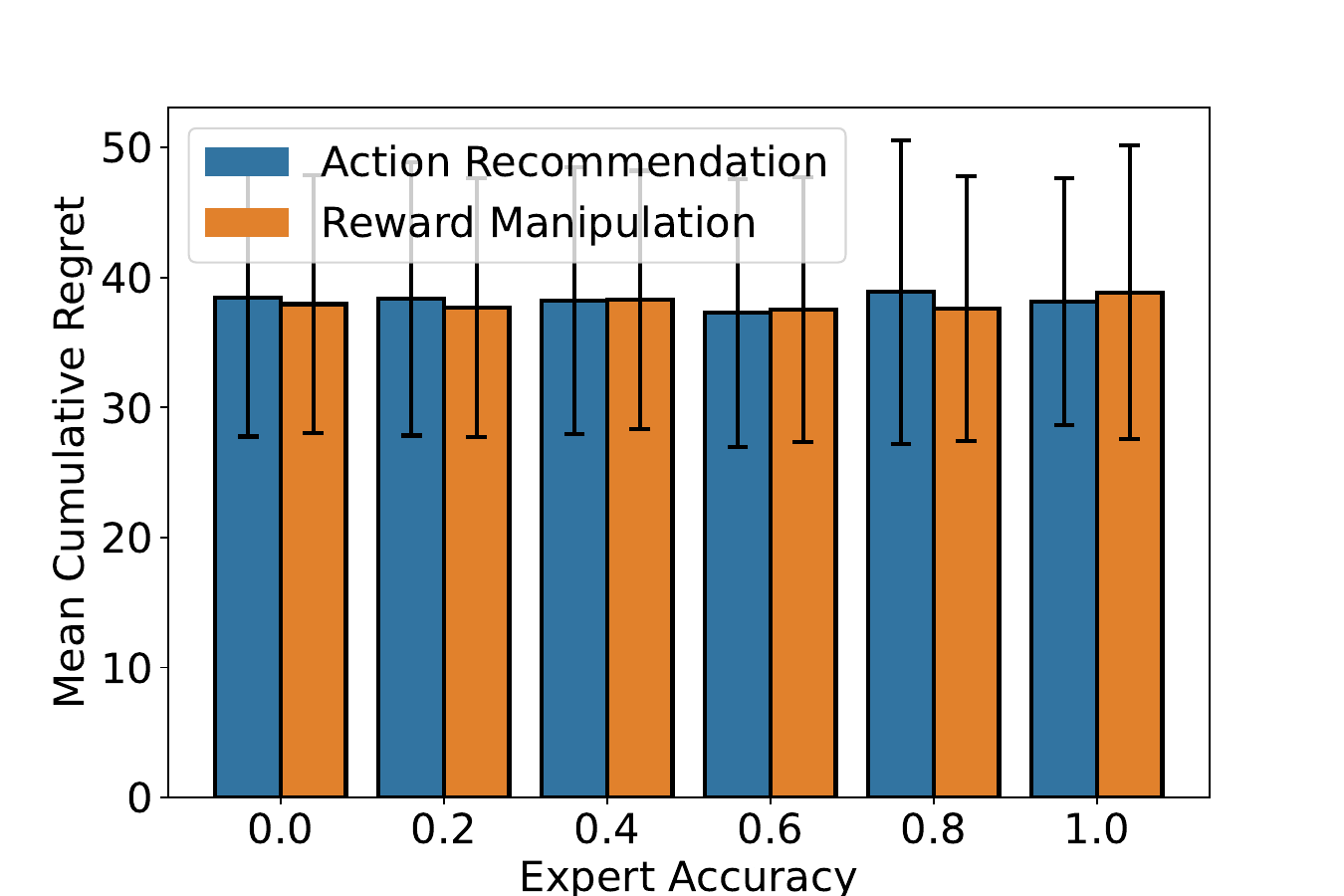}
        
        \caption{PPO-LSTM (Media Mill)}
    \end{subfigure}
    \hfill
    \begin{subfigure}[b]{0.24\linewidth}
        \includegraphics[width=\linewidth]{figures/expert_accuracy_range/media_mill/reinforce/expert_accuracy_range.pdf}
        \caption{Reinforce (Media Mill)}
    \end{subfigure}
    \hfill
    \begin{subfigure}[b]{0.24\linewidth}
        \includegraphics[width=\linewidth]{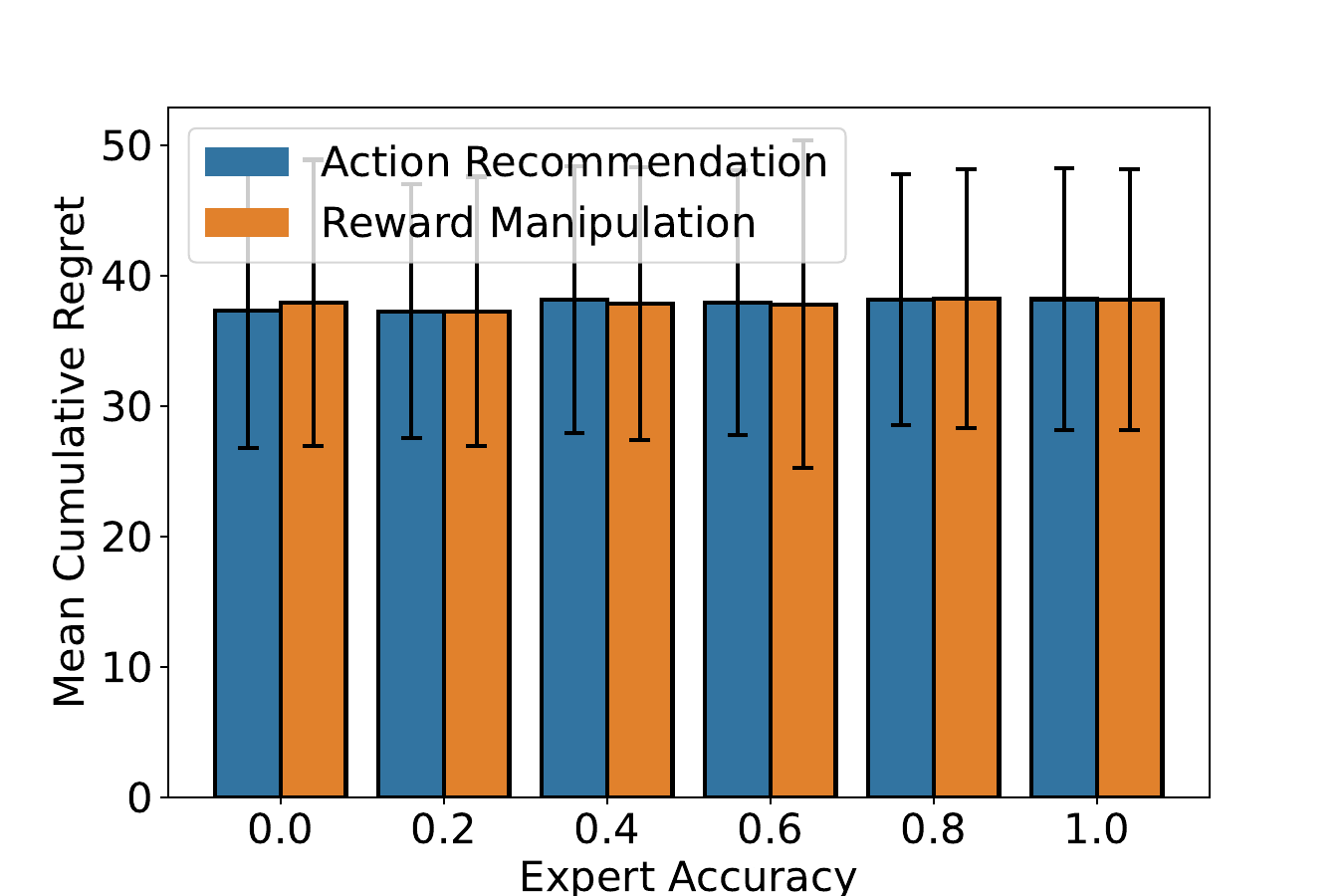}
        \caption{Actor Critic (Media Mill)}
    \end{subfigure}
    \hfill
    \begin{subfigure}[b]{0.24\linewidth}
        \includegraphics[width=\linewidth]{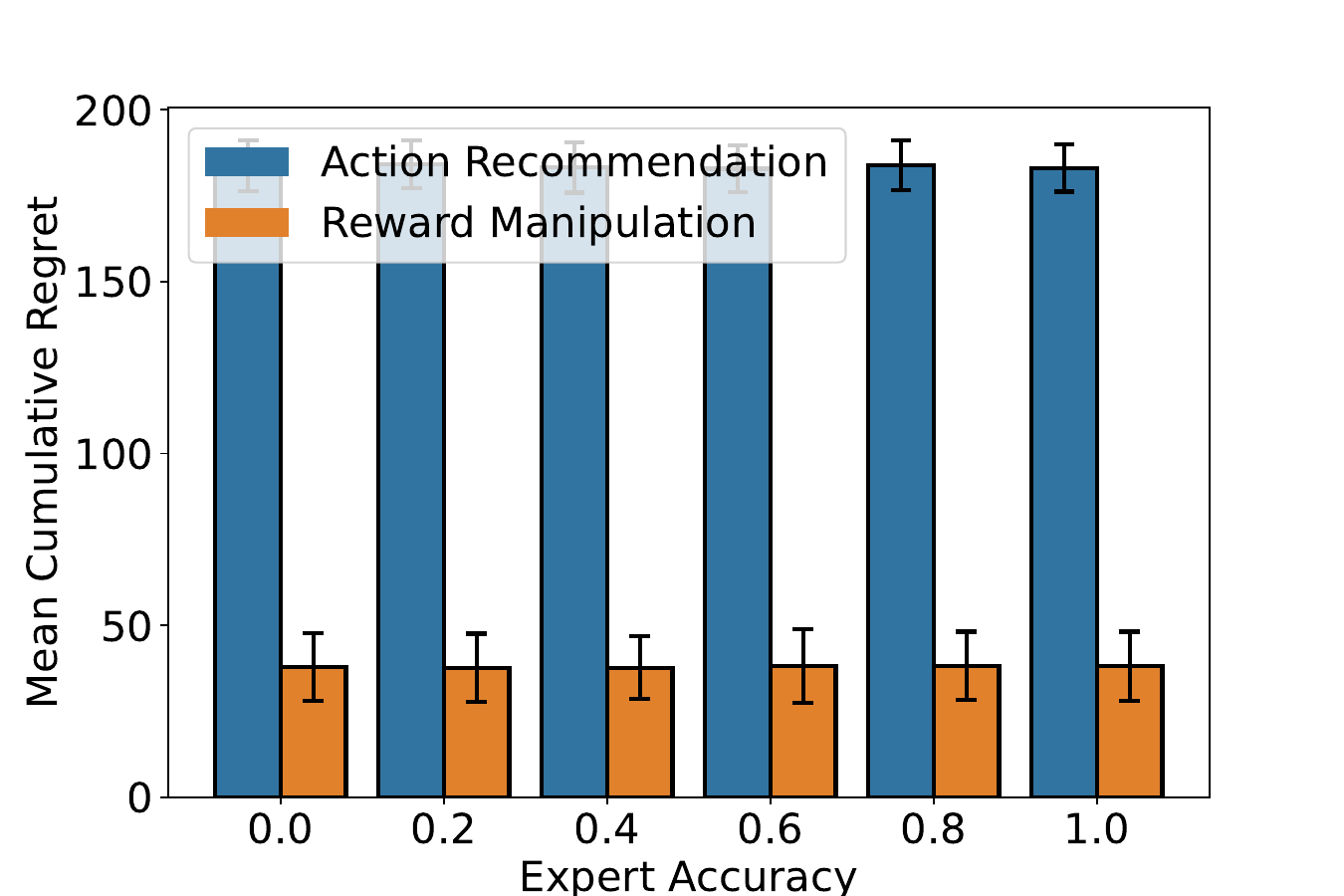}
        \caption{Bootstrapped-TS (Media Mill)}
    \end{subfigure}
       

    \caption{Comparison of expert feedback for different learners based on different expert qualities. The results show that mean cumulative regret for different datasets and algorithms vary in a different manner for the two feedback schemes considered. Higher levels of expert does not necessary results in better performance. }
    \label{fig:app_perf_expert_level}

\end{figure*}

We first present the effect of different expert quality on the two types of feedback discussed in Sections~\ref{sec:action_recommendation} and~\ref{sec:reward_penalty}. Note that we can compute the entropy of policy $\pi$ for the PPO, PPO-LSTM, Reinforce, Actor-Critic and LinearUCB and Bootstrapped Thompson sampling. We now present the results associated with different expert levels in for the four environments discussed in Section~\ref{sec:experiments}. Figure~\ref{fig:app_perf_expert_level} shows the variation of different expert qualities for different range of learners. The bar plot in orange shows the model performance when reward manipulation is used as a feedback from the human expert and the bar plot in blue shows the model performance when action recommendation as a feedback from human feedback. Notably, higher expert quality does not universally translate to better performance—a counterintuitive relationship whose subtleties we analyze in subsequent sections. Our analysis shows that for different expert levels the effectiveness of incorporating human feedback depends on the learner. Comparison of expert levels with model performance for other learners are shown in Appendix~\ref{sec:app_expert_quality_perf}.

\begin{figure*}[H]
    
    \centering
    \begin{subfigure}[b]{0.32\textwidth}
    \hspace{1.5cm}\setlength{\fboxsep}{1pt}\colorbox{lightgray!30}{\textbf{Bibtex}}
        \includegraphics[width=\linewidth]{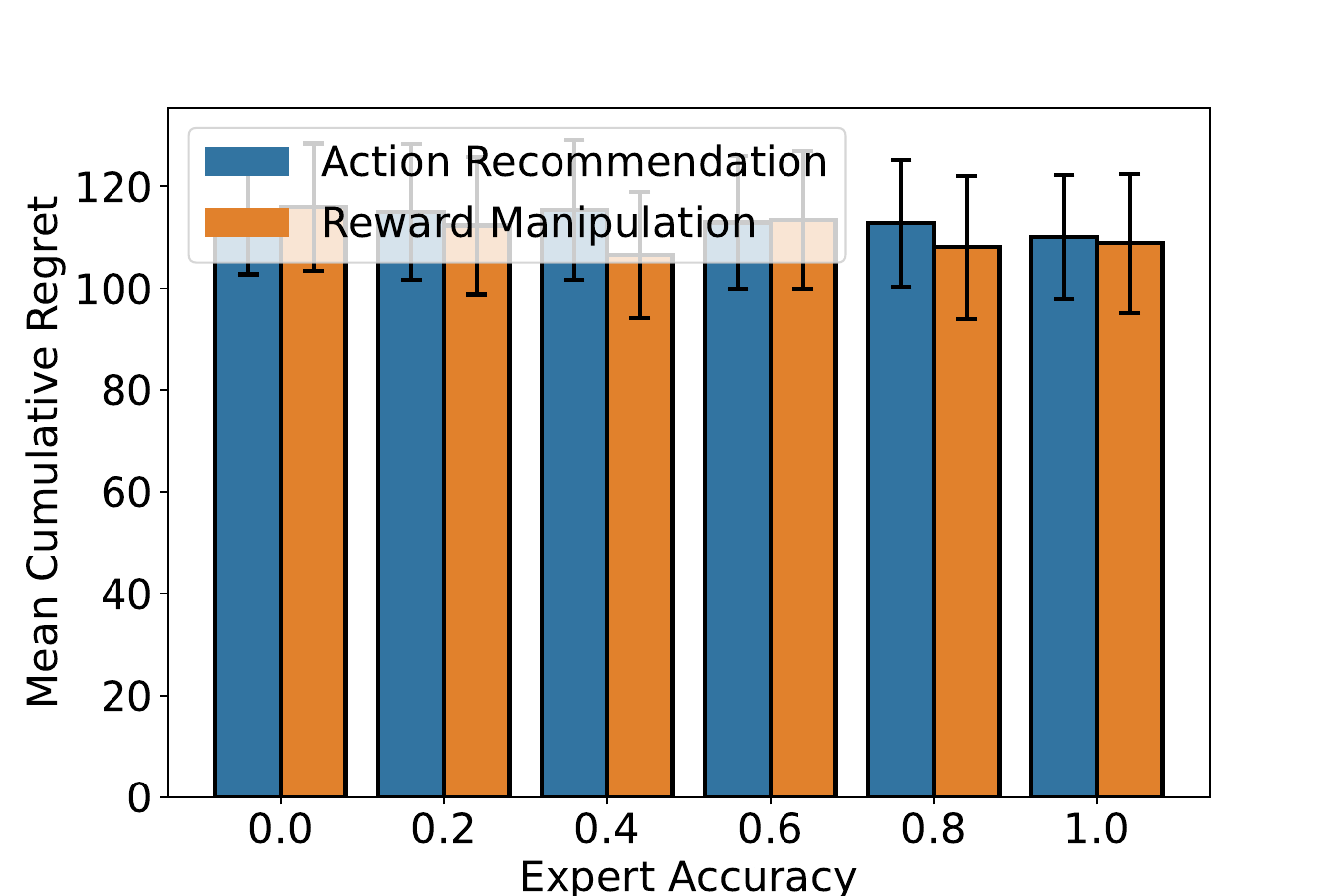}
        
    \end{subfigure}
    \hfill
    \begin{subfigure}[b]{\textwidth}
        \hspace{1.5cm}\setlength{\fboxsep}{1pt}\colorbox{lightgray!30}{\textbf{Delicious}}
        \includegraphics[width=\linewidth]{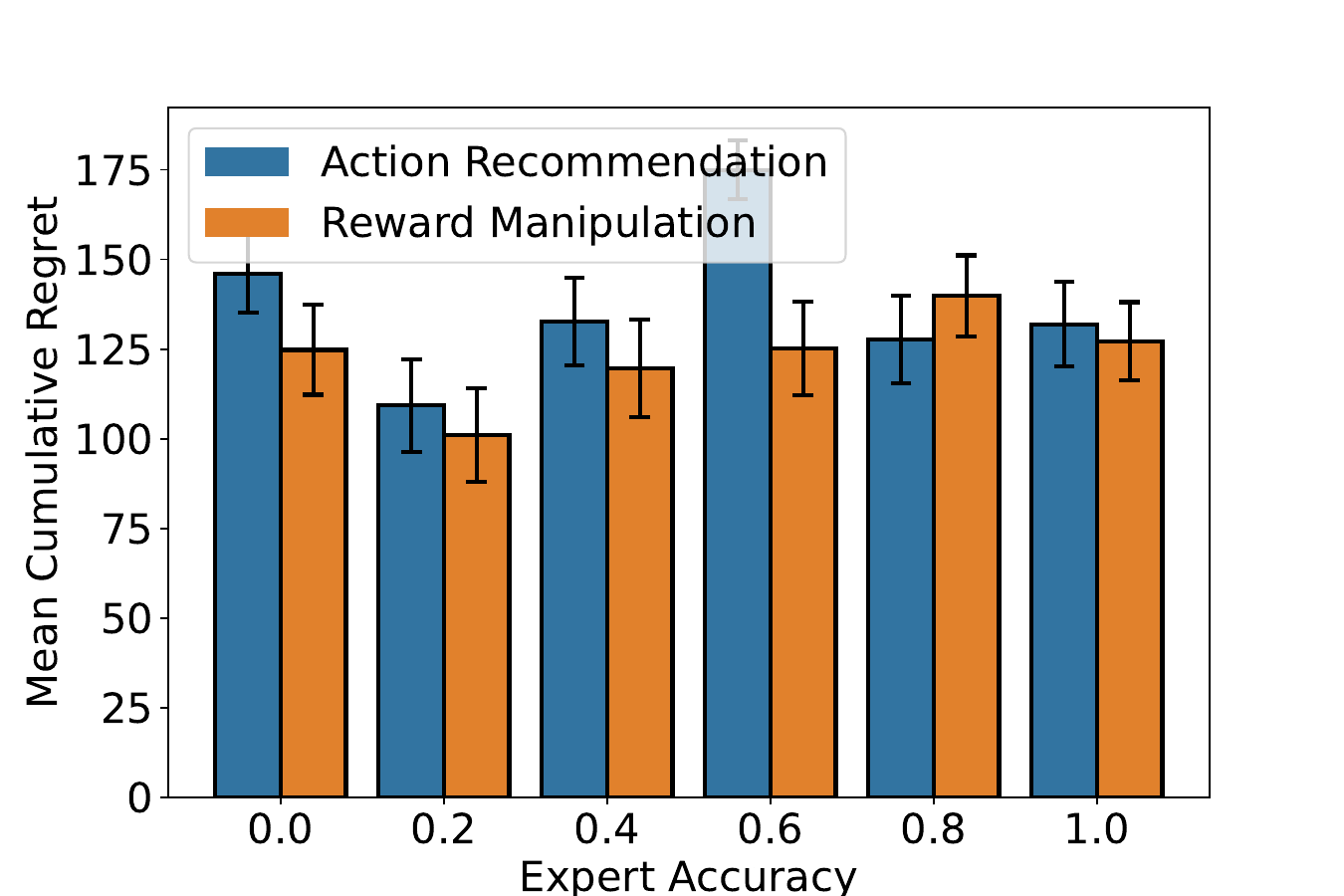}
        
    \end{subfigure}
    \hfill
    \begin{subfigure}[b]{\textwidth}
        \hspace{1.5cm}\setlength{\fboxsep}{1pt}\colorbox{lightgray!30}{\textbf{Media\_Mill}}
        \includegraphics[width=\linewidth]{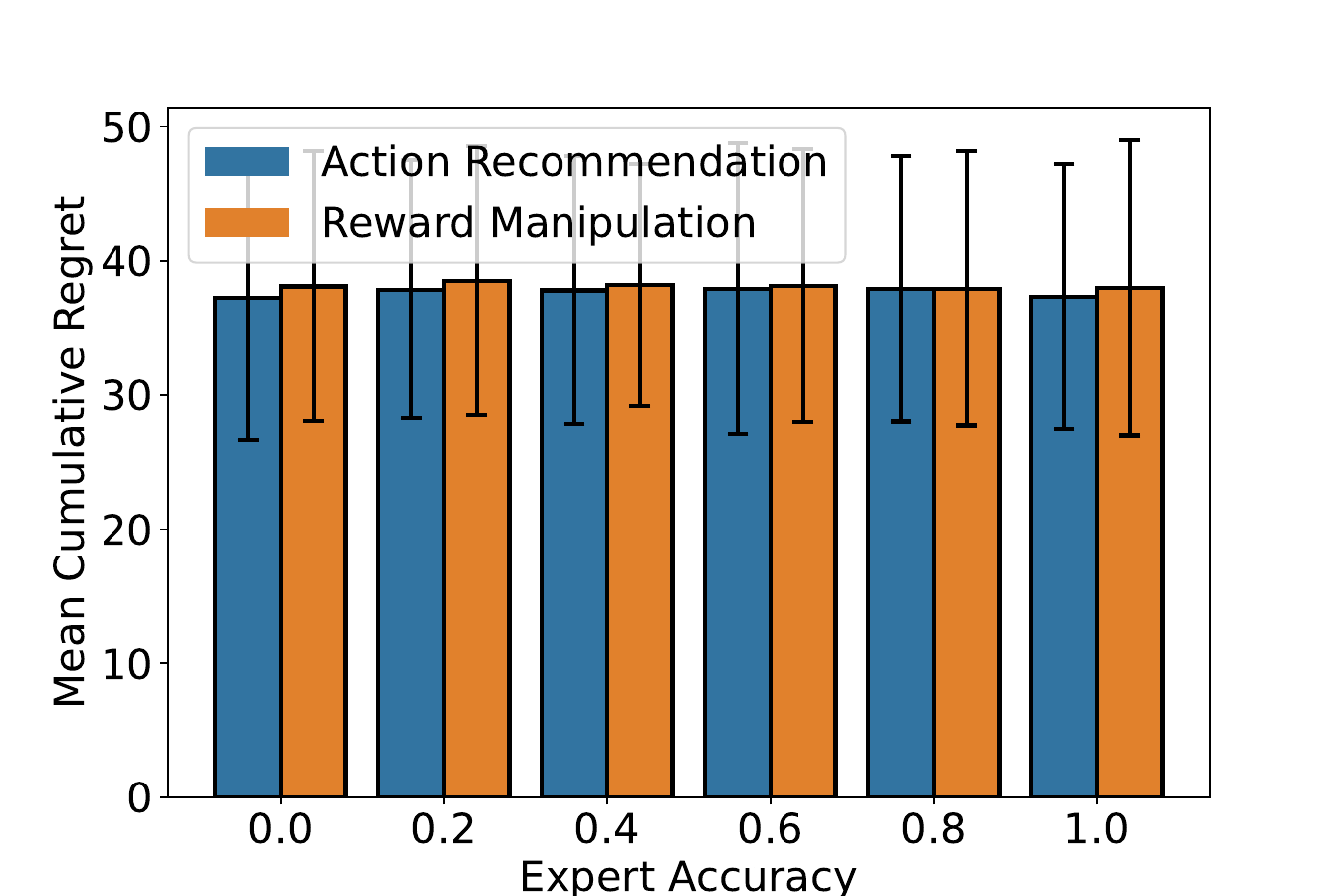}
        
    \end{subfigure}
    \hfill
        
   \text{PPO}\\

    \begin{subfigure}[b]{0.32\textwidth}
        \includegraphics[width=\linewidth]{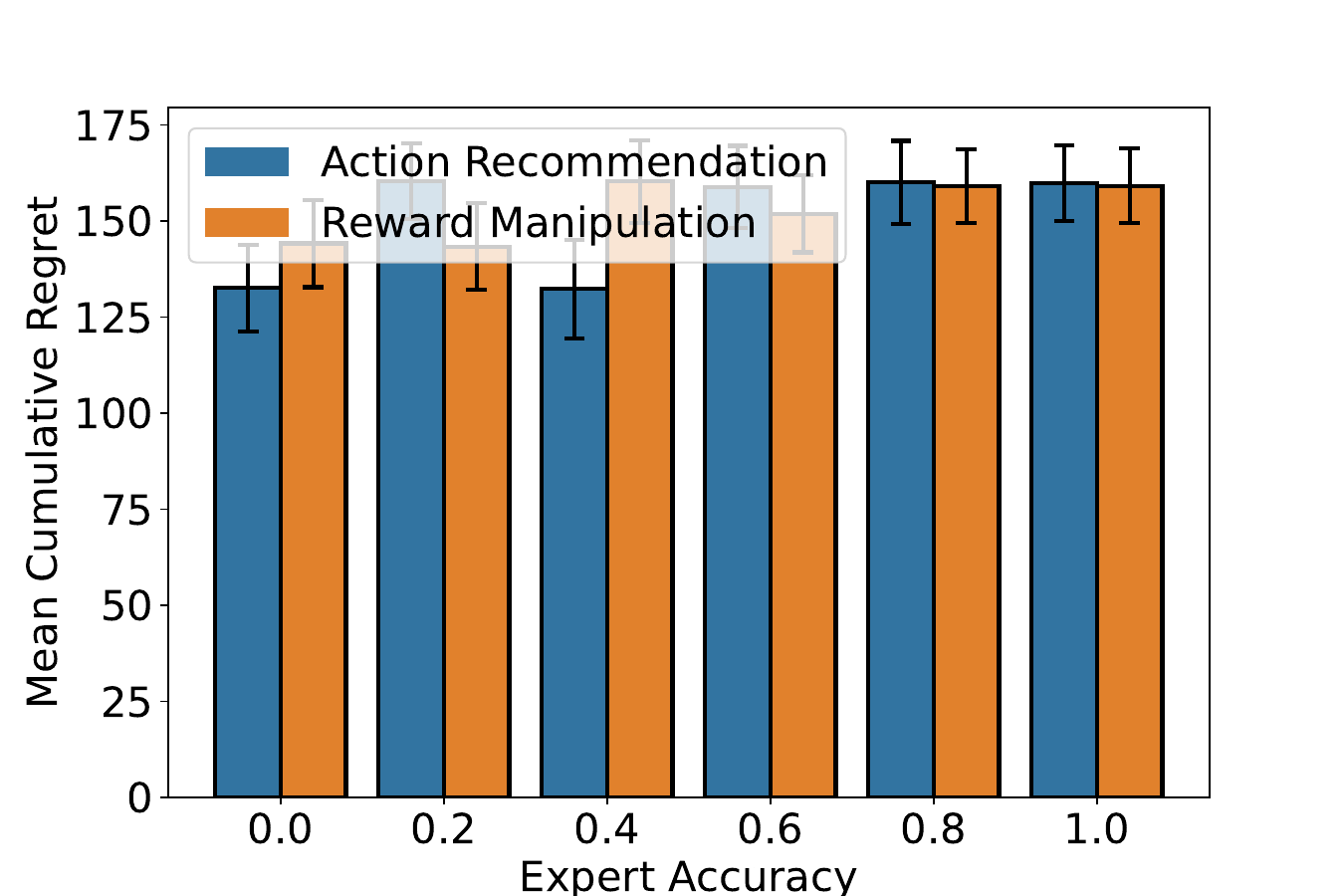}
        
    \end{subfigure}
    \hfill
    \begin{subfigure}[b]{0.32\textwidth}
        \includegraphics[width=\linewidth]{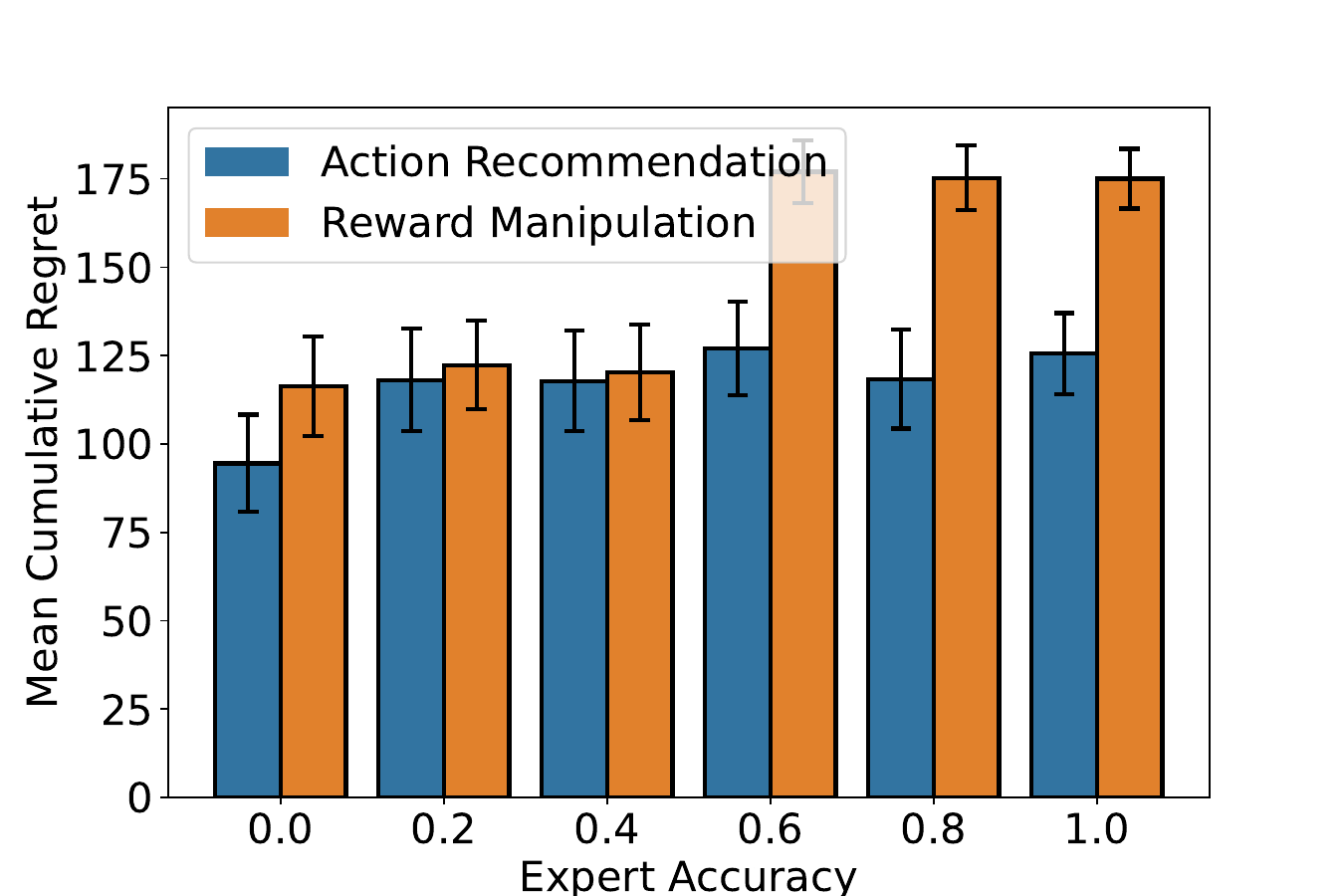}
    \end{subfigure}
    \hfill
    \begin{subfigure}[b]{0.32\textwidth}
        \includegraphics[width=\linewidth]{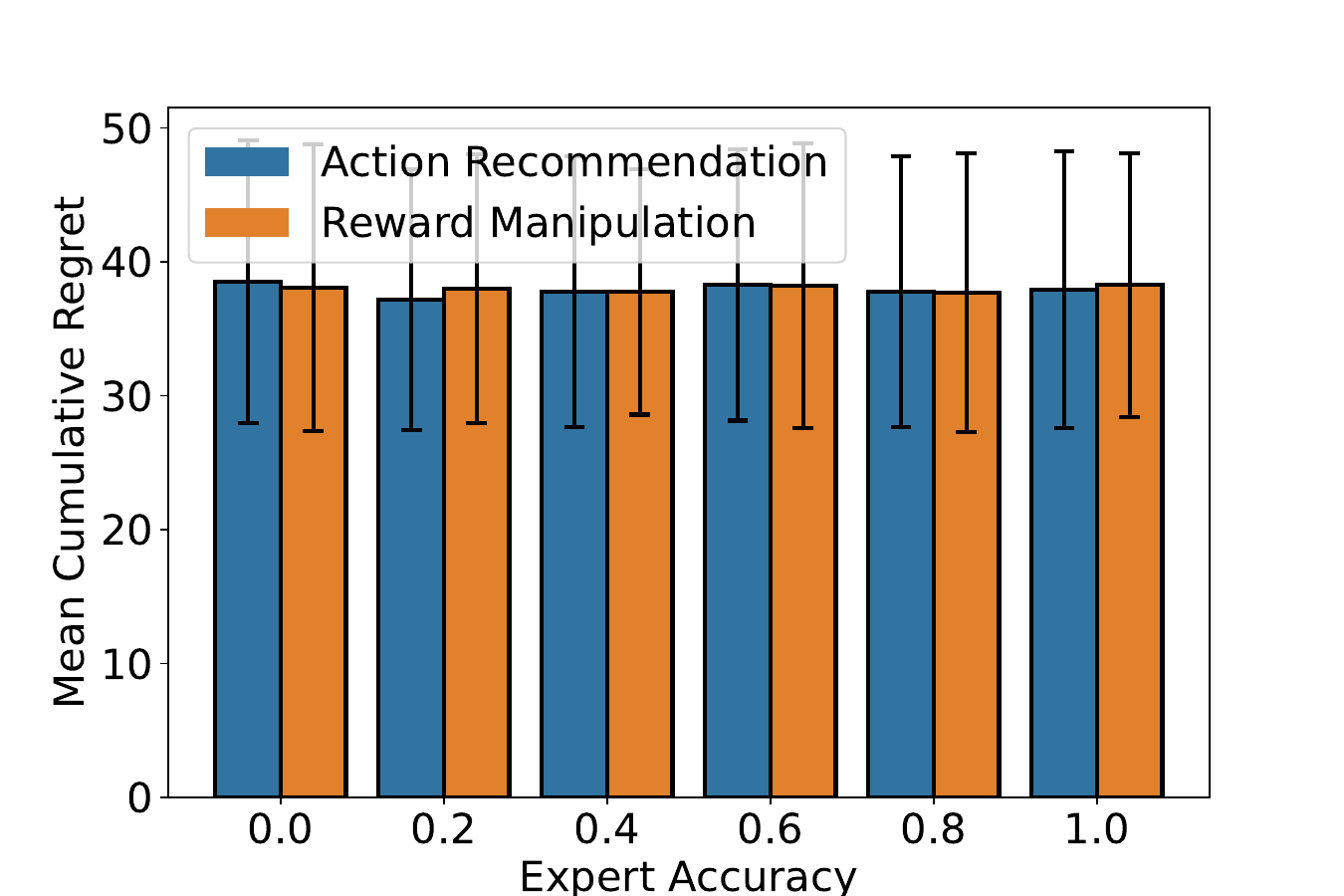}
        
    \end{subfigure}
       
   \text{Reinforce} \\
    
    \begin{subfigure}[b]{0.32\textwidth}
        \includegraphics[width=\linewidth]{figures/expert_accuracy_range/bibtex/reinforce/expert_accuracy_range.pdf}
        
    \end{subfigure}
    \hfill
    \begin{subfigure}[b]{0.32\textwidth}
        \includegraphics[width=\linewidth]{figures/expert_accuracy_range/delicious/reinforce/expert_accuracy_range.pdf}
        
    \end{subfigure}
    \hfill
    \begin{subfigure}[b]{0.32\textwidth}
        \includegraphics[width=\linewidth]{figures/expert_accuracy_range/media_mill/reinforce/expert_accuracy_range.pdf}
        
    \end{subfigure}
       
    \text{Linear UCB}\\

    \caption{Comparison of expert feedback for different learners based on different expert qualities. The results show that mean cumulative reward for different datasets and algorithms vary in a different manner for the two feedback schemes considered. Higher levels of expert does not necessary results in better performance. }
    \label{fig:expert_performance_comp} 
\end{figure*}

\begin{table*}[htbp]
    \centering
    \caption{Performance comparison of algorithms for different quality of expert feedback. The values in bold represent the minimum mean cumulative regret achieved across different levels of expert.} 
    \label{table:expert_range}
    \vspace{+2mm}
    \resizebox{\textwidth}{!}{%
    \begin{tabular}{lllcccc}
        \toprule
        \textbf{Feedback Type} & \textbf{Algorithm Name} & \textbf{Environment Name} & \multicolumn{1}{c}{\textbf{0.2}} & \multicolumn{1}{c}{\textbf{0.4}} & \multicolumn{1}{c}{\textbf{0.6}} & \multicolumn{1}{c}{\textbf{0.8}} \\
        \midrule
        Action Recommendation & PPO & Bibtex & $122.84957\pm8.26216$ & $121.19000\pm 7.14612$ & \bm{$120.88720\pm5.29698$} & $121.78527\pm6.20433$ \\[5pt]
        Reward Manipulation & PPO & Bibtex & $120.99013\pm 7.38448$ & \bm{$115.75083\pm5.24335$} & $121.48417\pm6.96902$ & $121.47040\pm 7.24336$ \\[5pt]
        Action Recommendation & PPO & Delicious & $\mathbf{111.38593\pm2.51805}$ & $134.23813\pm2.29251$ & $175.17900\pm1.52139$ & $136.76483\pm9.70185$ \\[5pt]
        Reward Manipulation & PPO & Delicious & $\mathbf{101.41833\pm2.39055}$ & $119.30050\pm2.20132$ & $127.08153\pm3.21616$ & $139.94187\pm2.04167$ \\[5pt]
        Action Recommendation & PPO-LSTM & Media\_Mill & $38.43170\pm1.76982$ & $38.44773\pm1.77304$ & $38.56507\pm 1.77324$ & \bm{$38.38180\pm1.76374$} \\[5pt]
        Reward Manipulation & PPO-LSTM & Media\_Mill & $38.61077\pm1.82181$ & $38.59887\pm1.80596$ & \bm{$38.40867\pm1.78106$} & $38.59343\pm1.76473$ \\[5pt]
        
        Action Recommendation & Bootstrapped-TS & Bibtex & $196.99950\pm0.53562$ & $196.99950+-0.53562$ & $196.99950\pm0.53562$ & $\mathbf{196.97300\pm0.54129}$ \\[5pt]
        Reward Manipulation & Bootstrapped-TS & Bibtex & $197.06000\pm0.45578$ & $\mathbf{197.06000\pm0.45578}$ & $197.06000\pm0.45578$ & $197.06000\pm0.45578$ \\[5pt]  
        \bottomrule
    \end{tabular}%
    }
\end{table*}

\begin{figure*}[!htb]
    \centering
    \begin{minipage}[b]{0.31\linewidth}
        \centering
        \includegraphics[width=\linewidth]{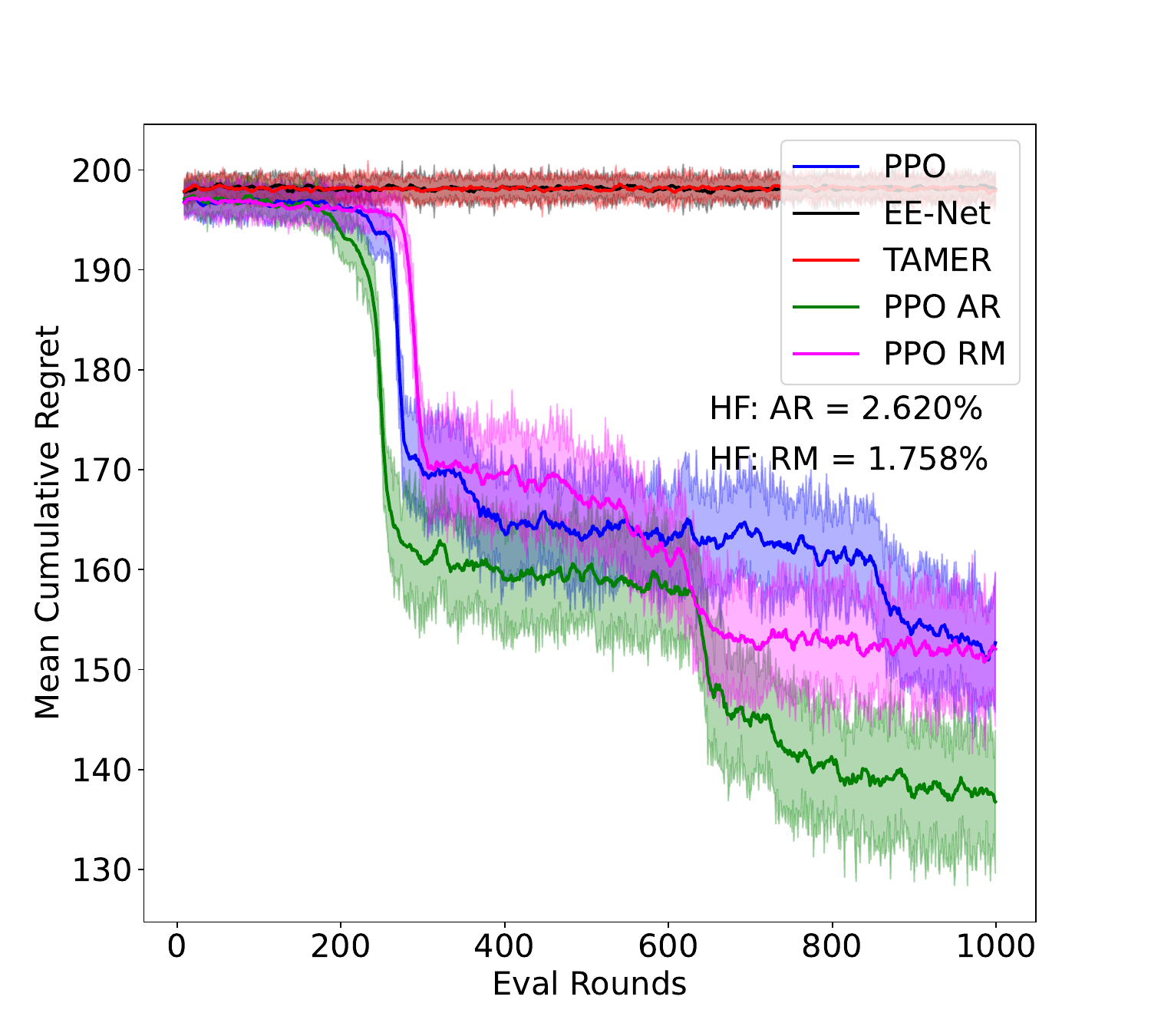}
        \captionsetup{labelformat=empty}
        \text{Bibtex}
    \end{minipage}
    \hspace{0.02\textwidth}
    \begin{minipage}[b]{0.31\linewidth}
        \centering
        \includegraphics[width=\linewidth]{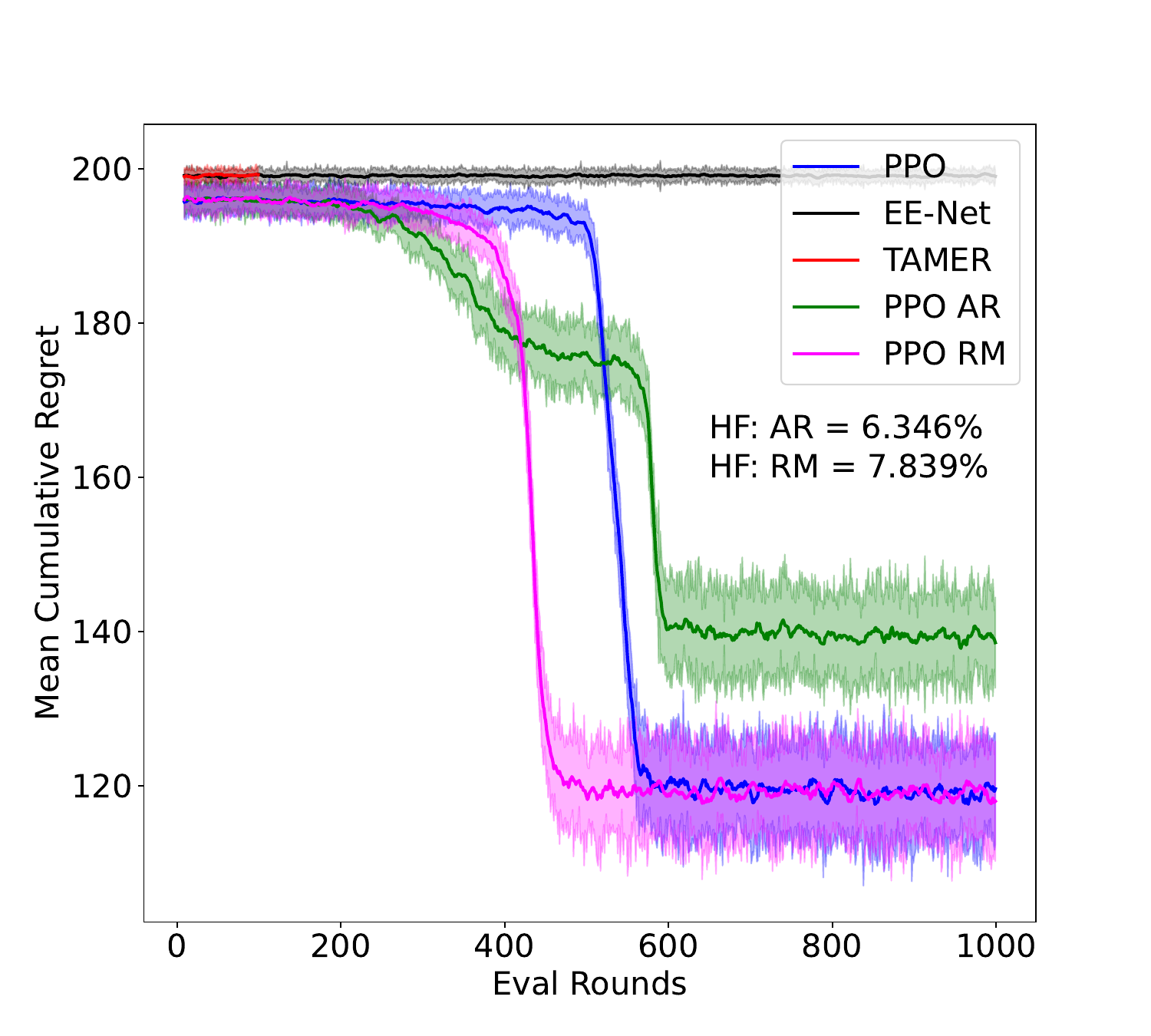}
        \captionsetup{labelformat=empty}
        \text{Delicious}
    \end{minipage}
    \hspace{0.02\textwidth}
    \begin{minipage}[b]{0.31\linewidth}
        \centering
        \includegraphics[width=\textwidth]{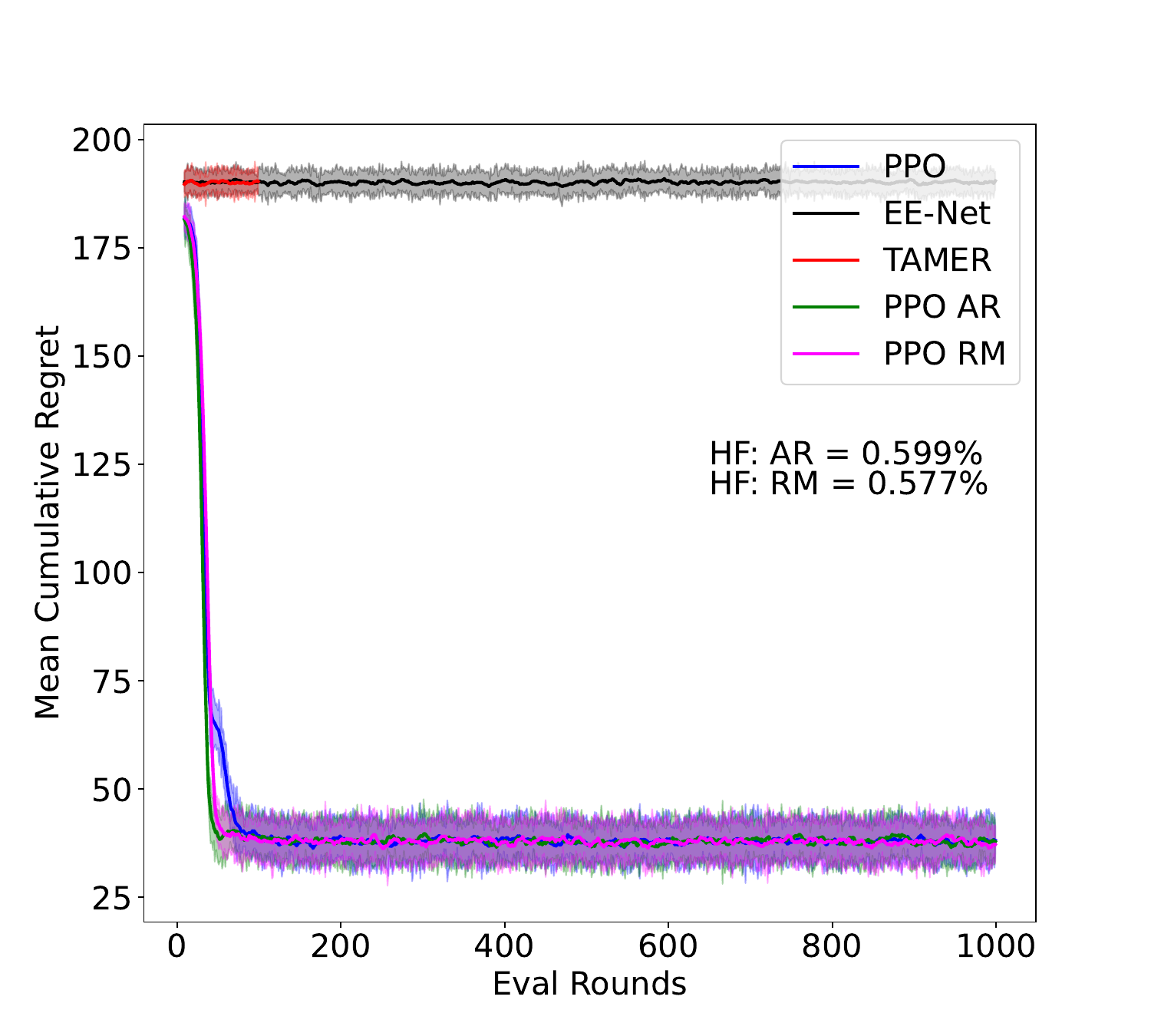}
        \captionsetup{labelformat=empty}
        \text{Media Mill}
    \end{minipage}
    \setcounter{figure}{2}
    \caption{Performance comparison of baselines and the proposed schemes. The figures show that using entropy based feedback leads to lower mean cumulative regret. The solid line represents the mean cumulative regret and the shaded region represents the $\pm$ 1 standard deviation across the mean.}
    \label{fig:baseline_comp}
\end{figure*}

\subsection{Incorporating entropy based feedback achieves higher performance compared to baselines}
We optimize the model performance across various expert levels and compare these results with baseline models, including EE-Net. Figure~\ref{fig:baseline_comp} presents the mean cumulative regret for the optimized expert level (as obtained from Table~\ref{table:expert_range}), highlighting the significant performance gains achieved by incorporating entropy-based feedback over the baselines.

Our analysis, conducted across all datasets, demonstrates that integrating entropy-based feedback—specifically Action Recommendation (AR) and Reward Modification (RM)—consistently outperforms both TAMER and EE-Net. Moreover, we observe that the proportion of steps during which the algorithm seeks human expert feedback varies across datasets. Importantly, the results reveal two key findings: 

Firstly, learners benefit substantially from entropy-based feedback compared to when no such feedback is provided. This improvement underscores the effectiveness of entropy thresholds in selectively involving human experts, thereby guiding the learning process. In fact, even with a modest number of queries to the human expert (less than 30\% of the total training steps), entropy-based feedback drives superior performance over the baseline models. Secondly, the final performance of the learners is not necessarily a monotonous function of the quality of the human feedback, as shown in Figure~\ref{fig:app_perf_expert_level}. 

Interestingly, the performance of AR and RM varies between datasets. For example, on the Bibtex dataset, AR performs worse compared to RM, while on the Delicious dataset, AR demonstrates the best performance among the three. This difference arises due to how penalties affect exploration: Bibtex, with fewer actions, benefits less from AR's action-space limitation, whereas Delicious, with many possible actions, sees AR accelerating convergence by narrowing down the action space early in the learning process. As a result, AR's advantage becomes more apparent in environments where an overwhelming number of actions could otherwise slow down the learner’s progress.

Further details regarding the proportion of expert queries for different levels of expert quality are provided in Appendix~\ref{sec:app_var_exp_query_percentage}.


\subsection{Effect of entropy threshold and expert accuracy on model performance}
Figure~\ref{fig:entr_thresh_expert_acc_ar_rp} (Appendix) presents bubble plots comparing model performance at different expert levels and entropy threshold values for both AR and RM feedback types. The size and color of each bubble represent the mean cumulative reward for the corresponding learner. 

We begin by analyzing the results for AR feedback. Generally, we observe that at higher entropy threshold values, the model's performance remains relatively stable across different expert levels. This behavior is expected, as higher entropy thresholds result in fewer queries to the human expert, reducing the impact of expert quality on performance. 

However, at lower entropy thresholds, an interesting pattern emerges: increasing expert quality can actually lead to a decrease in model performance. This phenomenon relates to the exploration-exploitation trade-off. At high expert levels, the expert consistently provides accurate recommendations, and since the model is designed to always accept these recommendations in the AR setting, the result is pure exploitation. Conversely, at lower expert levels, where recommendations are more random, the model is encouraged to explore a broader set of actions, which can ultimately yield higher cumulative rewards or lower cumulative regrets.

A similar pattern is observed with RM feedback. At higher entropy thresholds, the differences in performance between varying expert levels are minimal, as fewer queries are made to the expert. At lower entropy thresholds, however, we again see a decline in performance as expert quality increases. 

Further bubble plots illustrating these trends for other learners, under both AR and RM feedback, can be found in Appendix~\ref{sec:app_entr_threshold_expert_acc_perf}.

\captionsetup{font=footnotesize}

\subsection{Observed differences between feedback types}
Figure~\ref{fig:baseline_comp} illustrates how the two forms of feedback, AR and RM, interact differently with the underlying algorithms and datasets. The choice of feedback type should therefore depend on the specific application.

Our results generally indicate that at higher expert levels, AR tends to be more effective than RM. This is likely because AR directly influences the actions taken by the contextual bandit (CB), interfering less with its reward-based learning process. At low expert levels, however, AR can become disruptive, leading to poor exploration by prematurely narrowing the action space. In contrast, at high expert levels, AR provides clearer guidance for the bandit’s exploration, optimizing action selection while leaving the reward structure relatively intact.

Ultimately, this suggests that AR is particularly advantageous when expert quality is high, as it can effectively guide exploration without destabilizing the learning process.




\section{Conclusion}

In conclusion, this work introduces an effective entropy-based framework for incorporating human feedback into contextual bandits. By utilizing model entropy to trigger feedback solicitation, we significantly reduce the reliance on continuous human intervention, thus making the system more efficient and scalable. Our experiments show that even with low-quality human feedback, substantial performance gains can be achieved, underscoring the potential of entropy-based feedback mechanisms in various real-world applications. By dynamically selecting feedback opportunities based on model uncertainty, our approach provides a generalizable strategy for optimizing human intervention in learning systems, improving efficiency while minimizing unnecessary queries. Future work can explore further applications of entropy-based feedback in these domains, fostering more adaptive and intelligent human-in-the-loop learning frameworks.
 

\section{Impact statement}
This paper presents work whose goal is to advance the field of Machine Learning. There are many potential societal consequences of our work, none which we feel must be specifically highlighted here.

\nocite{langley00}

\bibliography{icml2025_conference}
\bibliographystyle{icml2025}

\appendix
\textbf{Other related areas}
Our work builds on several important research areas, including counterfactual reasoning, imitation learning, preference optimization, and entropy-based active learning. We draw inspiration from Tang and Wiens \cite{tang2023counterfactual}, whose counterfactual-augmented importance sampling informs our feedback framework, and extend DAGGER \cite{ross2011reduction} by dynamically incorporating expert feedback instead of using fixed imitation. We also acknowledge parallels with Active Preference Optimization (APO) \cite{das2024active}, adapting trajectory-level preference feedback to reward manipulation in more complex settings. Additionally, we connect with entropy-driven methods like BALD \cite{houlsby2011bayesian} and IDS \cite{russo2014learning}, adapting their principles for contextual bandit problems to balance information gain and decision-making efficiency in sequential exploration. These connections highlight how our approach advances real-time feedback integration and decision optimization.

\section{Effect of feedback qualities on different learners}
\label{sec:app_expert_quality_perf}

This section details the impact of feedback levels on different learners.
\begin{figure}[!thb]
    \centering
    
    \begin{subfigure}[b]{0.3\columnwidth}
        \centering
        \setlength{\fboxsep}{1pt}\colorbox{lightgray!30}{\textbf{Bibtex}}
        \includegraphics[width=\textwidth]{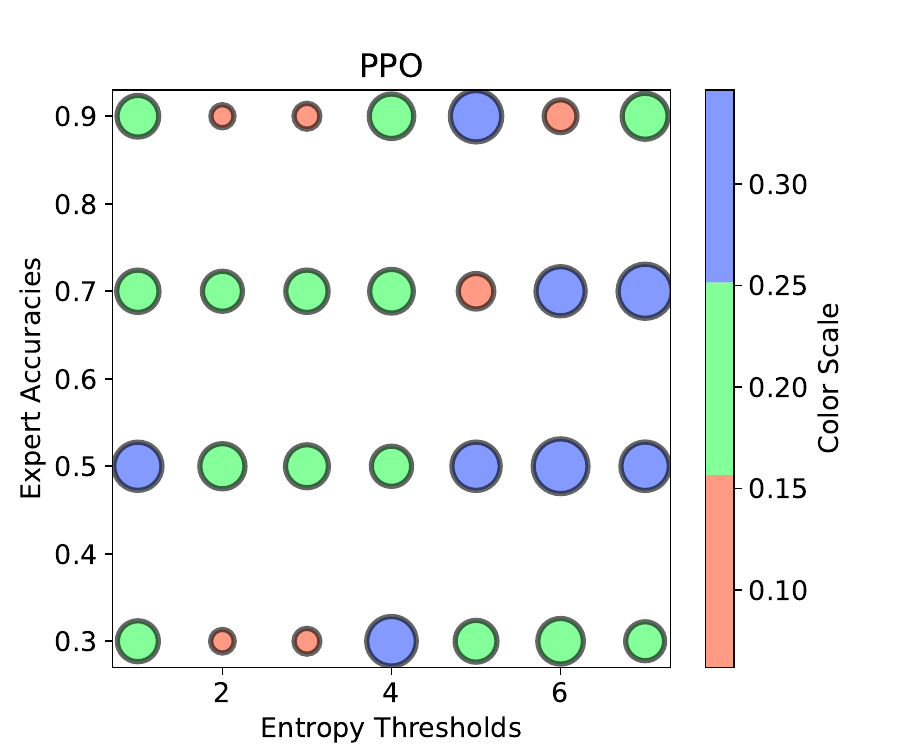}
     
    \end{subfigure}
    \hfill
    \begin{subfigure}[b]{0.3\columnwidth}
        \centering
        \setlength{\fboxsep}{1pt}\colorbox{lightgray!30}{\textbf{Delicious}}
        \includegraphics[width=\textwidth]{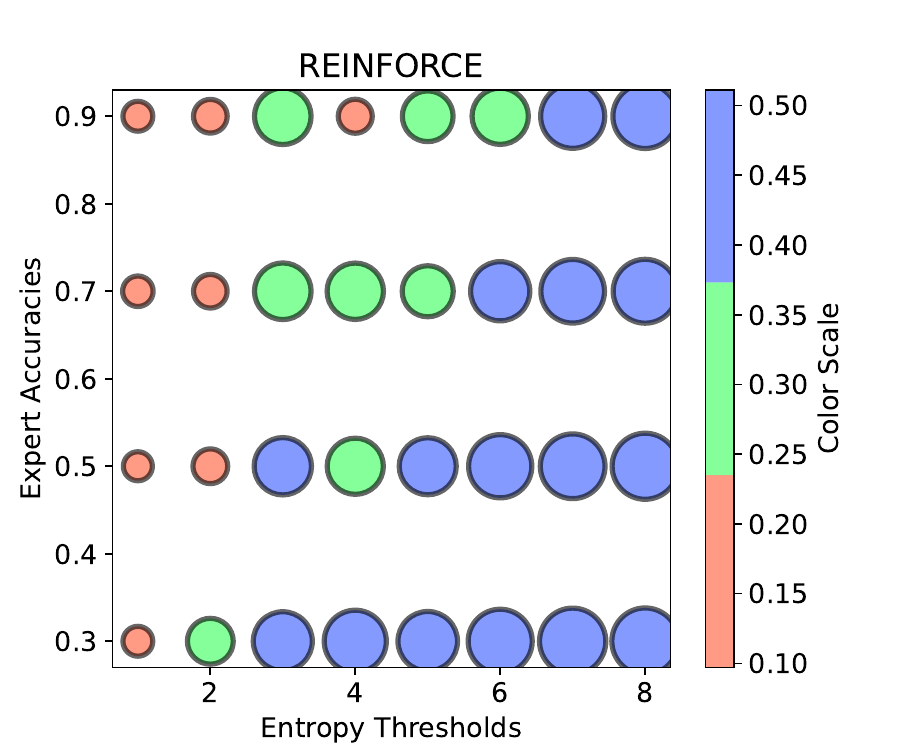}
       
    \end{subfigure}
    \hfill
    \begin{subfigure}[b]{0.3\columnwidth}
        \centering
        \setlength{\fboxsep}{1pt}\colorbox{lightgray!30}{\textbf{MediaMill}}
        \includegraphics[width=\textwidth]{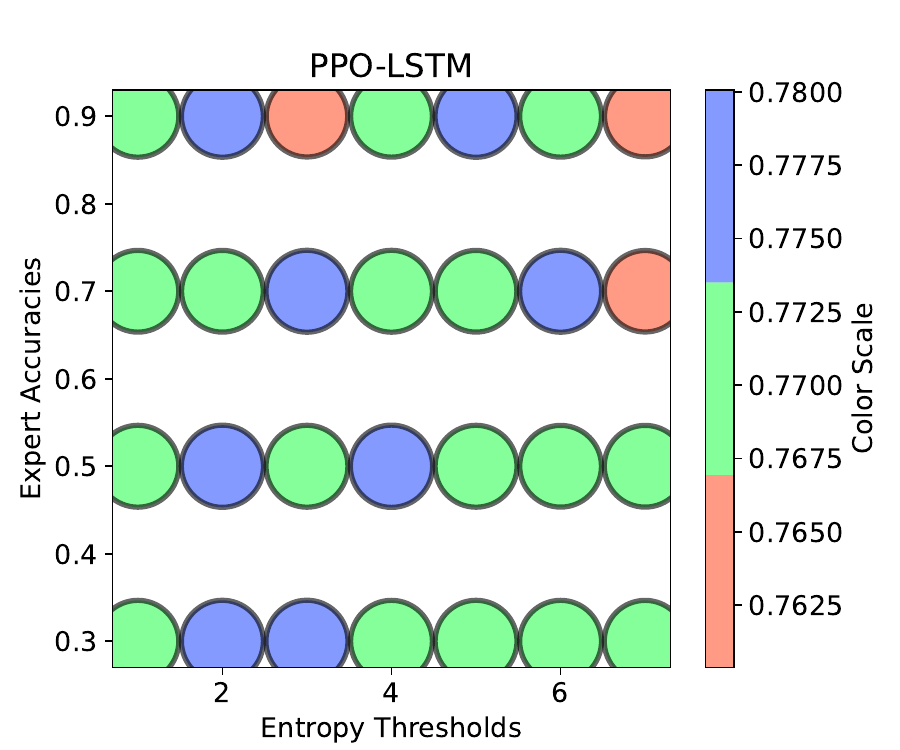}
       
    \end{subfigure}
    \setlength{\fboxsep}{1pt}\colorbox{lightgray!30}{\textbf{Action Recommendation}}
    \vspace{1em} 
    
    \begin{subfigure}[b]{0.3\columnwidth}
        \centering
        \includegraphics[width=\textwidth]{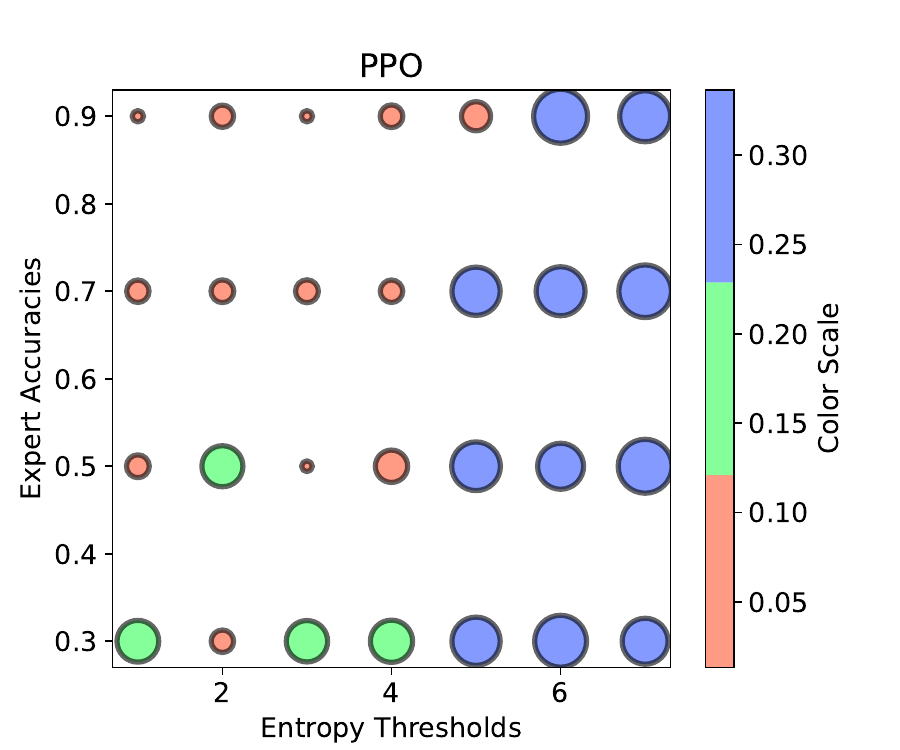}
    \end{subfigure}
    \hfill
    \begin{subfigure}[b]{0.3\columnwidth}
        \centering
        \includegraphics[width=\textwidth]{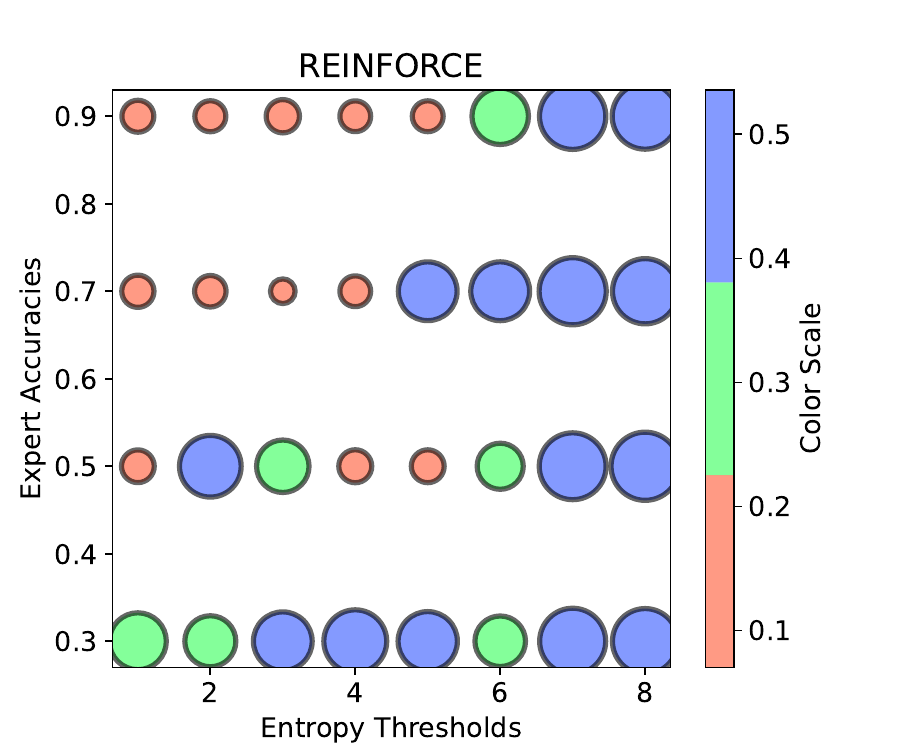}
    \end{subfigure}
    \hfill
    \begin{subfigure}[b]{0.3\columnwidth}
        \centering
        \includegraphics[width=\textwidth]{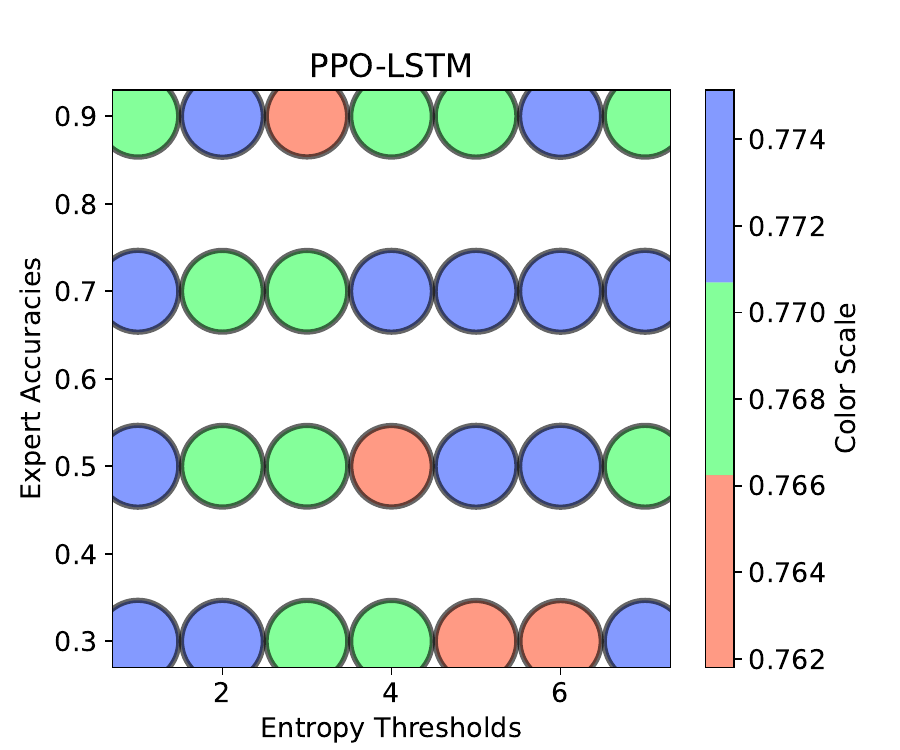}
    \end{subfigure}
    \setlength{\fboxsep}{1pt}\colorbox{lightgray!30}{\textbf{Reward Manipulation}}
    
    \caption{Comparison of model performance for different values of entropy and expert accuracies for feedback: Action Recommendation and Reward Manipulation. The size and color of each bubble in the bubble plots represent the magnitude of the mean cumulative reward.}
    \label{fig:entr_thresh_expert_acc_ar_rp}
\end{figure}

\section{Effect of entropy threshold and expert levels on model performance}
\label{sec:app_entr_threshold_expert_acc_perf}

This section studies the impact of the entropy threshold on performance.
\captionsetup{font=footnotesize}
\begin{figure}[H]
     \caption{Comparison of model performance for different values of entropy and expert accuracies for feedback: Action Recommendation. The size and color of each bubble in the bubble plots represent the magnitude of the mean cumulative reward.}
    \label{fig:app_entr_thresh_expert_acc_action_recommendation}
    \centering
    \begin{subfigure}[b]{0.32\columnwidth}
        \centering
        \setlength{\fboxsep}{1pt}\colorbox{lightgray!30}{\textbf{Bibtex}}
        \includegraphics[width=\columnwidth]{figures/plot_analysis_entropy_expert/bibtex/action_restriction_accuracy/ppo_entropy_expert.pdf}
    \end{subfigure}
    \hfill
    \begin{subfigure}[b]{0.32\columnwidth}
        \centering
        \setlength{\fboxsep}{1pt}\colorbox{lightgray!30}{\textbf{Delicious}}
        \includegraphics[width=\linewidth]{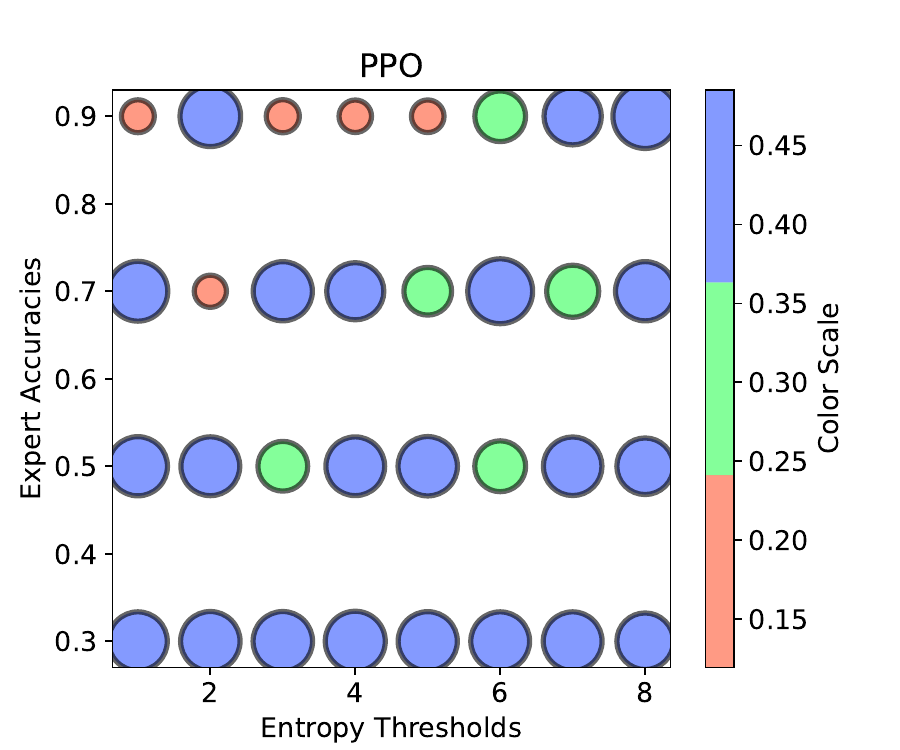}
    \end{subfigure}
    \hfill
    \begin{subfigure}[b]{0.32\columnwidth}
        \centering
        \setlength{\fboxsep}{1pt}\colorbox{lightgray!30}{\textbf{MediaMill}}
        \includegraphics[width=\linewidth]{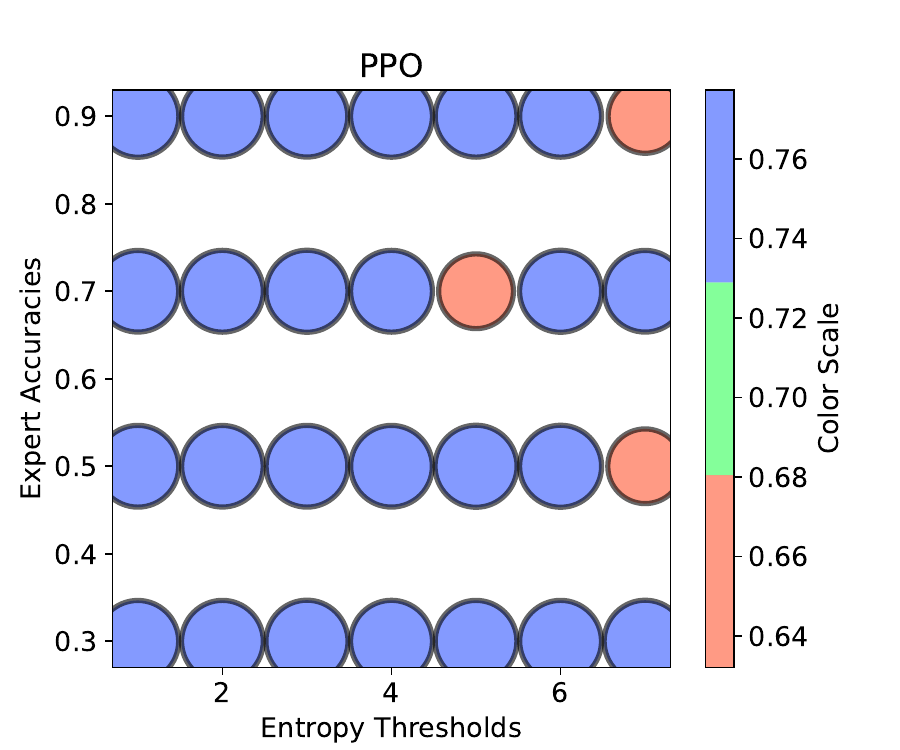}
    \end{subfigure}
    \hfill
    
    \begin{subfigure}[b]{0.32\columnwidth}
        \centering
        \includegraphics[width=\linewidth]{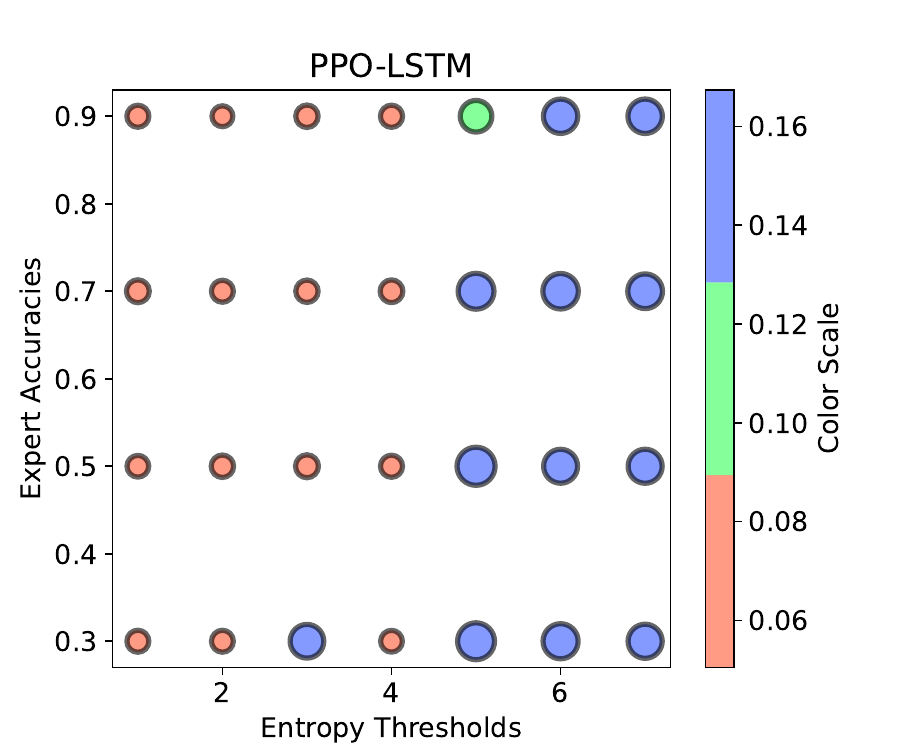}
    \end{subfigure}
    \hfill
    \begin{subfigure}[b]{0.32\columnwidth}
        \centering
        \includegraphics[width=\linewidth]{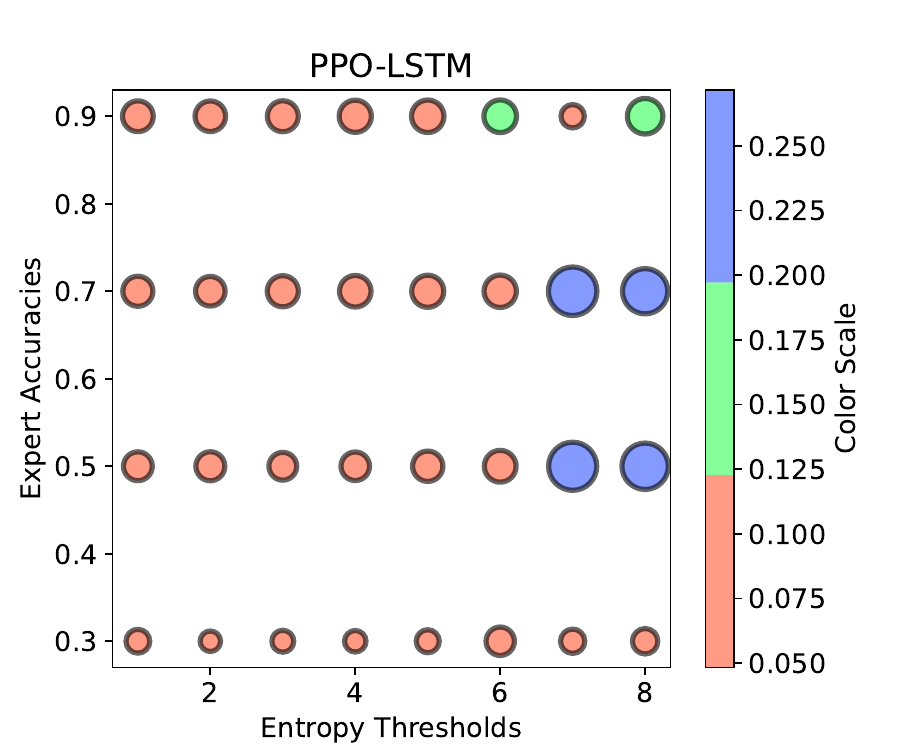}
    \end{subfigure}
    \hfill
    \begin{subfigure}[b]{0.32\columnwidth}
        \centering
        \includegraphics[width=\linewidth]{figures/plot_analysis_entropy_expert/media_mill/action_restriction_accuracy/ppo-lstm_entropy_expert.pdf}
    \end{subfigure}
    \hfill

    \begin{subfigure}[b]{0.32\columnwidth}
        \centering
        \includegraphics[width=\linewidth]{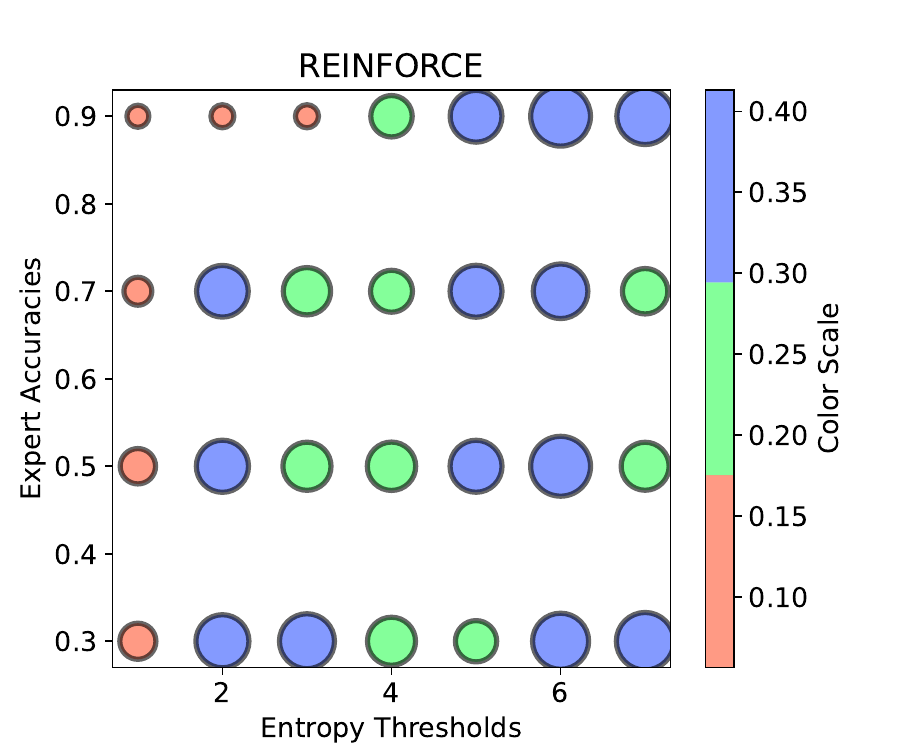}
    \end{subfigure}
    \hfill
    \begin{subfigure}[b]{0.32\columnwidth}
        \centering
        \includegraphics[width=\linewidth]{figures/plot_analysis_entropy_expert/delicious/action_restriction_accuracy/reinforce_entropy_expert.pdf}
    \end{subfigure}
    \hfill
    \begin{subfigure}[b]{0.32\columnwidth}
        \centering
        \includegraphics[width=\linewidth]{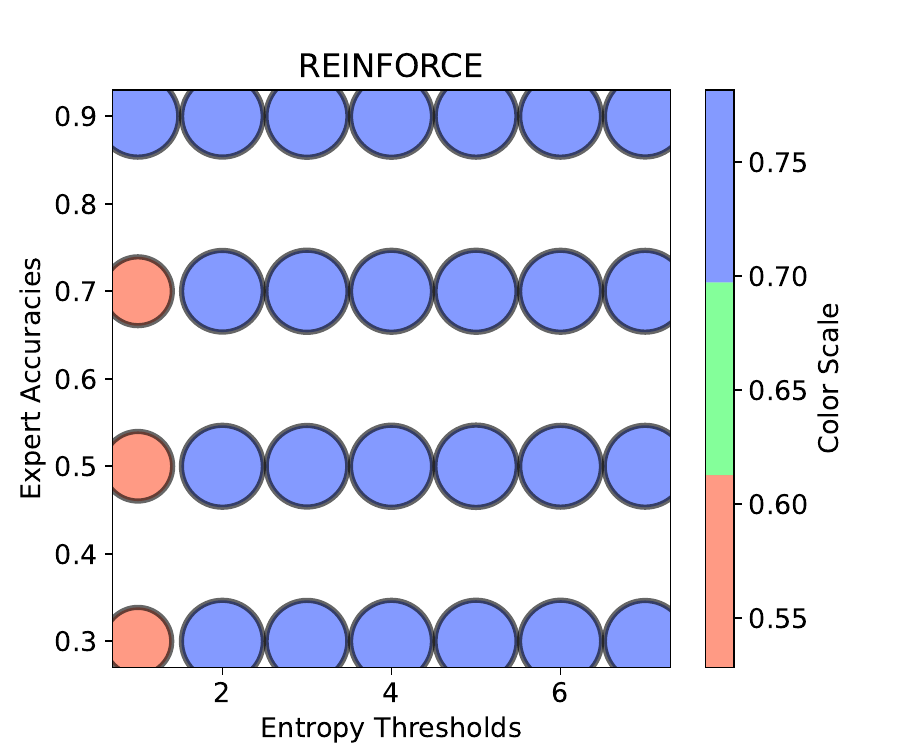}
    \end{subfigure}
    \hfill

    \begin{subfigure}[b]{0.32\columnwidth}
        \centering
        \includegraphics[width=\linewidth]{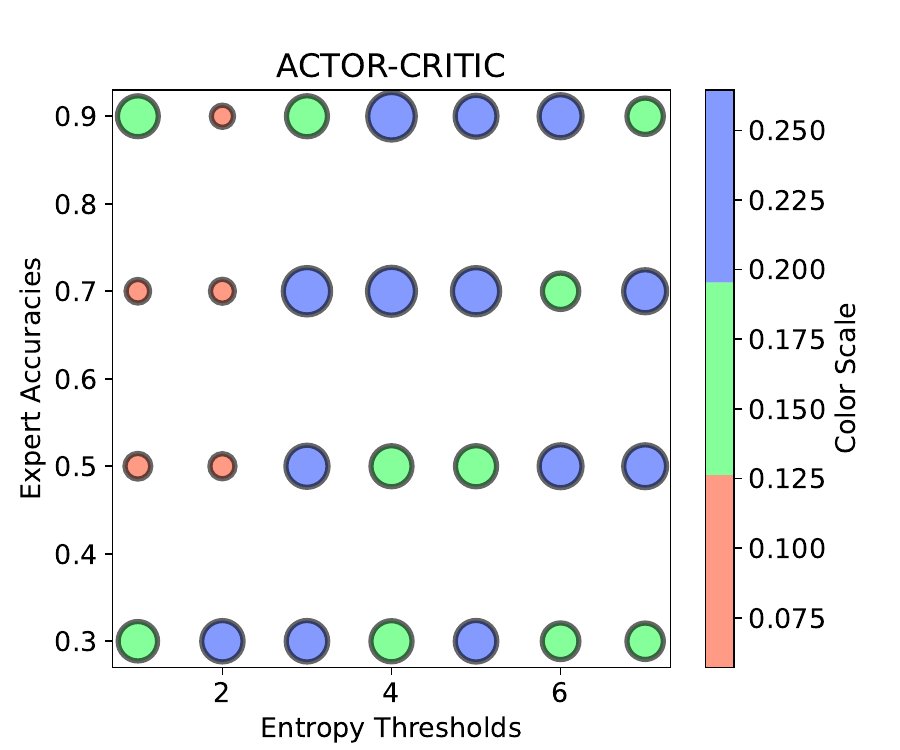}
    \end{subfigure}
    \hfill
    \begin{subfigure}[b]{0.32\columnwidth}
        \centering
        \includegraphics[width=\linewidth]{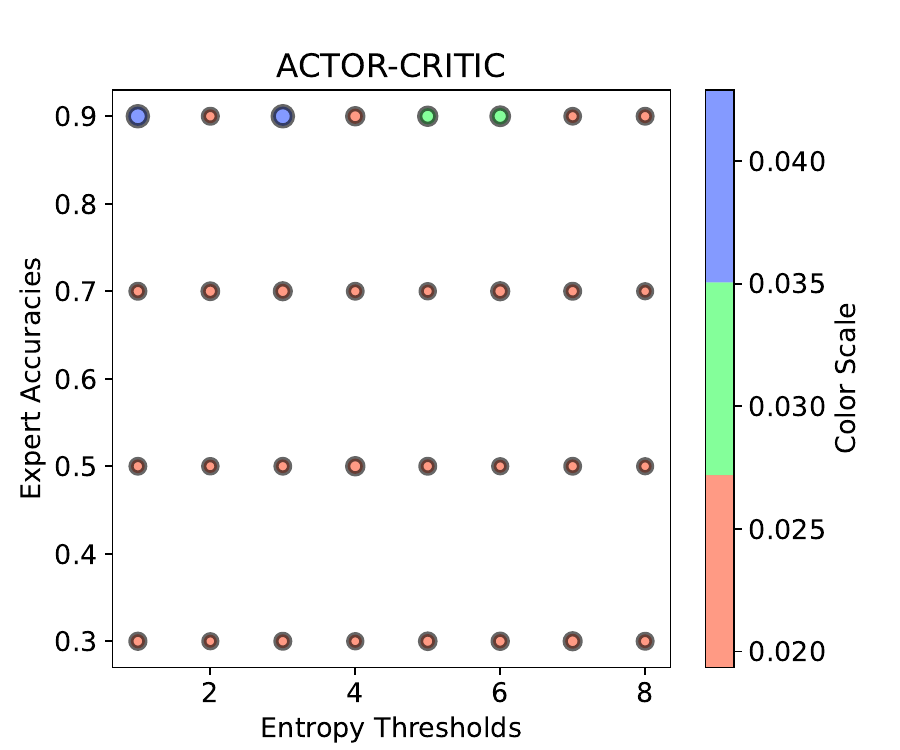}
    \end{subfigure}
    \hfill
    \begin{subfigure}[b]{0.32\columnwidth}
        \centering
        \includegraphics[width=\linewidth]{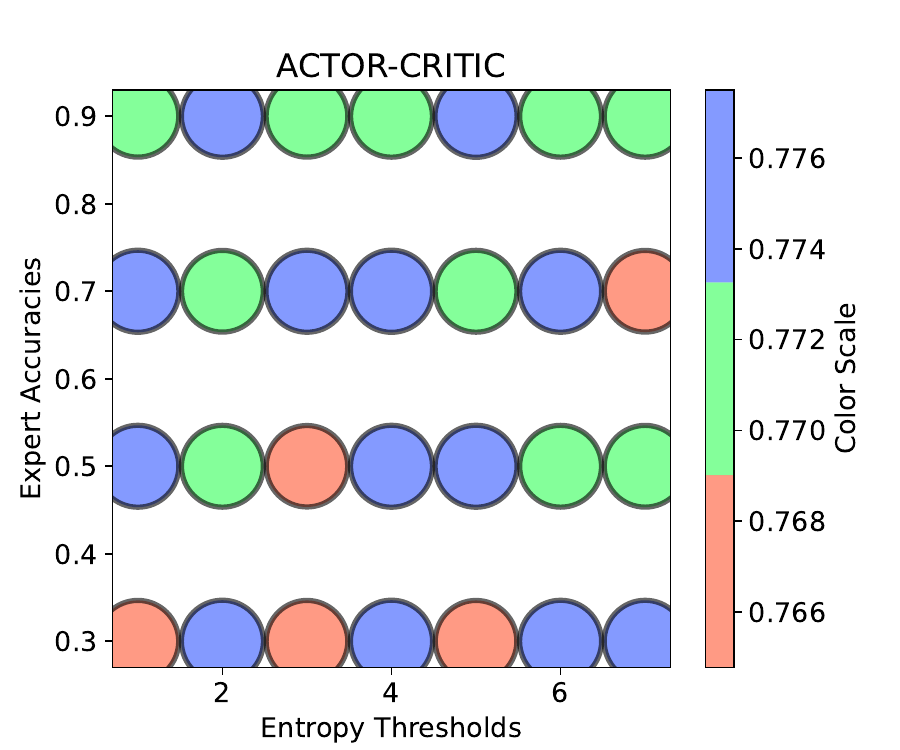}
    \end{subfigure}
    \hfill

     \begin{subfigure}[b]{0.32\columnwidth}
        \centering
        \includegraphics[width=\linewidth]{figures/plot_analysis_entropy_expert/media_mill/action_restriction_accuracy/actor-critic_entropy_expert.pdf}
    \end{subfigure}
    \hfill
     \begin{subfigure}[b]{0.32\columnwidth}
        \centering
        \includegraphics[width=\linewidth]{figures/plot_analysis_entropy_expert/media_mill/action_restriction_accuracy/actor-critic_entropy_expert.pdf}
    \end{subfigure}
    \hfill
     \begin{subfigure}[b]{0.32\columnwidth}
        \centering
        \includegraphics[width=\linewidth]{figures/plot_analysis_entropy_expert/media_mill/action_restriction_accuracy/actor-critic_entropy_expert.pdf}
    \end{subfigure}
    \hfill
    
   
\end{figure}

\captionsetup{font=footnotesize}
\begin{figure}[H]
     \caption{Comparison of model performance for different values of entropy and expert accuracies for feedback: Reward Manipulation. The size and color of each bubble in the bubble plots represent the magnitude of the mean cumulative reward.}
    \label{fig:entr_thresh_expert_acc_reward_manipulation}
    \centering
    \begin{subfigure}[b]{0.32\columnwidth}
        \centering
        \setlength{\fboxsep}{1pt}\colorbox{lightgray!30}{\textbf{Bibtex}}
        \includegraphics[width=\columnwidth]{figures/plot_analysis_entropy_expert/bibtex/reward_penalty_accuracy/ppo_entropy_expert.pdf}
    \end{subfigure}
    \hfill
    \begin{subfigure}[b]{0.32\columnwidth}
        \centering
        \setlength{\fboxsep}{1pt}\colorbox{lightgray!30}{\textbf{Delicious}}
        \includegraphics[width=\linewidth]{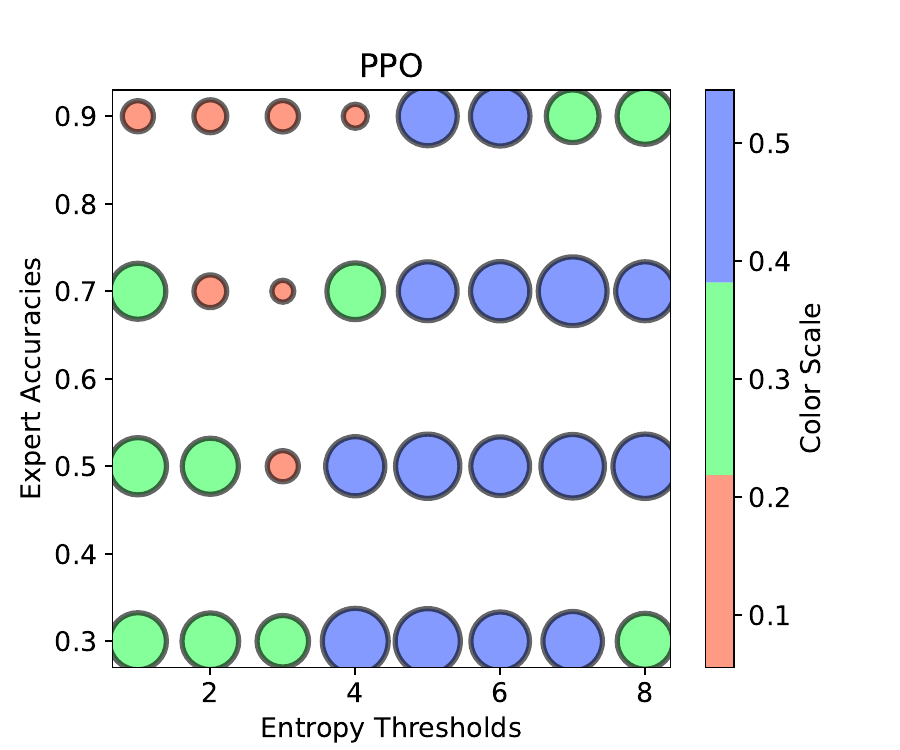}
    \end{subfigure}
    \hfill
    \begin{subfigure}[b]{0.32\columnwidth}
        \centering
        \setlength{\fboxsep}{1pt}\colorbox{lightgray!30}{\textbf{MediaMill}}
        \includegraphics[width=\linewidth]{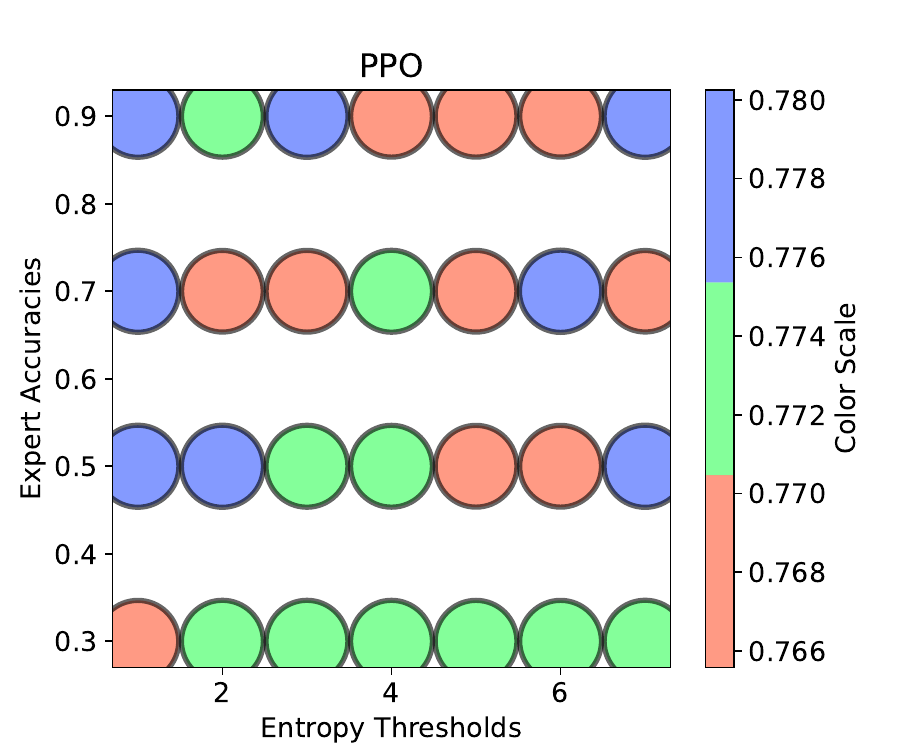}
    \end{subfigure}
    \hfill
    
    \begin{subfigure}[b]{0.32\columnwidth}
        \centering
        \includegraphics[width=\linewidth]{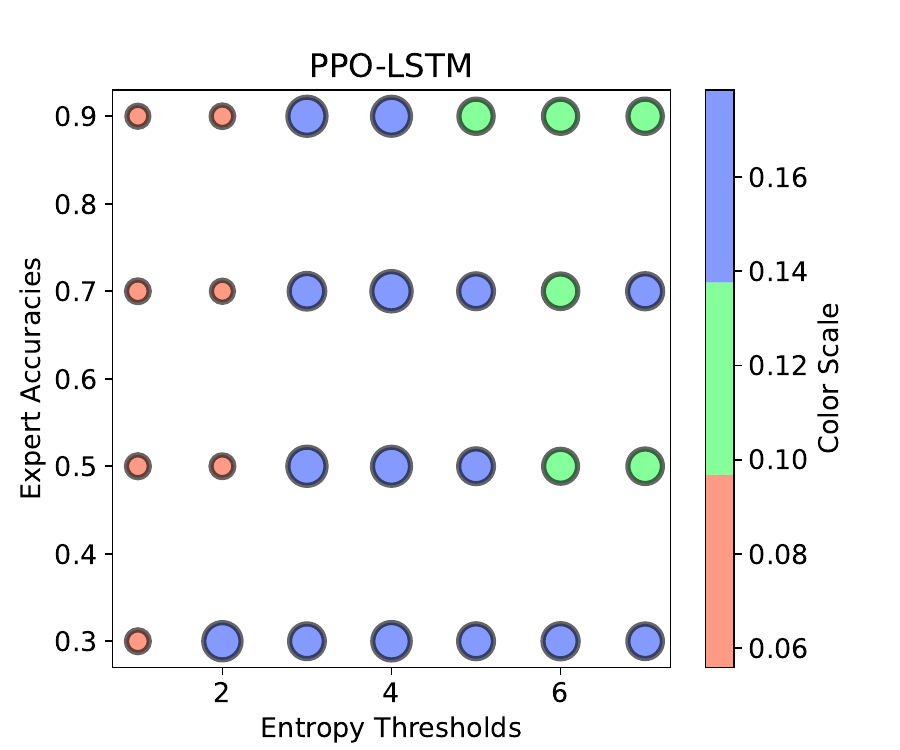}
    \end{subfigure}
    \hfill
    \begin{subfigure}[b]{0.32\columnwidth}
        \centering
        \includegraphics[width=\linewidth]{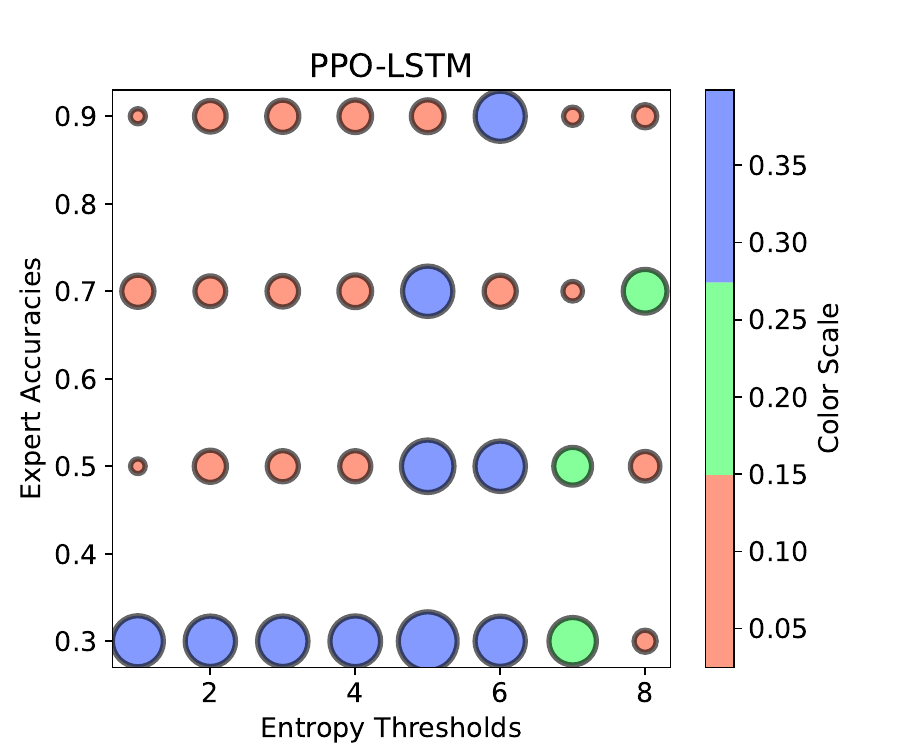}
    \end{subfigure}
    \hfill
    \begin{subfigure}[b]{0.32\columnwidth}
        \centering
        \includegraphics[width=\linewidth]{figures/plot_analysis_entropy_expert/media_mill/reward_penalty_accuracy/ppo-lstm_entropy_expert.pdf}
    \end{subfigure}
    \hfill

    \begin{subfigure}[b]{0.32\columnwidth}
        \centering
        \includegraphics[width=\linewidth]{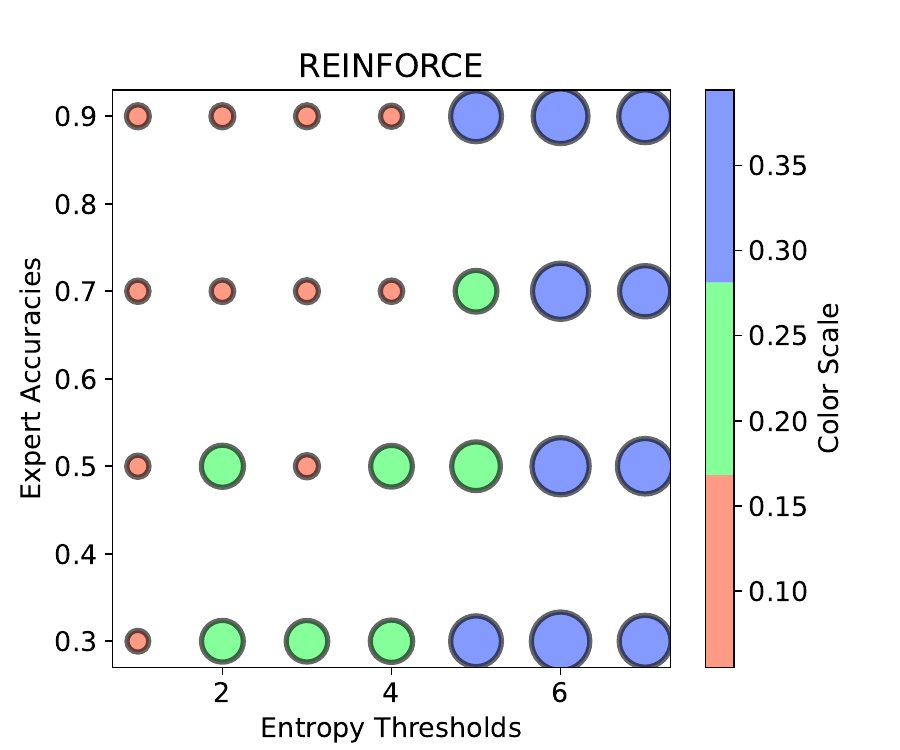}
    \end{subfigure}
    \hfill
    \begin{subfigure}[b]{0.32\columnwidth}
        \centering
        \includegraphics[width=\linewidth]{figures/plot_analysis_entropy_expert/delicious/reward_penalty_accuracy/reinforce_entropy_expert.pdf}
    \end{subfigure}
    \hfill
    \begin{subfigure}[b]{0.32\columnwidth}
        \centering
        \includegraphics[width=\linewidth]{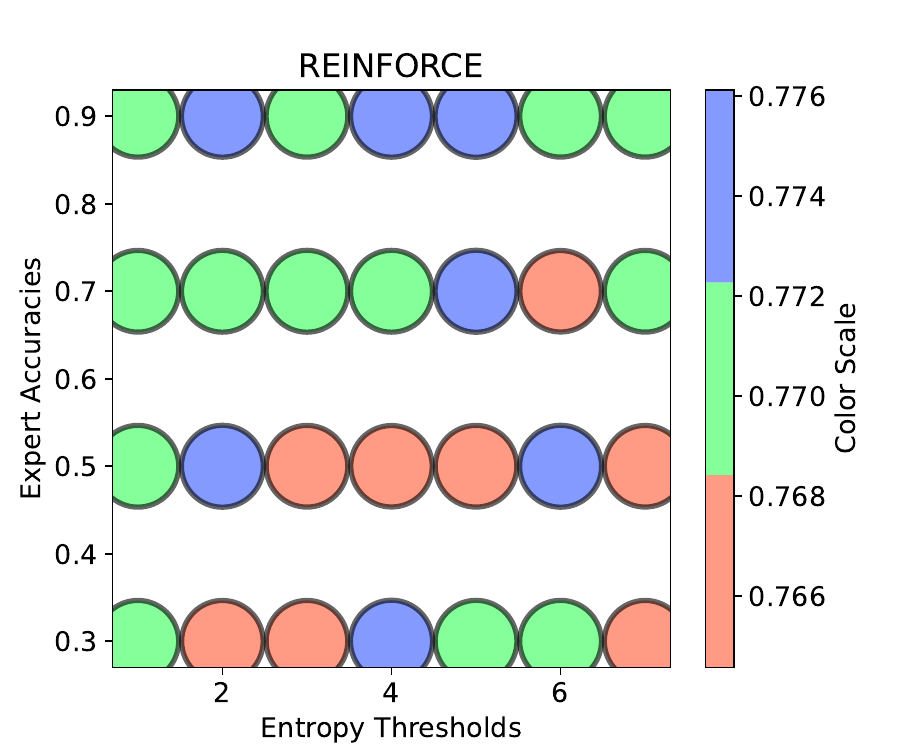}
    \end{subfigure}
    \hfill

    \begin{subfigure}[b]{0.32\columnwidth}
        \centering
        \includegraphics[width=\linewidth]{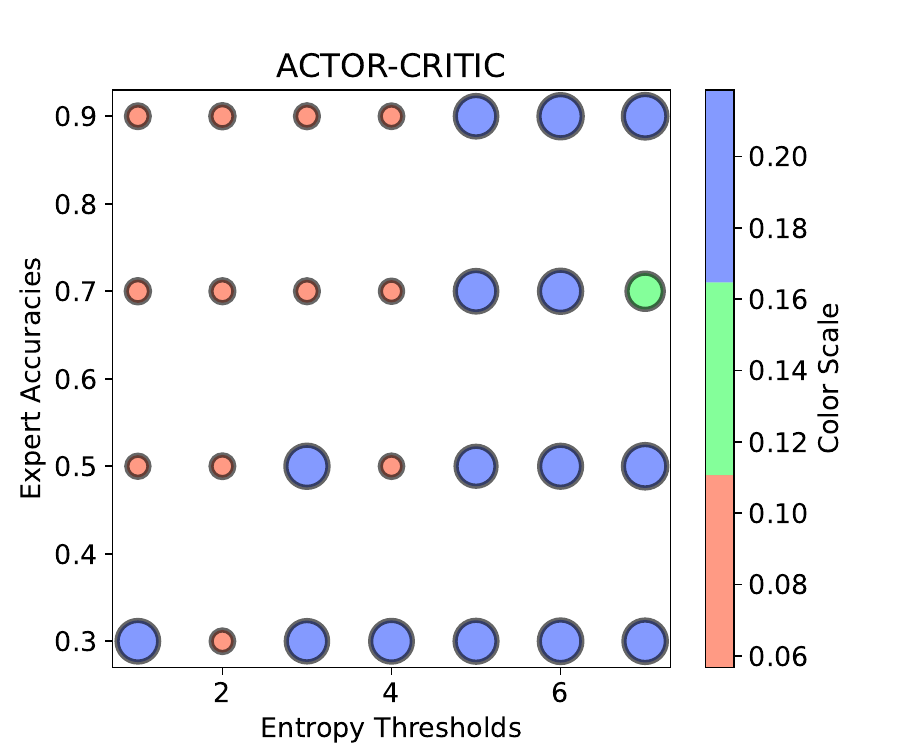}
    \end{subfigure}
    \hfill
    \begin{subfigure}[b]{0.32\columnwidth}
        \centering
        \includegraphics[width=\linewidth]{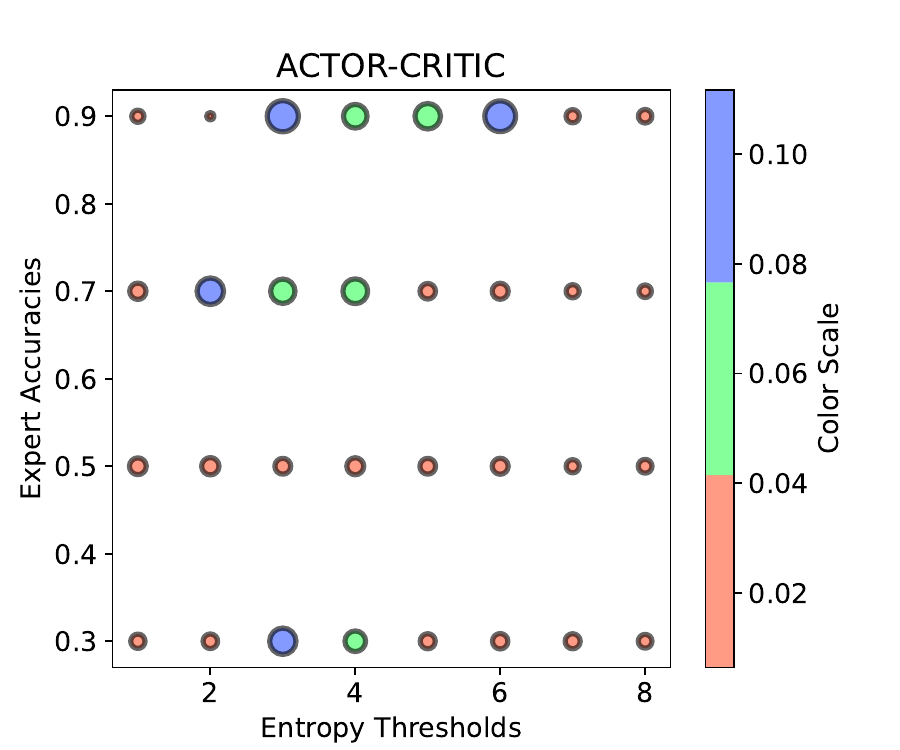}
    \end{subfigure}
    \hfill
    \begin{subfigure}[b]{0.32\columnwidth}
        \centering
        \includegraphics[width=\linewidth]{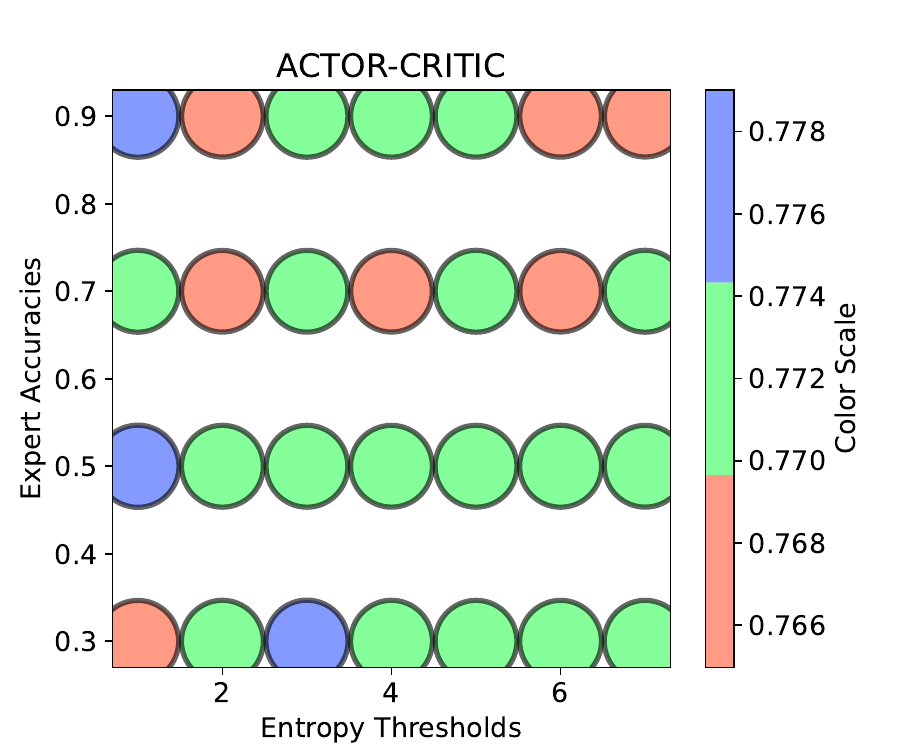}
    \end{subfigure}
    \hfill

\end{figure}

\section{Variation in the percentage of steps for expert queries based on entropy threshold}
\label{sec:app_var_exp_query_percentage}

This section studies the variation of expert queries for the two feedback types.

\captionsetup{font=footnotesize}
\begin{figure}[t]
     \caption{Variation of expert queries made for different models based on entropy for feedback type: Action Recommendation}
    \label{fig:app_var_entr_percentage_ar}
    \centering
    \begin{subfigure}[b]{0.32\columnwidth}
        \centering
        \setlength{\fboxsep}{1pt}\colorbox{lightgray!30}{\textbf{Bibtex}}
        \includegraphics[width=\columnwidth]{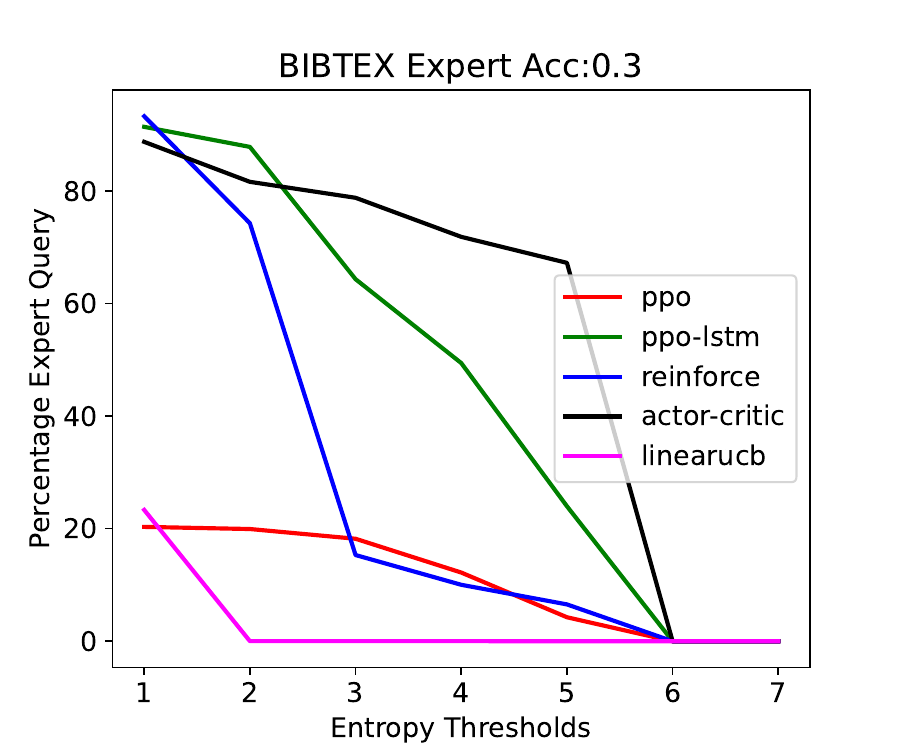}
    \end{subfigure}
    \hfill
    \begin{subfigure}[b]{0.32\columnwidth}
        \centering
        \setlength{\fboxsep}{1pt}\colorbox{lightgray!30}{\textbf{Delicious}}
        \includegraphics[width=\linewidth]{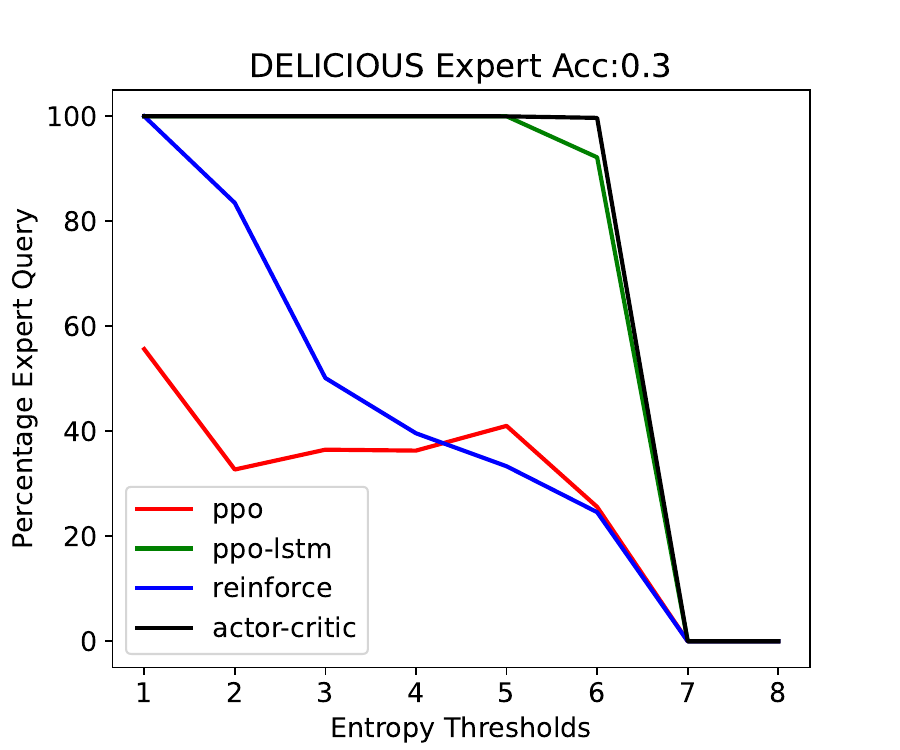}
    \end{subfigure}
    \hfill
    \begin{subfigure}[b]{0.32\columnwidth}
        \centering
        \setlength{\fboxsep}{1pt}\colorbox{lightgray!30}{\textbf{MediaMill}}
        \includegraphics[width=\linewidth]{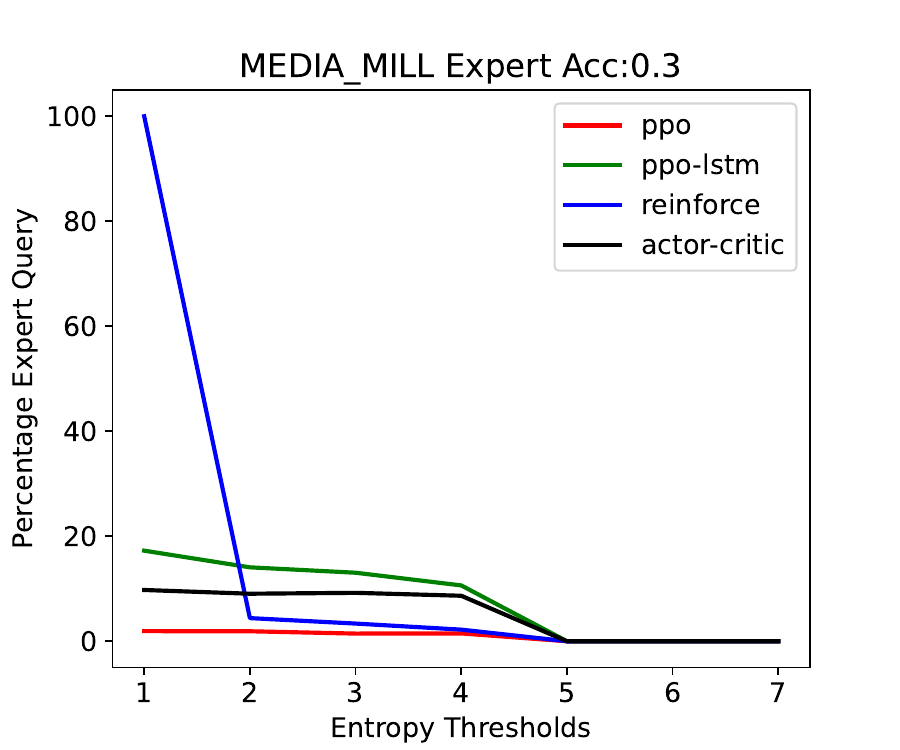}
    \end{subfigure}
    \hfill
    
    \begin{subfigure}[b]{0.32\columnwidth}
        \centering
        \includegraphics[width=\linewidth]{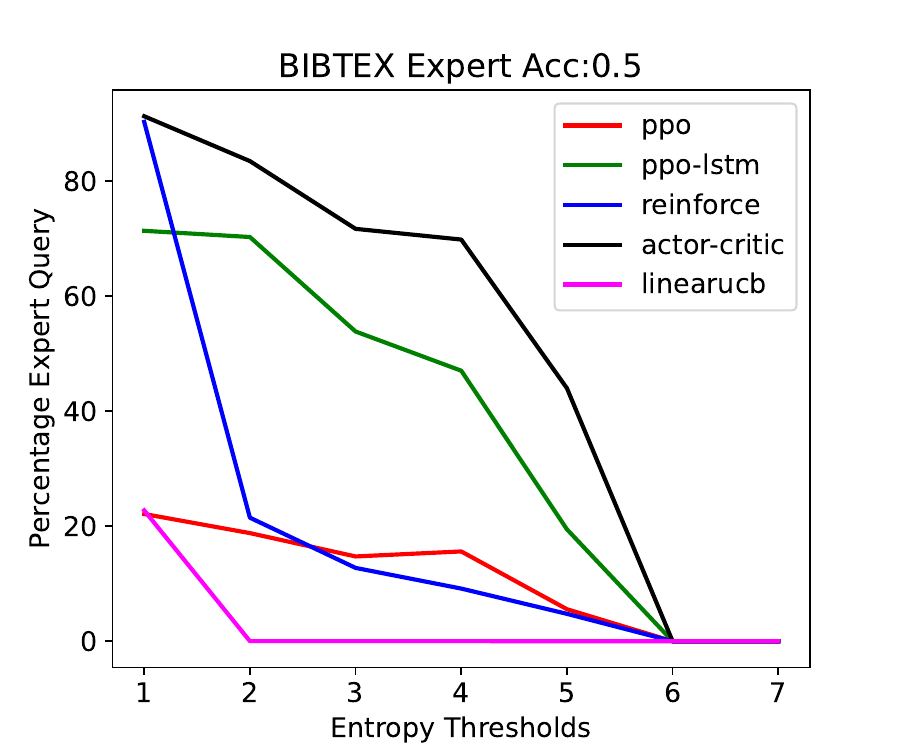}
    \end{subfigure}
    \hfill
    \begin{subfigure}[b]{0.32\columnwidth}
        \centering
        \includegraphics[width=\linewidth]{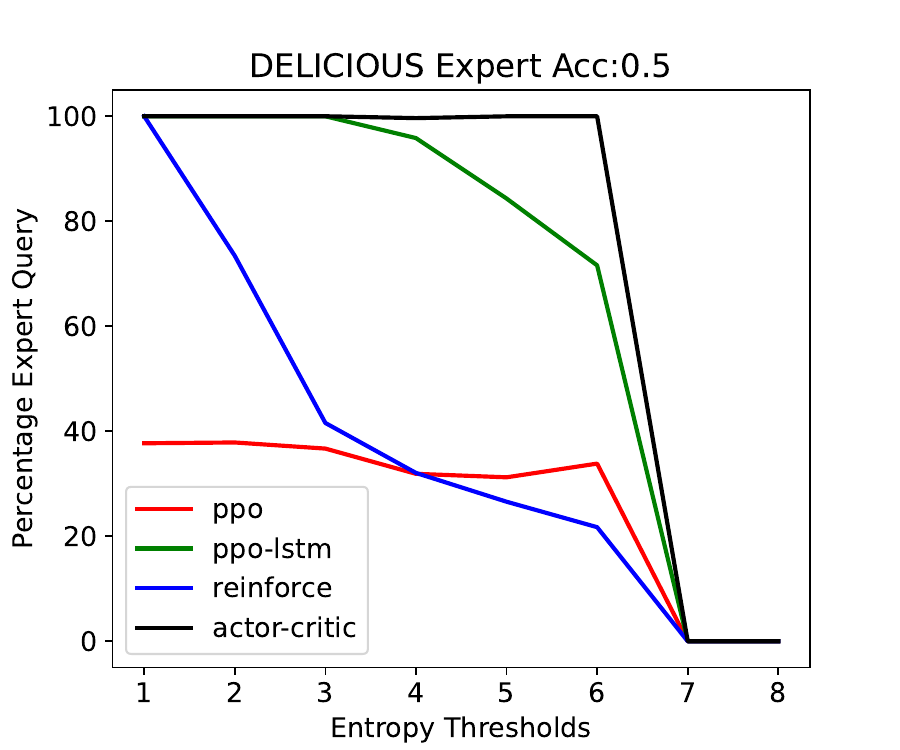}
    \end{subfigure}
    \hfill
    \begin{subfigure}[b]{0.32\columnwidth}
        \centering
        \includegraphics[width=\linewidth]{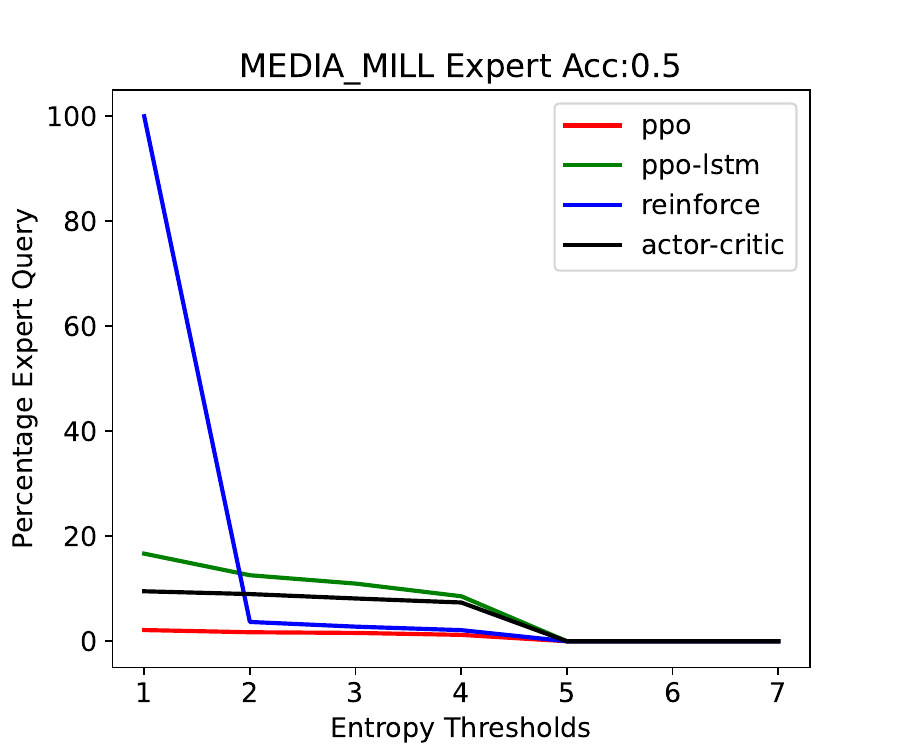}
    \end{subfigure}
    \hfill

    \begin{subfigure}[b]{0.32\columnwidth}
        \centering
        \includegraphics[width=\linewidth]{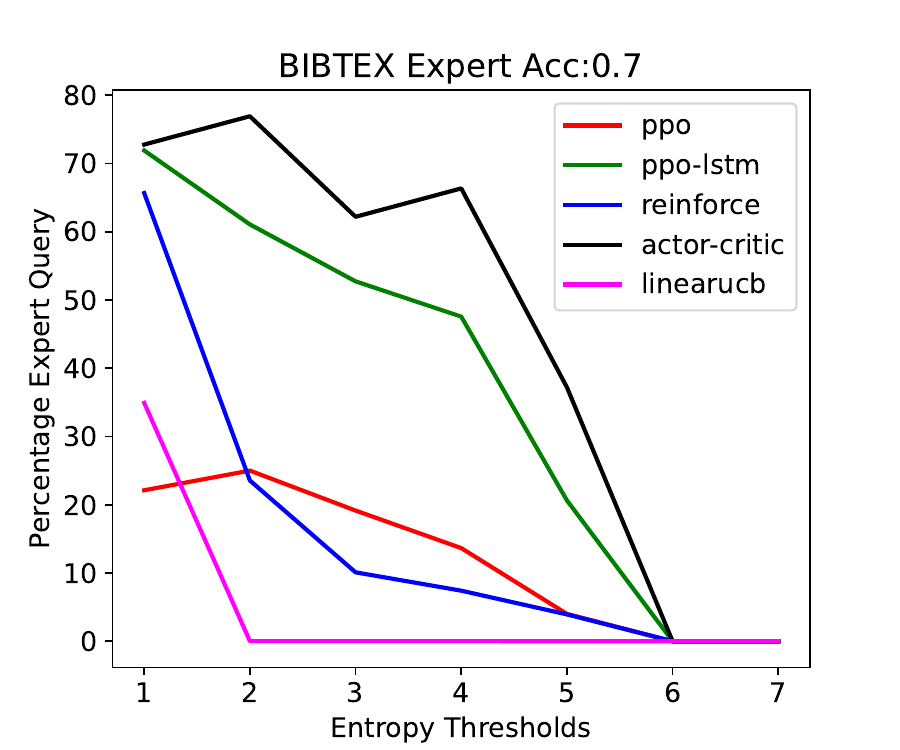}
    \end{subfigure}
    \hfill
    \begin{subfigure}[b]{0.32\columnwidth}
        \centering
        \includegraphics[width=\linewidth]{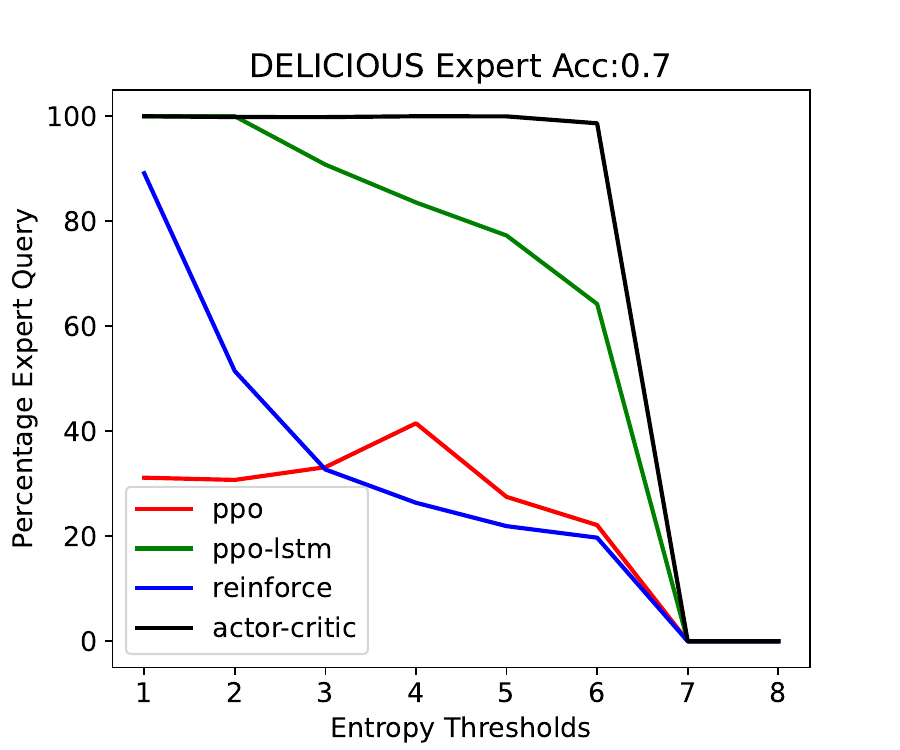}
    \end{subfigure}
    \hfill
    \begin{subfigure}[b]{0.32\columnwidth}
        \centering
        \includegraphics[width=\linewidth]{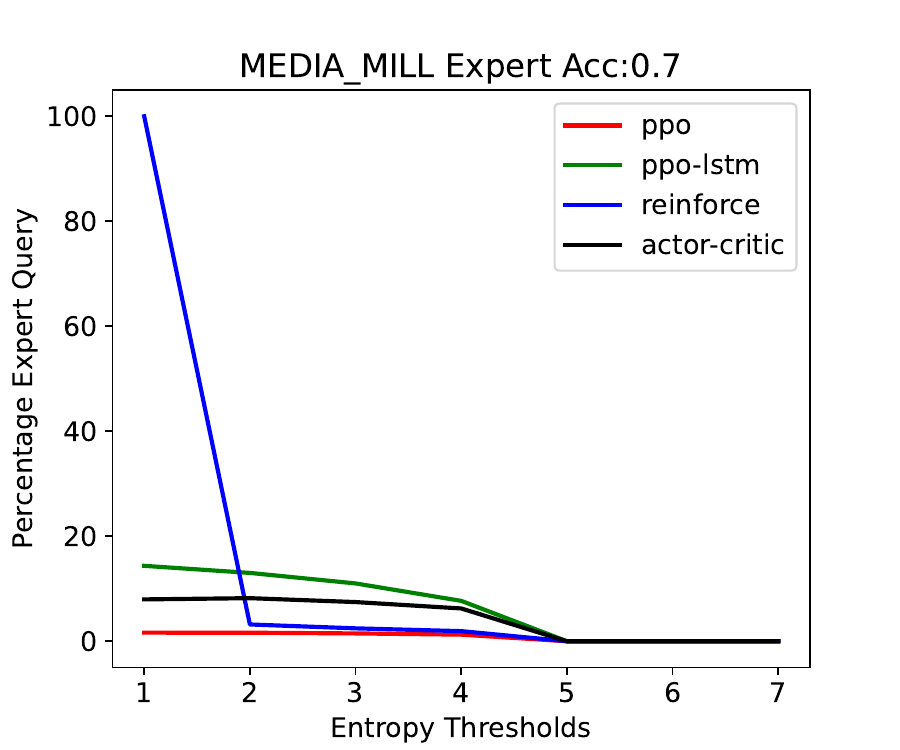}
    \end{subfigure}
    \hfill

    \begin{subfigure}[b]{0.32\columnwidth}
        \centering
        \includegraphics[width=\linewidth]{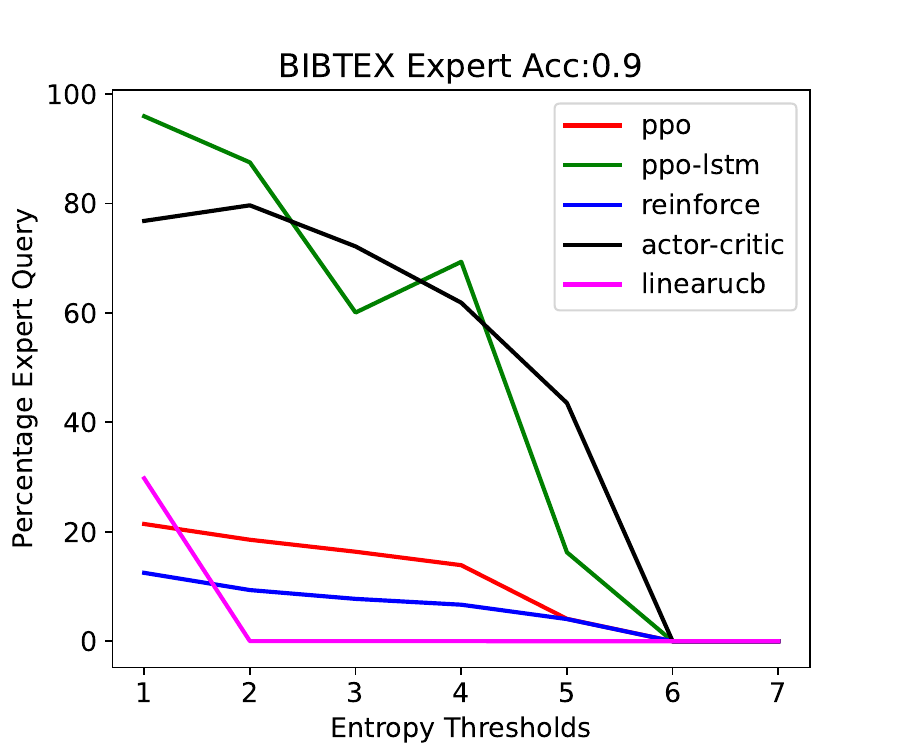}
    \end{subfigure}
    \hfill
    \begin{subfigure}[b]{0.32\columnwidth}
        \centering
        \includegraphics[width=\linewidth]{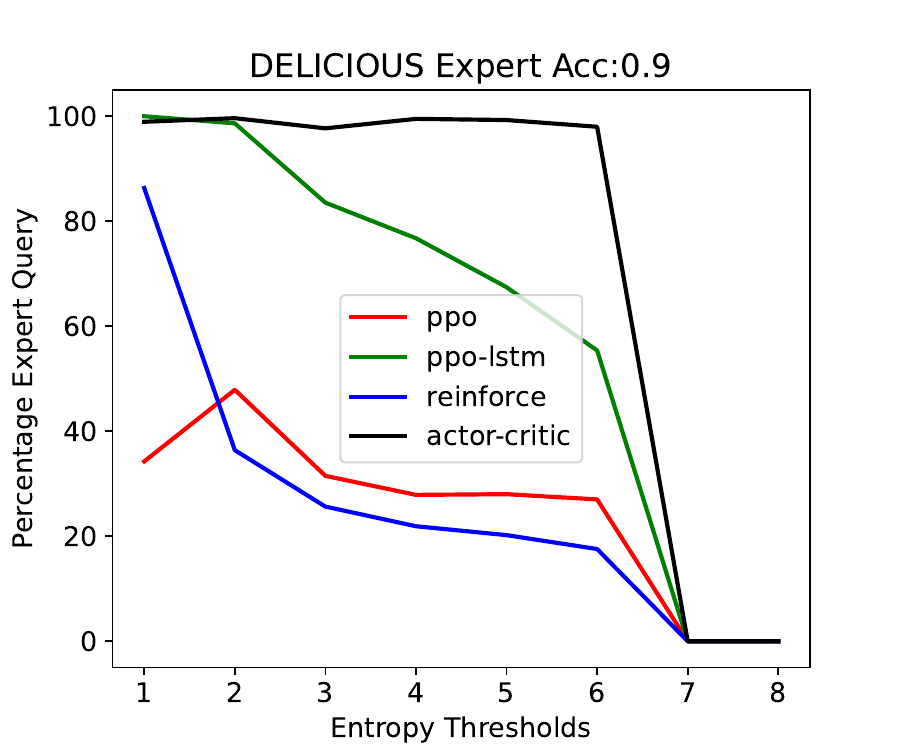}
    \end{subfigure}
    \hfill
    \begin{subfigure}[b]{0.32\columnwidth}
        \centering
        \includegraphics[width=\linewidth]{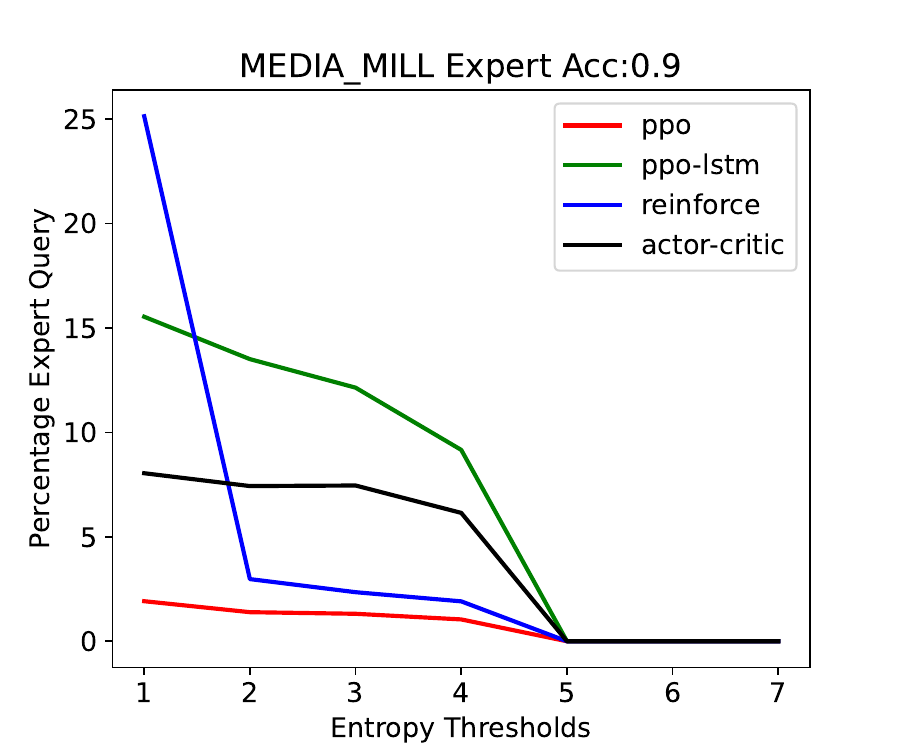}
    \end{subfigure}
    \hfill

\end{figure}


\captionsetup{font=footnotesize}
\begin{figure}[t]
     \caption{Variation of expert queries made for different models based on entropy for feedback type: Reward Manipulation}
    \label{fig:app_var_entr_percentage_rm}
    \centering
    \begin{subfigure}[b]{0.32\columnwidth}
        \centering
        \setlength{\fboxsep}{1pt}\colorbox{lightgray!30}{\textbf{Bibtex}}
        \includegraphics[width=\columnwidth]{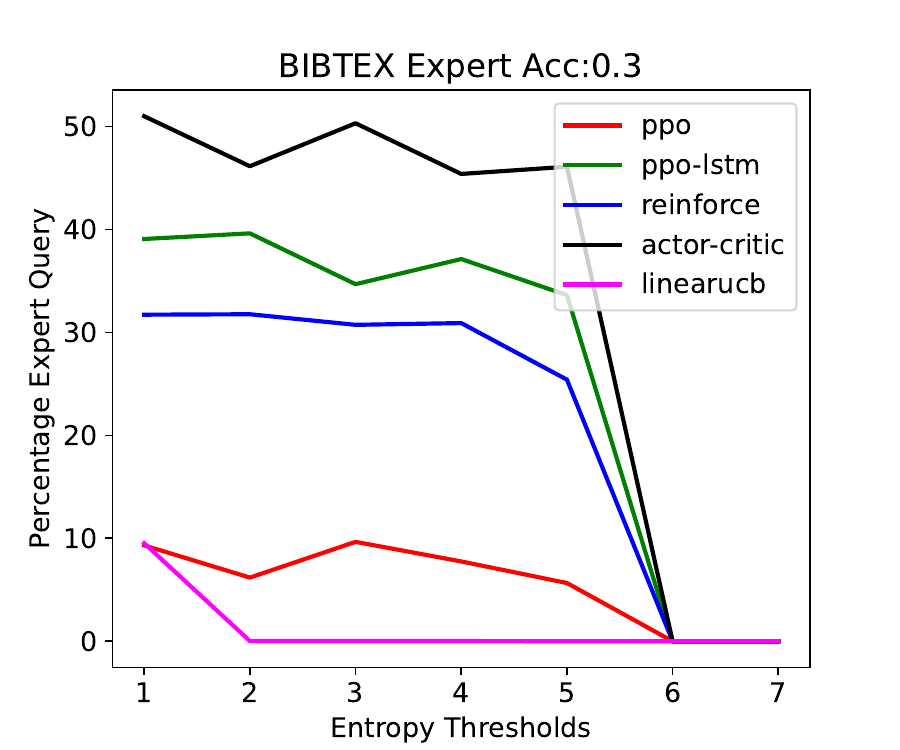}
    \end{subfigure}
    \hfill
    \begin{subfigure}[b]{0.32\columnwidth}
        \centering
        \setlength{\fboxsep}{1pt}\colorbox{lightgray!30}{\textbf{Delicious}}
        \includegraphics[width=\linewidth]{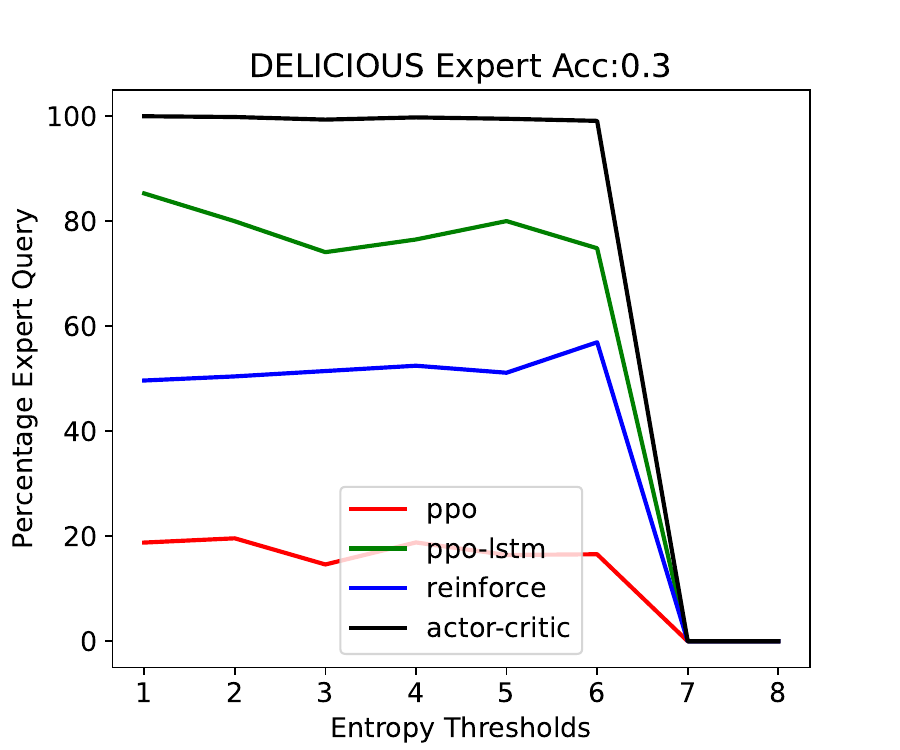}
    \end{subfigure}
    \hfill
    \begin{subfigure}[b]{0.32\columnwidth}
        \centering
        \setlength{\fboxsep}{1pt}\colorbox{lightgray!30}{\textbf{MediaMill}}
        \includegraphics[width=\linewidth]{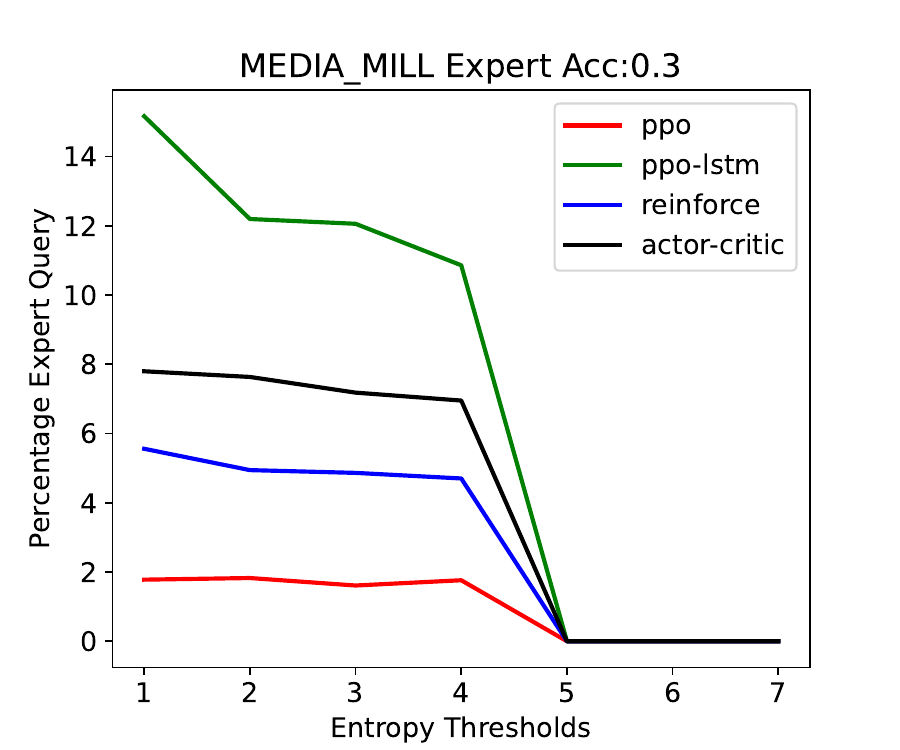}
    \end{subfigure}
    \hfill
    
    \begin{subfigure}[b]{0.32\columnwidth}
        \centering
        \includegraphics[width=\linewidth]{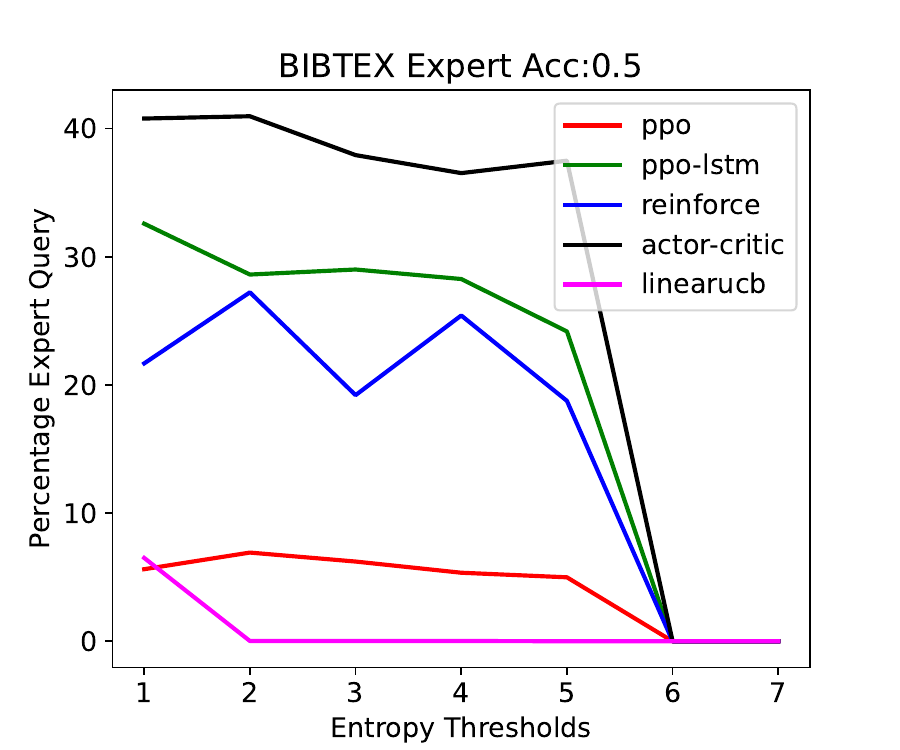}
    \end{subfigure}
    \hfill
    \begin{subfigure}[b]{0.32\columnwidth}
        \centering
        \includegraphics[width=\linewidth]{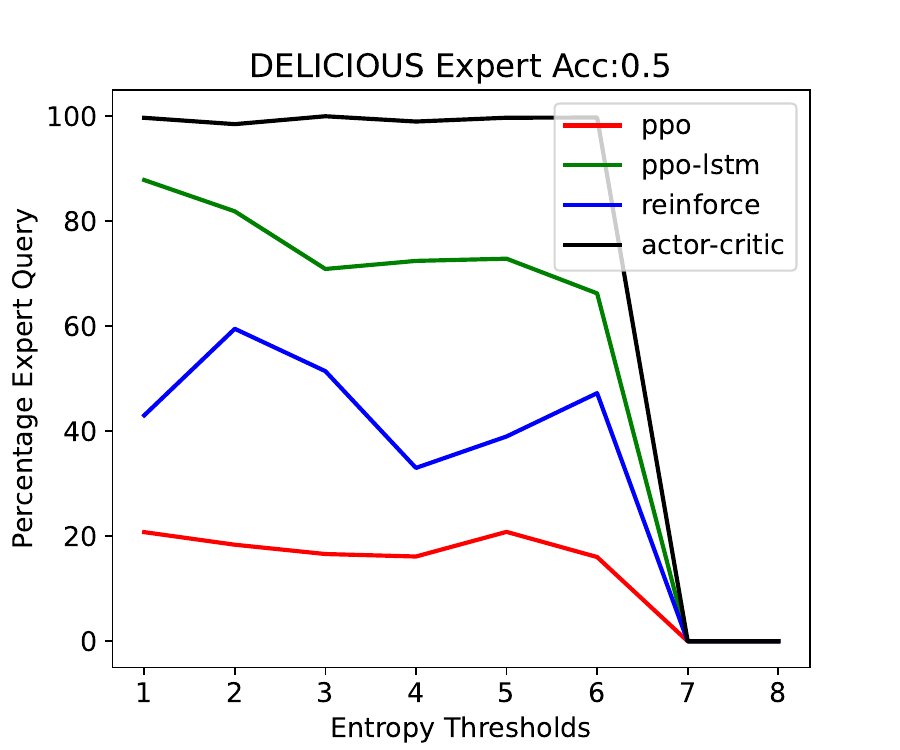}
    \end{subfigure}
    \hfill
    \begin{subfigure}[b]{0.32\columnwidth}
        \centering
        \includegraphics[width=\linewidth]{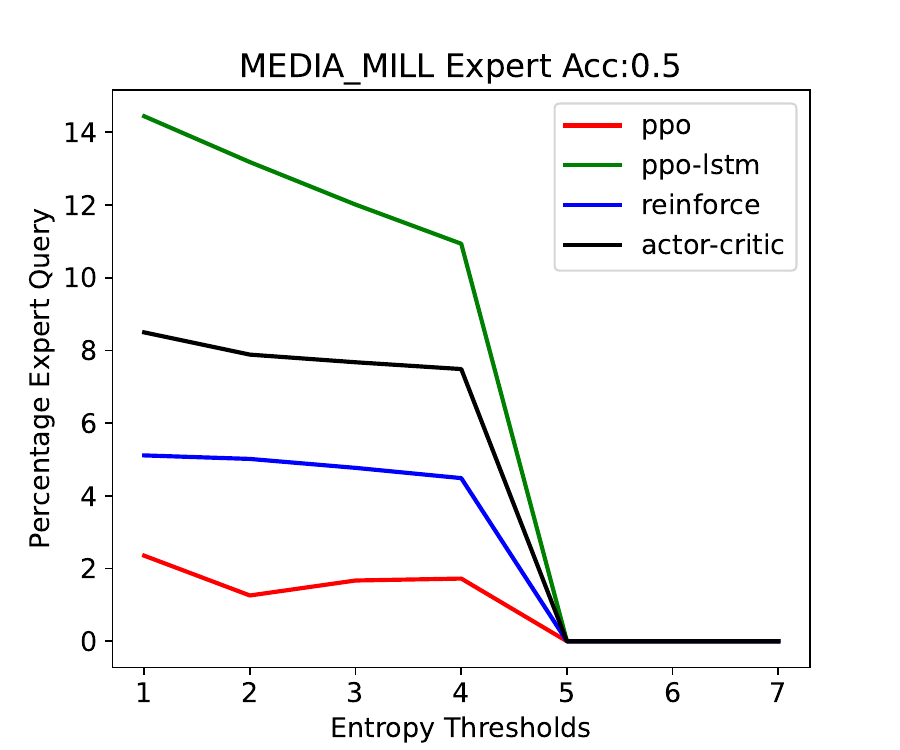}
    \end{subfigure}
    \hfill

    \begin{subfigure}[b]{0.32\columnwidth}
        \centering
        \includegraphics[width=\linewidth]{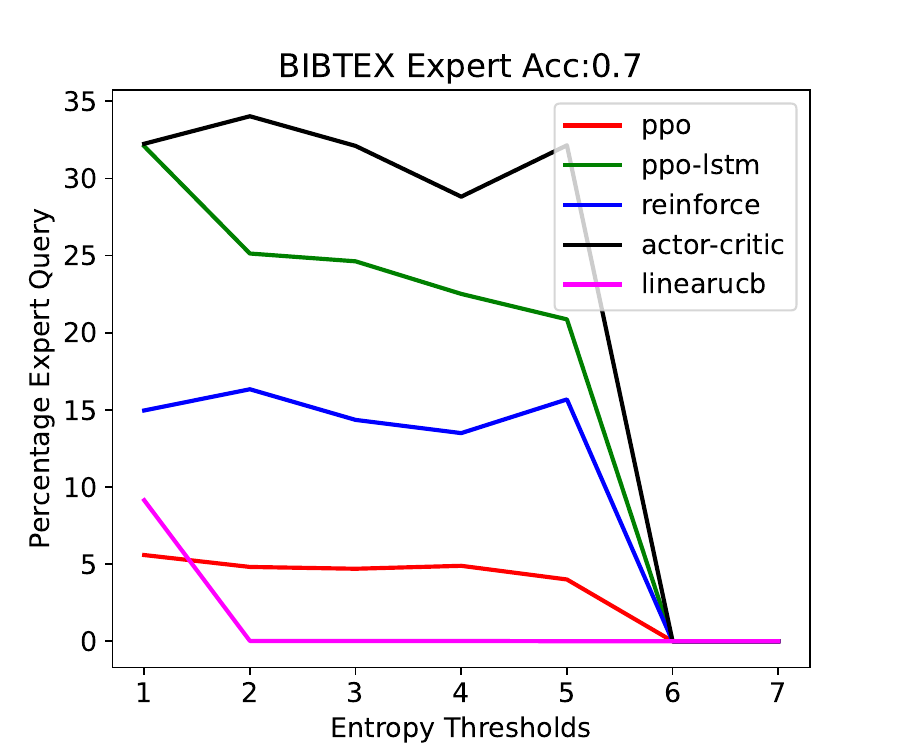}
    \end{subfigure}
    \hfill
    \begin{subfigure}[b]{0.32\columnwidth}
        \centering
        \includegraphics[width=\linewidth]{figures/entropy_vs_feedback_percentage/bibtex/reward_penalty_accuracy/reward_penalty_accuracy_hf_percentage_expert_acc_0.7.pdf}
    \end{subfigure}
    \hfill
    \begin{subfigure}[b]{0.32\columnwidth}
        \centering
        \includegraphics[width=\linewidth]{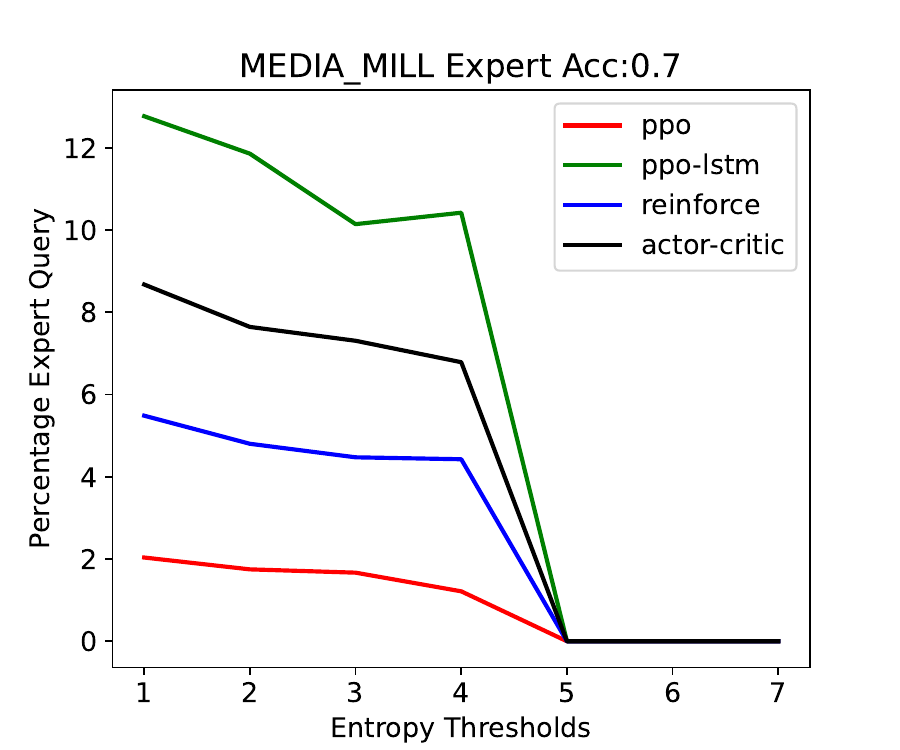}
    \end{subfigure}
    \hfill

    \begin{subfigure}[b]{0.32\columnwidth}
        \centering
        \includegraphics[width=\linewidth]{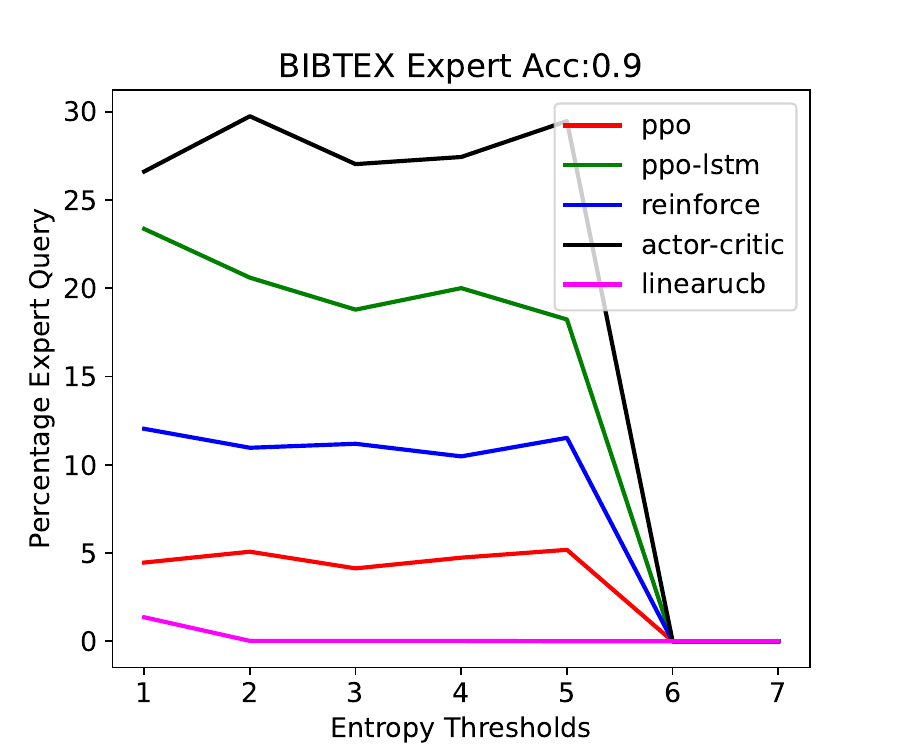}
    \end{subfigure}
    \hfill
    \begin{subfigure}[b]{0.32\columnwidth}
        \centering
        \includegraphics[width=\linewidth]{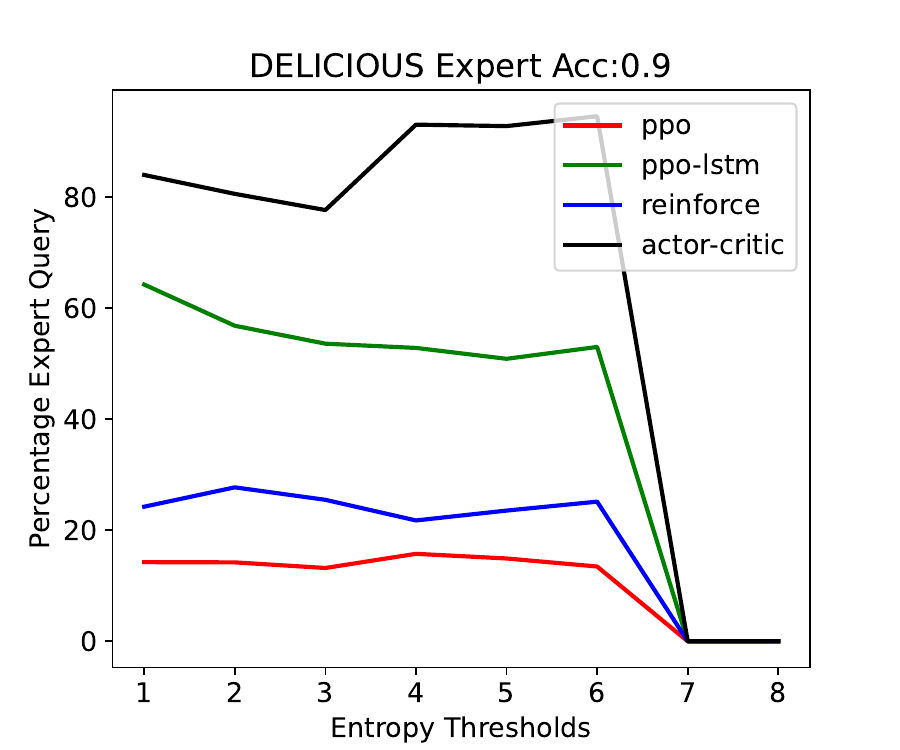}
    \end{subfigure}
    \hfill
    \begin{subfigure}[b]{0.32\columnwidth}
        \centering
        \includegraphics[width=\linewidth]{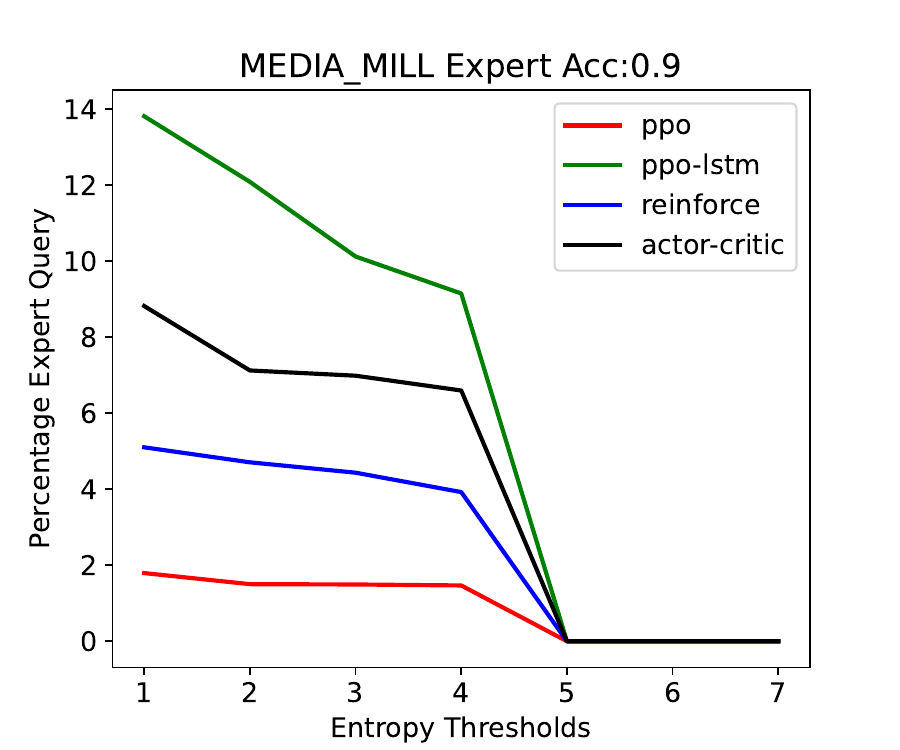}
    \end{subfigure}
    \hfill

\end{figure}
\section{Performance of different algorithms based on different expert levels}

This section provides more details on the performance as a function of different expert levels.
\clearpage
\newpage
\begin{table*}[!htbp]
    \centering
    \caption{Performance comparison of algorithms for different quality of expert feedback. The values in bold represent the maximum mean cumulative reward achieved at different levels of expert.} 
    \label{table:app_expert_range}
    
    \resizebox{\textwidth}{!}{%
    \begin{tabular}{lllcccc}
        \toprule
        \textbf{Feedback Type} & \textbf{Alg. Name} & \textbf{Env. Name} & \multicolumn{1}{c}{\textbf{0.3}} & \multicolumn{1}{c}{\textbf{0.5}} & \multicolumn{1}{c}{\textbf{0.7}} & \multicolumn{1}{c}{\textbf{0.9}} \\
        \midrule
        AR & PPO & Bibtex & \bm{$0.27349 \pm 0.00167$} & $0.26383 \pm 0.00091$ & $0.20268 \pm 0.00260$ & $0.16763 \pm 0.00092$ \\[5pt]
        RM & PPO & Bibtex & $0.27827 \pm 0.00312$ & $0.27470 \pm 0.00165$ & $0.16965 \pm 0.00202$ & $\mathbf{0.31021 \pm 0.00278}$ \\[5pt]
        AR & PPO & Media\_Mill & $0.76862 \pm 0.00137$ & $0.76842 \pm 0.00230$ & $\mathbf{0.77206 \pm 0.00124}$ & $0.76662 \pm 0.00134$ \\[5pt]
        RM & PPO & Media\_Mill & $0.76683 \pm 0.00190$ & $0.76530 \pm 0.00128$ & $0.76895 \pm 0.00291$ & $\mathbf{0.77545 \pm 0.00151}$ \\[5pt]
        AR & PPO & Delicious & $\mathbf{0.51770 \pm 0.00220}$ & $0.36824 \pm 0.00191$ & $0.37114 \pm 0.00208$ & $0.46170 \pm 0.00130$ \\[5pt]
        RM & PPO & Delicious & $\mathbf{0.48187 \pm 0.00113}$ & $0.29682 \pm 0.00230$ & $0.36717 \pm 0.00215$ & $0.40190 \pm 0.00165$ \\[5pt]
        AR & PPO-LSTM & Bibtex & $\mathbf{0.13464 \pm 0.00086}$ & $0.11283 \pm 0.00204$ & $0.11533 \pm 0.00090$ & $0.02363 \pm 0.00063$ \\[5pt]
        RM & PPO-LSTM & Bibtex & $\mathbf{0.14413 \pm 0.00052}$ & $0.14157 \pm 0.00186$ & $0.13750 \pm 0.00095$ & $0.14304 \pm 0.00136$ \\[5pt]
        AR & PPO-LSTM & Media\_Mill & $0.76836 \pm 0.00155$ & $0.77318 \pm 0.00141$ & $\mathbf{0.77504 \pm 0.00058}$ & $0.77113 \pm 0.00120$ \\[5pt]
        RM & PPO-LSTM & Media\_Mill & $0.76973 \pm 0.00114$ & $\mathbf{0.77447 \pm 0.00177}$ & $0.76748 \pm 0.00187$ & $0.76197 \pm 0.00373$ \\[5pt]
        AR & PPO-LSTM & Delicious & $\mathbf{0.12497 \pm 0.00140}$ & $0.11567 \pm 0.00091$ & $0.11793 \pm 0.00203$ & $0.11698 \pm 0.00100$ \\[5pt]
        RM & PPO-LSTM & Delicious & $\mathbf{0.28802 \pm 0.00123}$ & $0.26663 \pm 0.00204$ & $0.09600 \pm 0.00092$ & $0.26014 \pm 0.00151$ \\[5pt]
        AR & Reinforce & Bibtex & $0.24346 \pm 0.00128$ & $0.27678 \pm 0.00159$ & $\mathbf{0.29793 \pm 0.00134}$ & $0.11714 \pm 0.00133$ \\[5pt]
        RM & Reinforce & Bibtex & $0.21970 \pm 0.00090$ & $0.24939 \pm 0.00148$ & $0.25543 \pm 0.00166$ & \bm{$0.25662 \pm 0.00137$} \\[5pt]
        AR & Reinforce & Media\_Mill & $0.08715 \pm 0.00139$ & $0.35710 \pm 0.00214$ & $0.63323 \pm 0.00296$ & \bm{$0.63446 \pm 0.00155$} \\[5pt]
        RM & Reinforce & Media\_Mill & $0.77292 \pm 0.00310$ & $0.77098 \pm 0.00177$ & $0.77183 \pm 0.00111$ & $\mathbf{0.77339 \pm 0.00129}$ \\[5pt]
        AR & Reinforce & Delicious & $\mathbf{0.37394 \pm 0.00165}$ & $0.35349 \pm 0.00121$ & $0.37268 \pm 0.00230$ & $0.24432 \pm 0.00258$ \\[5pt]
        RM & Reinforce & Delicious & $0.04502 \pm 0.00067$ & $\mathbf{0.15057 \pm 0.00138}$ & $0.07441 \pm 0.00142$ & $0.07983 \pm 0.00091$ \\[5pt]
        AR & Actor-Critic & Bibtex & $0.14119 \pm 0.00107$ & $0.21240 \pm 0.00068$ & $\mathbf{0.23825 \pm 0.00093}$ & $0.15231 \pm 0.00208$ \\[5pt]
        RM & Actor-Critic & Bibtex & $0.17242 \pm 0.00126$ & $\mathbf{0.23110 \pm 0.00149}$ & $0.19961 \pm 0.00119$ & $0.19822 \pm 0.00149$ \\[5pt]
        AR & Actor-Critic & Media\_Mill & $0.76394 \pm 0.00118$ & $\mathbf{0.77449 \pm 0.00242}$ & $0.76325 \pm 0.00085$ & $0.76966 \pm 0.00076$ \\[5pt]
        RM & Actor-Critic & Media\_Mill & $0.76749 \pm 0.00205$ & $0.77507 \pm 0.00124$ & \bm{$0.77664 \pm 0.00099$} & $0.76347 \pm 0.00203$ \\[5pt]
        AR & Actor-Critic & Delicious & $0.02017 \pm 0.00084$ & $0.02213 \pm 0.00031$ & $0.02629 \pm 0.00054$ & $\mathbf{0.03498 \pm 0.00036}$ \\[5pt]
        RM & Actor-Critic & Delicious & $0.02292 \pm 0.00051$ & $0.02334 \pm 0.00070$ & \bm{$0.02354 \pm 0.00034$} & $0.02154 \pm 0.00069$ \\[5pt]
        AR & LinearUCB & Bibtex & $\mathbf{0.02478 \pm 0.00068}$ & $0.02280 \pm 0.00056$ & $0.02145 \pm 0.00066$ & $0.02002 \pm 0.00055$ \\[5pt]
        RM & LinearUCB & Bibtex & $0.02369 \pm 0.00080$ & $0.02532 \pm 0.00079$ & $0.02518 \pm 0.00049$ & $\mathbf{0.03527 \pm 0.00115}$ \\[5pt]
        AR & LinearUCB & Media\_Mill & $0.00321 \pm 0.00028$ & $0.00259 \pm 0.00029$ & $\mathbf{0.17961 \pm 0.00117}$ & $0.17399 \pm 0.00084$ \\[5pt]
        RM & LinearUCB & Media\_Mill & $0.00059 \pm 0.00004$ & $0.00058 \pm 0.00007$ & $\mathbf{0.19890 \pm 0.00087}$ & $0.05337 \pm 0.00136$ \\[5pt]
        AR & LinearUCB & Delicious & $\mathbf{0.02430 \pm 0.00053}$ & $0.01818 \pm 0.00036$ & $0.02064 \pm 0.00061$ & $0.05308 \pm 0.00066$ \\[5pt]
        RM & LinearUCB & Delicious & $0.01664 \pm 0.00022$ & $\mathbf{0.10018 \pm 0.00161}$ & $0.01889 \pm 0.00051$ & $0.08540 \pm 0.00063$ \\[5pt]
        AR & Bootstrapped-TS & Bibtex & $0.22537 \pm 0.00196$ & $0.19911 \pm 0.00105$ & $0.21668 \pm 0.00144$ & $\mathbf{0.24097 \pm 0.00137}$ \\[5pt]
        RM & Bootstrapped-TS & Bibtex & $0.15276 \pm 0.00101$ & $\mathbf{0.27697 \pm 0.00103}$ & $0.18423 \pm 0.00087$ & $0.18468 \pm 0.00278$ \\[5pt]
        AR & Bootstrapped-TS & Media\_Mill & $0.00000 \pm 0.00000$ & $0.00000 \pm 0.00000$ & $0.00000 \pm 0.00000$ & $0.00000 \pm 0.00000$ \\[5pt]
        RM & Bootstrapped-TS & Media\_Mill & $0.00000 \pm 0.00000$ & $0.00000 \pm 0.00000$ & $0.00000 \pm 0.00000$ & $0.00000 \pm 0.00000$ \\[5pt]
        AR & Bootstrapped-TS & Delicious & $0.00000 \pm 0.00000$ & $0.00000 \pm 0.00000$ & $0.00000 \pm 0.00000$ & $0.00000 \pm 0.00000$ \\[5pt]
        RM & Bootstrapped-TS & Delicious & $0.00000 \pm 0.00000$ & $0.00000 \pm 0.00000$ & $0.00000 \pm 0.00000$ & $0.00000 \pm 0.00000$ \\
        \bottomrule
    \end{tabular}%
    }
\end{table*}

\section{Hyper parameters}
We provide the hyperparameters for the policy based RL algorithms and the range of values of entropy thresholds that we consider for each dataset. 
\subsection{Hyperparameters for policy based RL algorithms}
\label{sec:app_hyperparams_rl}
\begin{table*}[!htbp]
    \centering
    \caption{HyperParameters for Policy based Algorithms. AFD=Advantage function discount.}
    \vspace{+2mm}
     \resizebox{\columnwidth}{!}{%
    \begin{tabular}{lccccc}
        \toprule
        \textbf{Algorithms} & \textbf{Training Epochs} & \textbf{Learning Rate} & \textbf{AFD} & \textbf{Clipping} & \textbf{Batch Size} \\
        \midrule
        PPO          & $5000$ & $0.005$  & $0.1$  & $0.1$ & $32$ \\
        PPO-LSTM     & $5000$ & $0.001$  & $0.95$ &$ 0.1$ & $32$ \\
        Reinforce    & $5000$ & $0.0002$ & -    & -   & -  \\
        Actor Critic & $5000$ & $0.002$  & -    & -   & $32$ \\
        \bottomrule
    \end{tabular}%
    }
    \label{tab:hyperparams_policy_based_algo}
\end{table*}

\subsection{Range of entropy thresholds considered}
\label{subsec:entropy_range}
\begin{table}[H]
    \centering
    \caption{Entropy thresholds for different environments $\lambda$}
    \begin{tabular}{lc}
        \toprule
        \textbf{Item} & \textbf{$\lambda$ values} \\
        \midrule
        Bibtex & $2.5$, $3.5$, $5.0$, $6.5$, $9.0$ \\
        Media Mill & $1.5$, $2.5$, $3.0$, $4.5$, $7.0$ \\
        Delicious & $1.5$, $2.5$, $4.5$, $6.5$, $9.0$ \\
        Yahoo & $1.5$, $2.5$, $4.5$, $7.0$, $9.0$ \\
        \bottomrule
    \end{tabular}
    \label{tab:lambda_values}
\end{table}



\end{document}